\documentclass[acmsmall,screen,nonacm]{acmart}
\usepackage{multirow}
\usepackage{graphicx} 
\usepackage{subfigure}
\AtBeginDocument{%
  }

\setcopyright{acmlicensed}
\copyrightyear{2018}
\acmYear{2018}
\acmDOI{XXXXXXX.XXXXXXX}

\acmJournal{TIST}
\acmVolume{37}
\acmNumber{4}
\acmArticle{111}
\acmMonth{8}

\begin{document}

%%
%% The "title" command has an optional parameter,
%% allowing the author to define a "short title" to be used in page headers.
\title{Causal Flow-based Variational Auto-Encoder for Disentangled Causal Representation Learning}

%%
%% The "author" command and its associated commands are used to define
%% the authors and their affiliations.
%% Of note is the shared affiliation of the first two authors, and the
%% "authornote" and "authornotemark" commands
%% used to denote shared contribution to the research.
\author{Di Fan}
\email{fandi@zju.edu.cn}
\orcid{0009-0001-6357-7849}
\affiliation{
  \institution{School of Mathematical Sciences, Zhejiang
University}
  % \streetaddress{866 Yuhangtang rd}
  \city{Xihu Qu}
  \state{Hangzhou}
  \country{China}
  \postcode{310012}
}

\author{Yannian Kou}
\orcid{0009-0001-3100-1570}
\affiliation{
  \institution{School of Mathematical Sciences, Zhejiang
University}
  % \streetaddress{866 Yuhangtang rd}
  \city{Xihu Qu}
  \state{Hangzhou}
  \country{China}
  \postcode{310012}
}
\email{kouyannian@zju.edu.cn}

\author{Chuanhou Gao}
\authornote{Corresponding author: Chuanhou Gao
}
\orcid{0000-0001-9030-2042}
\affiliation{
  \institution{School of Mathematical Sciences, Zhejiang
University}
  % \streetaddress{866 Yuhangtang rd}
  \city{Xihu Qu}
  \state{Hangzhou}
  \country{China}
  \postcode{310012}
}
\email{gaochou@zju.edu.cn}

%%
%% By default, the full list of authors will be used in the page
%% headers. Often, this list is too long, and will overlap
%% other information printed in the page headers. This command allows
%% the author to define a more concise list
%% of authors' names for this purpose.
\renewcommand{\shortauthors}{Fan et al.}

%%
%% The abstract is a short summary of the work to be presented in the
%% article.
\begin{abstract}
% Disentangled representation learning aims to learn low-dimensional representations of data, where each dimension corresponds to an underlying generative factor. Currently, Variational Auto-Encoder (VAE) is widely used for disentangled representation learning, with the majority of methods assuming independence among generative factors. However, in real-world scenarios, generative factors typically exhibit complex causal relationships. We thus design  a new supervised learning technique for the VAE-based framework called the Disentangled Causal Variational Auto-Encoder (DCVAE), which incorporates a variant of autoregressive flows referred to as causal flows, capable of learning effective causal disentangled representations and generate causally disentangled outcomes. adding supervised regularization 保证了Our model can effectively learn causal disentangled representations. The performance of DCVAE is evaluated on both synthetic and real-world datasets, demonstrating its outstanding capability in achieving causal disentanglement and performing intervention experiments. Moreover, DCVAE exhibits remarkable performance on downstream tasks and has the potential to learn the true causal structure among factors.
Disentangled representation learning aims to learn low-dimensional representations where each dimension corresponds to an underlying generative factor. While the Variational Auto-Encoder (VAE) is widely used for this purpose, most existing methods assume independence among factors, a simplification that does not hold in many real-world scenarios where factors are often interdependent and exhibit causal relationships. To overcome this limitation, we propose the Disentangled Causal Variational Auto-Encoder (DCVAE), a novel supervised VAE framework that integrates causal flows into the representation learning process, enabling the learning of more meaningful and interpretable disentangled representations. We evaluate DCVAE on both synthetic and real-world datasets, demonstrating its superior ability in causal disentanglement and intervention experiments. Furthermore, DCVAE outperforms state-of-the-art methods in various downstream tasks, highlighting its potential for learning true causal structures among factors.
\end{abstract}

%%
%% The code below is generated by the tool at http://dl.acm.org/ccs.cfm.
%% Please copy and paste the code instead of the example below.
%%


%%
%% Keywords. The author(s) should pick words that accurately describe
%% the work being presented. Separate the keywords with commas.

\received{20 February 2007}
\received[revised]{12 March 2009}
\received[accepted]{5 June 2009}

%%
%% This command processes the author and affiliation and title
%% information and builds the first part of the formatted document.
\maketitle
\section{Introduction}
Representation learning aims to learn data representations that simplify the extraction of information for constructing classifiers or predictors \citep{bengio2013representation}. Disentangled representation learning, a key advancement in this field, aims to factorize representations to effectively identify and disentangle latent factors in observed data \citep{locatello2019challenging}. Its ability to generate robust and interpretable representations has made it vital in domains such as computer vision and recommendation systems \citep{hsieh2018learning,ma2019learning,mu2021knowledge,zhang2021hybrid,yang2022disentangled,wang2023disentangled,jia2024self}.

One widely used framework for learning disentangled representations is the Variational Auto-Encoder (VAE) \citep{kingma2013auto}, which has gained significant popularity \citep{shao2011variational,gupta2023neural,sadok2024a}. The primary focus of existing research is imposing independent constraints on the posterior or aggregated posterior of latent variables $\mathbf{z}$ through KL divergence \citep{higgins2017beta, burgess2018understanding, kim2018disentangling, chen2018isolating, kumar2017variational, kim2019relevance, dupont2018learning}. However, these methods assume the independence of generative factors, which often does not align with real-world data, where latent factors are likely to have causal relationships. For example, in a human face image dataset, factors like smiling and mouth opening exhibit a causal relationship: smiling increases the probability of an open mouth, with smiling acting as the causal variable and mouth opening as the effect. Models based on independence assumptions struggle to learn disentangled representations in such cases \citep{trauble2021disentangled}. To address this, recent approaches have shifted towards causal disentangled representation learning \citep{suter2019robustly, reddy2022causally, brehmer2022weakly, buchholz2023learning, lippe2022citris, seigal2022linear}, and \citep{locatello2019challenging} highlights the importance of supervised learning for disentanglement, suggesting that unsupervised methods are insufficient. Consequently, Structural Causal Models (SCMs) and supervised methods are key to constructing a latent space that causally aligns with ground-truth factors \citep{shen2022weakly, yang2021causalvae, an2023causally}. However, these approaches either rely on pre-existing SCMs to impose causal relationships or focus on modeling the causal structure of generative factors in specific scenarios. In contrast, we propose incorporating causal relationships into the model using normalizing flows, a recent deep learning technique, thus eliminating the reliance on pre-existing SCMs.

Normalizing flows \citep{papamakarios2021normalizing} offer a flexible framework for constructing generative models, ensuring efficient sampling and accurate density estimation. Recently, they have gained prominence in the context of using VAE, where they enhance inference by approximating the posterior distribution more closely to the true data distribution \citep{rezende2015variational}. Among the various types of normalizing flows, autoregressive flows are particularly favored in VAE due to their ability to produce triangular Jacobian matrices \citep{huang2018neural, kingma2016improved}. Autoregressive flows have shown promise in learning causal orders between two variables
or pairs of multivariate variables, particularly in causal discovery tasks \citep{khemakhem2021causal}.
\citet{wehenkel2021graphical} introduced a graphical normalizing flow model, incorporating a conditioner based on an adjacency matrix to improve density estimation from a Bayesian network perspective, though without focusing on causal disentangled representations. Inspired by these advancements, we propose a novel approach, i.e., causal flows based on autoregressive flows. This model not only retains the benefits of traditional autoregressive flows in enhancing VAE’s inference capability but also incorporates causal structural information, enabling VAE to learn causal disentangled representations during inference. This represents the first application of flow models to this task.

In this paper, we introduce causal flows to capture causal relationships among generative factors. We then present a novel VAE-based framework, the Disentangled Causal Variational Auto-Encoder (DCVAE), which integrates VAE with causal flows to learn causal disentangled representations. After encoding the input data with DCVAE's encoder and processing it through the causal flows, we obtain the causal disentangled representations, which are subsequently fed into the decoder for reconstructing the original images. To increase the flexibility of the prior distribution, we further introduce a conditional prior in DCVAE.

The main contributions can be summarized as follows:
\begin{enumerate}
\item 
We introduce causal flows, an enhanced form of autoregressive flows that integrate causal structure information of ground-truth factors to improve representation expressiveness.
\item We propose the Disentangled Causal Variational Auto-Encoder (DCVAE), a novel VAE-based framework for learning causal disentangled representations.
% We theoretically prove that DCVAE satisfies disentanglement identifiability\footnote{We adopt the definition of disentanglement and model's identifiability in \citet{shen2022weakly}}.
\item To validate the effectiveness of DCVAE, we conduct extensive experiments on both synthetic and real-world datasets. The results demonstrate that DCVAE successfully generates representations with causal semantics, achieves causal disentanglement, and performs well in intervention experiments, including counterfactual image generation. Additionally, our model excels in sampling efficiency and distributional robustness in downstream tasks. 
\end{enumerate}

The remainder of the paper is organized as follows: Section \ref{2related work} reviews related works, followed by the introduction of preliminaries in section \ref{3Preliminaries}. Section \ref{sec:Causal Flows} presents our proposed causal flows, and section \ref{5the proposed model} details the overall framework of our model, DCVAE. In section \ref{6experiments}, we provide both quantitative and qualitative experiments, offering a comprehensive evaluation of our model’s performance. Finally, section \ref{7conclusion} summarizes the contributions of our approach and outlines potential directions for future research.
\section{Related Work}\label{2related work}
\subsection{Disentangled Representation Learning}
Disentangled representation learning aims to disentangle latent explanatory factors behind the observed data \citep{bengio2013representation}. It assumes that \textcolor{black}{high-dimensional} data is generated by low-dimensional, semantically meaningful generative factors, which are called ground-truth factors. Thus, a disentangled representation is characterized by changes in one dimension being solely caused by one factor of variation in the data. This pursuit, endowed with the ability to generate robust, controllable, and interpretable representations, has emerged as a challenging issue in machine learning. 

Generally, variational methods are widely employed for disentangled representation in images, utilizing an encoder-decoder framework to learn mutually independent latent factors. The variational approach employs a standard normal distribution as the prior for latent variables and then uses the variational posterior to approximate the unknown true posterior. By introducing new independent regularization terms into the original loss function, this framework is further extended, giving rise to various algorithms \citep{higgins2017beta, burgess2018understanding, kim2018disentangling, chen2018isolating, kumar2017variational, kim2019relevance, dupont2018learning,sonderby2016ladder}. $\beta$-VAE \citep{higgins2017beta} imposed a greater weight ($\beta > 1$) on the KL divergence between the variational posterior and prior, \textcolor{black}{thereby employing} a modified version of the VAE objective. In $\beta$-TCVAE \citep{chen2018isolating}, a total correlation penalty was incorporated into the objective to encourage the model to identify statistically independent factors in the data distribution. DIP-VAE \citep{kumar2017variational} introduced a regularizer on the expectation of the approximate posterior over observed data, promoting disentanglement. Unfortunately, \citet{locatello2019challenging} showed that the unsupervised learning of disentangled representations is theoretically impossible from observations without inductive biases. Therefore, several weakly supervised or supervised methods were proposed \citep{hosoya2018group,bouchacourt2018multi,shu2019weakly,locatello2020weakly}. For instance, \citet{locatello2020weakly} explored the task of learning disentangled representations from pairs of non-i.i.d. observations, where these observations share an unknown, random subset of factors of variation. They also investigated the impact of different supervision modalities. 
Furthermore, \citet{khemakhem2020variational} introduced a conditional VAE, assuming that latent variables are conditionally independent given some additionally observed variables. The latent variable model they proposed, particularly VAE, generated a provable disentangled representation under appropriate conditions. 

However, the most popular existing methods for disentanglement fail to effectively learn disentangled representations for data with strongly correlated factors, a scenario commonly encountered in real-world situations \citep{shen2022weakly,trauble2021disentangled}.

\subsection{Causal Representation Learning}\label{sec:Causal Representation Learning}
Recognizing the limitations of the aforementioned methods in addressing general scenarios, there has been growing interest in achieving causal disentangled representations through VAE. However, research in this area remains relatively limited.

\citet{scholkopf2022causality} highlighted the importance of causal disentangled representations, providing a conceptual foundation. The disentangled causal mechanisms explored in \citet{suter2019robustly} and \citet{reddy2022causally} assume conditional independence of underlying factors given a shared confounder. This assumption imposes strict constraints on the latent structure of generative factors, excluding cases where real factors may have causal relationships with one another.
For models that do not impose causal graph constraints, supervised methods are often used. \citet{locatello2019challenging} emphasized the importance of supervised learning for disentanglement, suggesting that unsupervised methods are insufficient. Consequently, Structural Causal Models (SCMs) and supervised techniques play a critical role in constructing a latent space that causally aligns with ground-truth factors \citep{shen2022weakly, yang2021causalvae, an2023causally}. \citet{yang2021causalvae} proposed an SCM layer to model the causal generative mechanisms of data, extending iVAE \citep{khemakhem2020variational} by using a conditional prior based on ground-truth factors, operating in a fully supervised manner. Similarly, DEAR \citep{shen2022weakly} utilized an SCM to construct the prior distribution in a weakly supervised setting. \citet{an2023causally} emphasized the importance of training a disentangled decoder and introduced a supervised learning approach for VAE.
In our work, we also leverage supervision signals from concept labels. We believe this supervised approach is both feasible and applicable to real-world datasets. For example, in the case of human face datasets, features such as gender and smiling status can be easily extracted from raw data.

Other models have also imposed structure on the latent space of VAEs. For example, Graph VAE \citep{he2018variational} applied a chained structure and imposed a structural causal model (SCM) on the VAE’s latent space. However, the main goal of Graph VAE is to enhance VAE’s expressive power, rather than specifically focusing on disentangling latent causal factors. Several other methods have been proposed for learning causal disentangled representations, extending beyond generative models like VAE \citep{suter2019robustly, scholkopf2022causality, kocaoglu2017causalgan}. However, unlike these methods, our model introduces causal flows within VAE, resulting in a significantly different model structure.

Relatively speaking, our proposed model is designed to handle general scenarios where generative factors exhibit more complex causal relationships. Unlike SCM-based methods, we utilize the strengths of flow models to achieve disentanglement. This approach not only leverages the flexible fitting capacity of flow models but also benefits from the causal insights offered by SCMs. In other words, our model integrates causal information while enhancing the expressive power of VAE’s inference process. Importantly, our primary focus is on representation learning, with an emphasis on the disentangled representations generated by the encoder.
To the best of our knowledge, our model is the first VAE-based framework to achieve causal disentanglement without relying on complex SCMs. Furthermore, it successfully learns causal disentangled representations without imposing constraints on the causal graph of generative factors.
\section{Preliminaries}\label{3Preliminaries}
\subsection{Variational Auto-Encoder}
Let $\{\mathbf{x}^{(j)}\}_{j=1}^N$ denote i.i.d training data, $\mathbf{x}\in \mathbb{R}^n$ be the observed variables and $\mathbf{z}\in \mathbb{R}^d$ be the latent variables. The dataset $\mathcal{X}$ has an empirical data distribution denoted as $q_{\mathcal{X}}$. The \emph{generative model} defined over $\mathbf{x}$ and $\mathbf{z}$ is $p_{\boldsymbol{{\boldsymbol{\theta}}}}(\mathbf{x},\mathbf{z})=p(\mathbf{z})p_{\boldsymbol{\theta}}(\mathbf{x}|\mathbf{z})$, where ${\boldsymbol{\theta}}$ is the parameter of the \emph{decoder}. Typically, $p(\mathbf{z})={\mathcal N}({\boldsymbol{0}},{\mathbf I})$, $p_{\boldsymbol{\theta}}(\mathbf{x}|\mathbf{z})={\mathcal N}(f_{\boldsymbol{\theta}}(\mathbf{z}),\sigma^2{\mathbf I})$, where $f_{\boldsymbol{\theta}}(\mathbf{z})$ is a neural network. The marginal likelihood $p_{\boldsymbol{\theta}}(\mathbf{x})=\int p_{\boldsymbol{\theta}}(\mathbf{x},\mathbf{z})d\mathbf{z}$ is intractable to maximize. Therefore, VAE \citep{kingma2013auto} introduces a parametric \emph{encoder} $q_{\textcolor{black}{\boldsymbol{\eta}}}(\mathbf{z}|\mathbf{x})={\mathcal N}(\mu_{\textcolor{black}{\boldsymbol{\eta}}}(\mathbf{x}),{\text{diag}}(\sigma_{\textcolor{black}{\boldsymbol{\eta}}}^2(\mathbf{x}))$, also called an \emph{inference model}, to obtain the variational lower bound on the marginal log-likelihood, i.e., the Evidence Lower Bound (ELBO):
\begin{align}
\label{VAE objective function}
    %\mathcal{L}(\mathbf{x},{\textcolor{black}{\boldsymbol{\eta}}},{\boldsymbol{\theta}})
    \rm{ELBO}(\textcolor{black}{\boldsymbol{\eta}},\boldsymbol{\theta})&=\mathbb{E}_{q_\mathcal{X}}\left[{\rm{log}}\,p_{\boldsymbol{\theta}}(\mathbf{x})-D_{{\rm{KL}}}(q_{\textcolor{black}{\boldsymbol{\eta}}}(\mathbf{z}|\mathbf{x}) \Vert p_{\boldsymbol{\theta}}(\mathbf{z}|\mathbf{x}))\right]\nonumber\\
    &=\mathbb{E}_{q_\mathcal{X}}\left[\mathbb{E}_{q_{\textcolor{black}{\boldsymbol{\eta}}}(\mathbf{z}|\mathbf{x})}({\rm{log}} \,p_{\boldsymbol{\theta}}(\mathbf{x},\mathbf{z})-{\rm{log}} \,q_{\textcolor{black}{\boldsymbol{\eta}}}(\mathbf{z}|\mathbf{x}))\right]\nonumber \\
    &=\mathbb{E}_{q_\mathcal{X}}\left[\mathbb{E}_{q_{\textcolor{black}{\boldsymbol{\eta}}}(\mathbf{z}|\mathbf{x})}{\rm{log}} \,p_{\boldsymbol{\theta}}(\mathbf{x}|\mathbf{z})-D_{{\rm{KL}}}(q_{\textcolor{black}{\boldsymbol{\eta}}}(\mathbf{z}|\mathbf{x}) \Vert p(\mathbf{z}))\right]
\end{align}
\textcolor{black}{where ${\textcolor{black}{\boldsymbol{\eta}}}$ is the parameter of the \emph{encoder}}. As can be seen from Eq. (\ref{VAE objective function}), maximizing $\rm{ELBO}(\textcolor{black}{\boldsymbol{\eta}},\boldsymbol{\theta})$
% $\mathcal{L}(\mathbf{x},{\textcolor{black}{\boldsymbol{\eta}}},{\boldsymbol{\theta}})$ 
will simultaneously maximize ${\rm{log}}\,p_{\boldsymbol{\theta}}(\mathbf{x})$ and minimize KL divergence $D_{{\rm{KL}}}(q_{\textcolor{black}{\boldsymbol{\eta}}}(\mathbf{z}|\mathbf{x}) \Vert p_{\boldsymbol{\theta}}(\mathbf{z}|\mathbf{x}))\geq 0$. Therefore, we wish $q_{\textcolor{black}{\boldsymbol{\eta}}}(\mathbf{z}|\mathbf{x})$ to be flexible enough to match the true posterior $p_{\boldsymbol{\theta}}(\mathbf{z}|\mathbf{x})$. At the same time, based on the third line of Eq. (\ref{VAE objective function}), which is often used as \textcolor{black}{the} objective function of VAE, we require that $q_{\textcolor{black}{\boldsymbol{\eta}}}(\mathbf{z}|\mathbf{x})$ is efficiently computable, differentiable, and sampled from.

\subsection{Autoregressive Normalizing Flows}\label{Autoregressive Normalizing Flows}
\emph{Normalizing flows} \citep{rezende2015variational} are effective solutions to the issues mentioned above. The flows construct flexible posterior distribution through expressing $q_{\textcolor{black}{\boldsymbol{\eta}}}(\mathbf{z}|\mathbf{x})$ as an expressive invertible and differentiable mapping $\boldsymbol{g}$ of a random variable with a relatively simple distribution, such as an isotropic normal. Typically, $\boldsymbol{g}$ is obtained by composing a sequence of invertible and differentiable transformations $\boldsymbol{g}^{(1)}, \boldsymbol{g}^{(2)},\ldots, \boldsymbol{g}^{(K)}$, i.e., $\boldsymbol{g}=\boldsymbol{g}^{(K)}\circ\cdots\circ\boldsymbol{g}^{(1)}, \boldsymbol{g}^{(k)}:\mathbb{R}^{d+n} \rightarrow \mathbb{R}^d, \forall k=1\ldots K$. If we define the initial random variable (the output of encoder) as $\mathbf{z}^{(0)}$ and the final output random variable as $\mathbf{z}^{(K)}$, then $\mathbf{z}^{(k)}=\boldsymbol{g}^{(k)}(\mathbf{z}^{(k-1)},\mathbf{x}), \forall k$. In this case, we can use $\boldsymbol{g}$ to obtain the conditional probability density function of $\mathbf{z}^{(K)}$ by applying the general probability-transformation formula \citep{papamakarios2021normalizing}:
\begin{equation}
\label{probability-transformation formula}
q_{{\textcolor{black}{\boldsymbol{\eta}}}}(\mathbf{z}^{(K)}|\mathbf{x})=q_{\textcolor{black}{\boldsymbol{\eta}}}(\mathbf{z}^{(0)}|\mathbf{x})\left|{\rm{det}}\, J_{\boldsymbol{g}(\mathbf{z}^{(0)},\mathbf{x})}\right|
\end{equation}
where ${\rm{det}}\, J_{\boldsymbol{g}(\mathbf{z}^{(0)},\mathbf{x})}$ is the Jacobian determinant of $\boldsymbol{g}$ with respect to $\mathbf{z}^{(0)}$.

\emph{Autoregressive flows} are one of the most popular normalizing flows \citep{huang2018neural,papamakarios2021normalizing,kingma2016improved}. By carefully designing the function $\boldsymbol{g}$, the Jacobian matrix in Eq. (\ref{probability-transformation formula}) becomes a lower triangular matrix.
%so the determinant can be computed in linear time. 
%Here we don't distinguish between $\boldsymbol{g}$ and $\boldsymbol{g}^{(K)}$, because if $\boldsymbol{g}^{(K)}$ has the following functional form, then $\boldsymbol{g}$ will also have the same form.
For illustration, we will only use a single-step flow with notation $\boldsymbol{g}$. Multi-layer flows are simply the composition of the function represented by a single-step flow, as mentioned earlier. And we will denote the input to the function $\boldsymbol{g}$ as $\mathbf{z}$ and its output as $\widetilde{\mathbf{z}}$. In the autoregressive flows, $\boldsymbol{g}$ has the following form:
\begin{align}
\label{autoregressive flows}
&\mathbf{\widetilde{z}}=\boldsymbol{g}(\mathbf{z},\mathbf{x})=\left[g_1(\mathbf{z}_1;\boldsymbol{h}_1 )\ldots g_d(\mathbf{z}_d;\boldsymbol{h}_d )  \right]^{\rm T}\\
&{\rm{where}} \quad \boldsymbol{h}_i=\boldsymbol{c}_i(\mathbf{\widetilde{z}}_{<i},\mathbf{x})\nonumber
\end{align}
where $g_i$, an invertible function of input $\mathbf{z}_i$, is termed as a \textbf{transformer}. Here $\mathbf{z}_i$ stands for the i-th element of vector $\mathbf{z}$, and $\boldsymbol{c}_i$ is the i-th \textbf{conditioner}, a function of the first $i-1$ elements of $\widetilde{\mathbf{z}}$, which determines part of parameters of the transformer $g_i$. We use neural networks to fit $\boldsymbol{c}$. 
\section{Causal Flows}\label{sec:Causal Flows}
In this section, we propose an extension to the autoregressive flows by incorporating an adjacency matrix $A$. The extended flows still involve functions with tractable Jacobian determinants. 

In autoregressive flows, 
%according to equations (\ref{autoregressive flows}) and the third section in \citet{khemakhem2021causal},
causal order is established among variables $\mathbf{\widetilde{z}}_1,\cdots, \mathbf{\widetilde{z}}_d$. Given the causal graph of the variables $\mathbf{\widetilde{z}}_1,\cdots, \mathbf{\widetilde{z}}_d$, let $A \in \mathbb{R}^{d\times d}$ denote its corresponding binary adjacency matrix, $A_{i,:}$ is the row vector of $A$ and $A_{i,j}$ is nonzero only if $\mathbf{{\widetilde z}}_j$ is the parent node of $\mathbf{{\widetilde z}}_i$, then $A$ corresponding to the causal order in autoregressive flows is a full lower-triangular matrix. The conditioner can be written in the form of \textcolor{black}{$\boldsymbol{c}_i(\mathbf{\widetilde{z}}\circ A_{i,:},\mathbf{x})$}, where $\circ$ is the element-wise product. 
%The binary adjacency matrix corresponding to $A$ is $\mathbf{I}_A = \mathbf{I}(A \neq 0)$, here $\mathbf{I}$ is the element-wise indicator function. 
If we utilize prior knowledge about the true causal structure among variables, i.e., if a certain causal structure
%$ (A$ or $\mathbf{I}_A)$ 
among variables is known, then $A$ is still a lower triangular matrix, but some of its entries are set to 0.
%according to the underlying causal graph
We can integrate such $A$ into the conditioner to include causal structure information in the model, which is also denoted as \textcolor{black}{$\boldsymbol{c}_i(\mathbf{\widetilde{z}}\circ A_{i,:},\mathbf{x})$}. We will refer to it as the $\textbf{causal conditioner}$ in the following.

We define autoregressive flows that use the causal conditioner as \textbf{Causal Flows}. %Besides the conditioner, we also need to specify the transformer to construct flows. 
The transformer can be any invertible function, and we focus on affine transformer, which is one of the simplest transformers. Therefore, causal flows $\boldsymbol{g}$ can be formulated as follows:
\begin{equation}
    \widetilde{\mathbf{z}}_i = g_i(\mathbf{z}_i;\boldsymbol{h}_i) = \mathbf{z}_i \, {\rm{exp}}(s_i(\widetilde{\mathbf{z}}\circ A_{i,:},\mathbf{x}))+t_i(\widetilde{\mathbf{z}}\circ A_{i,:},\mathbf{x})
\label{flows formulation}
\end{equation}
where $\boldsymbol{s}=\left[s_1,\cdots,s_d\right]^{\rm{T}}\in \mathbb{R}^d$ and $\boldsymbol{t}=\left[t_1,\cdots,t_d\right]^{\rm{T}}\in \mathbb{R}^d$ are defined by the conditioner, i.e., $\boldsymbol{h}_i=\left\{s_i,t_i\right\}$, and $s_1$ and $t_1$ are constants. 

Given that the derivative of the transformer with respect to $\mathbf{z}_i$ is ${\rm{exp}}(s_i(\widetilde{\mathbf{z}}\circ A_{i,:}, \mathbf{x}))$ and $A$ is lower-triangular, the log absolute Jacobian determinant is:
\begin{align}
\label{Jacabian}
    {\rm{log}}\,\left|{\rm{det}}\, J_{\boldsymbol{g}(\mathbf{z},\mathbf{x})}\right| &=\sum_{i=1}^{d}{\rm{log}}\,{\rm{exp}}(s_i(\widetilde{\mathbf{z}}\circ A_{i,:},\mathbf{x}))\\
    &=\sum_{i=1}^{d}s_i(\widetilde{\mathbf{z}}\circ A_{i,:},\mathbf{x})
\end{align}
Now, we are able to derive the log probability density function of $\widetilde{\mathbf{z}}$ using the following expression:
\begin{equation}
    {\rm log}\,q_{{\textcolor{black}{\boldsymbol{\eta}}}}(\widetilde{\mathbf{z}}|\mathbf{x})={\rm log}\,q_{\textcolor{black}{\boldsymbol{\eta}}}(\mathbf{z}|\mathbf{x})-\sum_{i=1}^{d}s_i(\widetilde{\mathbf{z}}\circ A_{i,:},\mathbf{x})
\end{equation}

Causal flows will not only empower our model to incorporate causal structure information but also through the utilization of flow model characteristics, enhance the expressive capacity of the learned representations. This augmentation, consequently, enriches the information encoded in the learned representations, thereby improving the overall proficiency of our model.

It is worth noting that the computation of autoregressive flows in Eq. (\ref{autoregressive flows}) needs to be performed sequentially, meaning that $\widetilde{\mathbf{z}}_{<i}$ must be calculated before $\widetilde{\mathbf{z}}_i$. Due to the sampling requirement in VAE, this approach may not be computationally efficient. % This is particularly true when dealing with high-dimensional latent variables and multiple layers of causal flows.
However, in causal disentanglement applications of VAE, the number of factors of interest is often relatively small. Additionally, we've found that using a single layer of causal flows and lower-dimensional latent variables is enough to lead to better results, so the computational cost of the model is not significantly affected by sequential sampling.
% as the encoder and decoder architectures play a key role in determining reconstructed image quality.

\section{The Proposed model}\label{5the proposed model}
This section focuses on addressing the issue of causal disentanglement in VAE. 
First, we introduce some notations. We denote $\boldsymbol{\xi} \in \mathbb{R}^m$ as the underlying ground-truth factors of interest for data $\mathbf{x}$, with distribution $p_{\boldsymbol{\xi}}$. For each underlying factor $\boldsymbol{\xi}_i$, we denote $\boldsymbol{y}_i$ as some continuous or discrete annotated observation satisfying $\boldsymbol{\xi}_i  = {\mathbb{E}}(\boldsymbol{y}_i| \mathbf{\mathbf{x}})$, where the superscript $i$ still denotes the i-th element of each vector. Let $\mathcal{D}=\left\{(\mathbf{x}^{(j)},\boldsymbol{y}^{(j)},\mathbf{u}^{(j)})\right\}_{j=1}^{N}$ denotes a labeled dataset, where $\boldsymbol{u}^{(j)}\in \mathbb{R}^k$ is the additional observed variable. Depending on the context, the variable $\mathbf{u}$ can take on various meanings, such as serving as the time index in a time series, a class label, or another variable that is observed concurrently \citep{hyvarinen2016unsupervised}.
% We get $\boldsymbol{\xi}_i = {\mathbb{E}}(\boldsymbol{y}_i | \mathbf{\mathbf{x}}, \mathbf{u})$, where $i=1,\cdots,m$. This is because if $\mathbf{u}$ is ground-truth factor $\boldsymbol{y}$, it is obviously true, otherwise, $\boldsymbol{\xi}_i = {\mathbb{E}}(\boldsymbol{y}_i | \mathbf{\mathbf{x}}, \mathbf{u})={\mathbb{E}}(\boldsymbol{y}_i | \mathbf{\mathbf{x}})$. 
\textcolor{black}{When $\mathbf{u}$ is the ground-truth factor $\boldsymbol{y}$, $\boldsymbol{\xi}_i = {\mathbb{E}}(\boldsymbol{y}_i | \mathbf{x}, \mathbf{u})$, which is obviously true. Otherwise, when $\mathbf{u}$ is not the ground-truth factor $\boldsymbol{y}$, i.e., when $\mathbf{y}$ and $\mathbf{u}$ are entirely unrelated, $\boldsymbol{\xi}_i = {\mathbb{E}}(\boldsymbol{y}_i | \mathbf{x}) = {\mathbb{E}}(\boldsymbol{y}_i | \mathbf{x}, \mathbf{u})$. Thus, we obtain $\boldsymbol{\xi}_i = {\mathbb{E}}(\boldsymbol{y}_i | \mathbf{x}, \mathbf{u})$, where $i = 1, \cdots, m$.}
% It should be noted that our subsequent discussion will center on the encoder-causal flows-decoder architecture in VAEs, and 
We will view the encoder and flows as \textcolor{black}{a} unified stochastic transformation $E$, with the learned representation $\widetilde{\mathbf{z}}$ as its final output, i.e., $\widetilde{\mathbf{z}}=E(\mathbf{x},\mathbf{u})$. Additionally, in the stochastic transformation $E(\mathbf{x},\mathbf{u})$, we use $\bar{E}(\mathbf{x},\mathbf{u})$ to denote its deterministic part, i.e., $\bar{E}(\mathbf{x},\mathbf{u}) = \mathbb{E}(E(\mathbf{x},\mathbf{u}) | \mathbf{x},\mathbf{u})$. 
% we use $\bar{E}(\mathbf{x})$ to denote the deterministic part of the stochastic transformation $E(\mathbf{x})$, i.e., $\bar{E}(\mathbf{x}) = \mathbb{E}(E(\mathbf{x}) | \mathbf{x})$. 

Now, we adopt the definition of causal disentanglement as follows:
\begin{definition} [Disentangled representation \citep{shen2022weakly}]   \label{definition}
Considering the underlying factor $\boldsymbol{\xi} \in \mathbb{R}^m$ of data $\mathbf{x}$, $E$ is said to learn a disentangled representation with respect to $\boldsymbol{\xi}$ if there exists a one-to-one function $r_i$ such that ${{\bar E}(\mathbf{x},\mathbf{u})}_i=r_i(\boldsymbol{\xi}_i), \forall i=1,\cdots,m$.
\end{definition}

%As noted in \citet{shen2022weakly}, 
The purpose of this definition is to guarantee some degree of alignment between the latent variable $E(\mathbf{x})$ and the underlying factor $\boldsymbol{\xi}$ in the model. 

\begin{figure}[t]
\begin{center}
\includegraphics[width=0.8\linewidth]{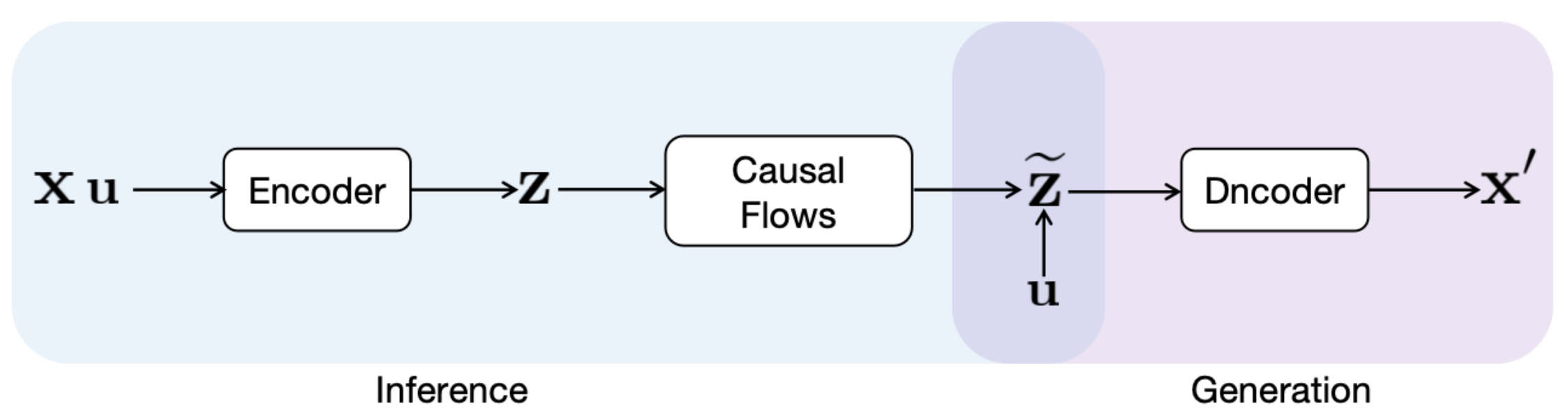}
\caption{Model structure of DCVAE.}
\label{fig:architecture}
\end{center}
\end{figure}

%To learn a disentangled representation that meets the definition mentioned above,
We now proceed to present the full probabilistic formulation of DCVAE. The model's structure is depicted in Figure \ref{fig:architecture}.
%下标表示不同样本，上标表示
The conditional generative model is defined as follows:
\begin{align}
     p_{\boldsymbol{{\theta}}}(\mathbf{x},\widetilde{\mathbf{z}}|\mathbf{u})&=p_{\mathbf{f}}(\mathbf{x}|\widetilde{\mathbf{z}}, \mathbf{u})p_{\mathbf{T},\boldsymbol{\lambda}}(\widetilde{\mathbf{z}}|\mathbf{u})\label{generative model}\\
     p_{\mathbf{f}}(\mathbf{x}|\widetilde{\mathbf{z}},\mathbf{u})&=p_{\mathbf{f}}(\mathbf{x}|\widetilde{\mathbf{z}})= p_{\boldsymbol{\zeta}}(\mathbf{x}-\mathbf{f}(\widetilde{\mathbf{z}}))\label{generative}
\end{align}
\textcolor{black}{with}
\begin{align}
p_{\mathbf{T},\boldsymbol{\lambda}}(\widetilde{\mathbf{z}}|\mathbf{u})=
\begin{cases}
        \frac{Q(\widetilde{\mathbf{z}}_{\leq m})e^{<\mathbf{T}(\widetilde{\mathbf{z}}_{\leq m}),\boldsymbol{\lambda}(\mathbf{u})>}}{Z(\mathbf{u})}\\
        \mathcal{N}(\boldsymbol{0}_{(d-m)\times 1},\mathbf{I}_{(d-m)\times(d-m)})
\end{cases}
\label{prior}
\end{align}
where $\boldsymbol{\theta}=(\mathbf{f},\mathbf{T},\boldsymbol{\lambda})\in \boldsymbol\Theta$ are model parameters, \textcolor{black}{$p_{\mathbf{T},\boldsymbol{\lambda}}(\widetilde{\mathbf{z}}|\mathbf{u}) > 0$ is almost surely $\forall(\widetilde{\mathbf{z}},\mathbf{u}) \in \widetilde{\mathbf{Z}}\times\mathbf{U}$ with $\widetilde{\mathbf{Z}}$ and $\mathbf{U}$ to be compact Hausdorff spaces.} Eq. (\ref{generative model}) describes the process of generating $\mathbf{x}$ from $\widetilde{\mathbf{z}}$. Eq. (\ref{generative}) indicates that $\mathbf{x}=\mathbf{f}(\widetilde{\mathbf{z}})+\boldsymbol\zeta$, where $p_{\boldsymbol \zeta}({\boldsymbol \zeta})=\mathcal{N}(\boldsymbol{0},{\mathbf{I}})$ and the decoder $\mathbf{f}(\widetilde{\mathbf{z}})$ is assumed to be an invertible function approximated by a neural network. \textcolor{black}{In Eq. (\ref{prior}), the first equation is used
%, we use the exponential conditional distribution \citep{pacchiardi2022score}
for the first $m$ dimensions of $\widetilde{\mathbf{z}}$ while the second one serves for the remaining $d-m$ dimensions to capture other non-interest factors for generation};
% Based on the properties of the exponential function we use (Theorem \ref{theorem2} in \ref{Appendix:theorem_proof}), we need to add the conditions .
 $\mathbf{T}: {\mathbb{R}}^d\rightarrow {\mathbb{R}}^{d\times l}$ is the sufficient statistic, ${\boldsymbol{\lambda}}:{\mathbb{R}}^k\rightarrow{\mathbb{R}}^{d\times l}$ is the corresponding parameter, $Q:{\mathbb{R}}^d\rightarrow {\mathbb{R}}$ is the base measure, $Z(\mathbf{u})$ is the normalizing constant and $<\cdot, \cdot>$ denotes the dot product.  If $d=m$, we will only use the conditional prior in the first line of (\ref{prior}). When causal relationships exist among the generative factors of data $\mathbf{x}$, indicating their non-mutual independence, incorporating information $\mathbf{u}$ alters the prior distribution from a factorial distribution to a distribution that better matches the real-world situation.

We define the inference model that utilizes causal flows as follows:
\begin{align}
    q_{{\textcolor{black}{\boldsymbol{\eta}}}}(\mathbf{z}|\mathbf{x},\mathbf{u})&=q_{{\boldsymbol{\epsilon}}}(\mathbf{z}-{\boldsymbol{\phi}}(\mathbf{x},\mathbf{u}))\label{encoder}\\
    \mathbf{z}&\sim q_{{\textcolor{black}{\boldsymbol{\eta}}}}(\mathbf{z}|\mathbf{x},\mathbf{u})\label{z}\\
    \widetilde{\mathbf{z}}&=\boldsymbol{g}(\mathbf{z},\mathbf{x})\label{zK}\\
    q_{{\textcolor{black}{\boldsymbol{\eta}}},{\boldsymbol{\gamma}}}(\widetilde{\mathbf{z}}|\mathbf{x},\mathbf{u})&=q_{\textcolor{black}{\boldsymbol{\eta}}}(\mathbf{z}|\mathbf{x},\mathbf{u})\prod \limits_{i=1}^{d}{\rm exp}(-s_i(\widetilde{\mathbf{z}}\circ A_{i,:},\mathbf{x}))\label{zK_function}
\end{align}
where ${\boldsymbol{\gamma}}=(\boldsymbol{s},\boldsymbol{t},A)\in \boldsymbol\Gamma$ denotes parameters of causal flows. Eq. (\ref{encoder}) indicates that $\mathbf{z}={\boldsymbol{\phi}}(\mathbf{x},\mathbf{u})+\boldsymbol\epsilon$, where the probability density of ${\boldsymbol\epsilon}$ is $q_{\boldsymbol\epsilon}(\boldsymbol\epsilon)=\mathcal{N}(\boldsymbol{0},{\mathbf{I}})$ and ${\boldsymbol{\phi}}(\mathbf{x},\mathbf{u})$ denotes the encoder. Eq. (\ref{z}) and (\ref{zK}) describe the process of transforming the original encoder output $\mathbf{z}$ into the final latent variable representation $\widetilde{\mathbf{z}}$ by using causal flows. Eventually, the posterior distribution obtained by the inference model is represented by Eq. (\ref{zK_function}). So the parameters of stochastic transformation $E(\mathbf{x},\mathbf{u})$ are $\textcolor{black}{\boldsymbol{\eta}}$ and ${\boldsymbol{\gamma}}$\textcolor{black}{, which we denote as $E_{\textcolor{black}{\boldsymbol{\eta}},\boldsymbol{\gamma}}(\mathbf{x},\mathbf{u})$}.

Now the dataset $\mathcal{X}$ has an empirical data distribution denoted by $q_{\mathcal{X}}(\mathbf{x},\mathbf{u})$. Our goal is to maximize the variational lower bound on the marginal likelihood $p_{\boldsymbol{\theta}}(\mathbf{x}|\mathbf{u})$. The labels $\boldsymbol{y}$ represent the ground-truth latent factors. We introduce a regularization term in the objective function to encourage consistency between $\boldsymbol{\xi}$ and $E(\mathbf{x,u})$. The loss function of \textbf{DCVAE} is formulated as follows:
\begin{align}
    \mathcal{L}(\textcolor{black}{\boldsymbol{\eta}},\boldsymbol{\gamma},\boldsymbol{\theta})=&-{\rm{ELBO}}(\boldsymbol{\phi,\gamma,\theta})+\beta_{sup}\mathcal{L}_{sup}(\textcolor{black}{\boldsymbol{\eta}},\boldsymbol{\gamma})\nonumber\\
        =&-\mathbb{E}_{q_\mathcal{X}}[\mathbb{E}_{q_{{\textcolor{black}{\boldsymbol{\eta}}},{\boldsymbol{\gamma}}}(\widetilde{\mathbf{z}}|\mathbf{x},\mathbf{u})}{\rm{log}} \,p_{\mathbf{f}}(\mathbf{x}|\widetilde{\mathbf{z}},\mathbf{u})\nonumber\\
        &-D_{{\rm{KL}}}(q_{{\textcolor{black}{\boldsymbol{\eta}}},{\boldsymbol{\gamma}}}(\widetilde{\mathbf{z}}|\mathbf{x},\mathbf{u})\Vert p_{\mathbf{T},\boldsymbol{\lambda}}(\widetilde{\mathbf{z}}|\mathbf{u}))] \nonumber\\
        &+\quad \beta_{sup}{\mathbb E}_{(\mathbf{x},\boldsymbol{y},\mathbf{u})}\left[l_{sup}(\textcolor{black}{\boldsymbol{\eta}},\boldsymbol{\gamma})\right]\label{loss}
\end{align}
where $\beta_{sup}>0$ is a hyperparameter, $l_{sup}(\textcolor{black}{\boldsymbol{\eta}},\boldsymbol{\gamma})= \sum_{i=1}^m(\boldsymbol{y}_i-\textcolor{black}{{\bar E}_{\textcolor{black}{\boldsymbol{\eta}},\boldsymbol{\gamma}}(\mathbf{x},\mathbf{u})}_i)^2$ is the Mean Squared Error if ${\boldsymbol y}_i$ is the continuous observation, and $l_{sup}(\textcolor{black}{\boldsymbol{\eta}},\boldsymbol{\gamma})=\sum_{i=1}^{m}-\boldsymbol{y}_i{\rm log}\,\sigma(\textcolor{black}{{\bar E}_{\textcolor{black}{\boldsymbol{\eta}},\boldsymbol{\gamma}}(\mathbf{x},\mathbf{u})}_i)-(1-\boldsymbol{y}_i){\rm log}\,(1-\sigma(\textcolor{black}{{\bar E}_{\textcolor{black}{\boldsymbol{\eta}},\boldsymbol{\gamma}}(\mathbf{x},\mathbf{u})}_i))$ is the cross-entropy loss if ${\boldsymbol y}_i$ is the binary label. The loss term $\mathcal{L}_{sup}$ aligns the factor of interest $\boldsymbol{\xi}\in \mathbb{R}^m$ with the first $m$ dimensions of the latent variable $\widetilde{\mathbf{z}}$, in order to satisfy the Definition \ref{definition} \citep{locatello2020weakly, shen2022weakly}.

As demonstrated in \citep{shen2022weakly}, even under the supervision of latent variables, previously employed disentangled representation learning methods, based on the assumption of independent priors, fail to achieve true disentanglement when causal relationships exist among the latent factors of interest (i.e., the disentanglement identifiability discussed earlier in Definition \ref{definition}). To address the identifiability issue raised in \citep{shen2022weakly}, we incorporate additional information, denoted as $\mathbf{u}$.
We utilize this additional information in two ways. First, we propose integrating it into conditional priors to regularize the learned posterior of $\mathbf{z}$. This regularization allows us to focus on priors that are more aligned with the true generating factors of interest. Moreover, the use of these conditional priors ensures that the learned representation is indeed disentangled, as defined in Definition \ref{definition}, thus addressing the identifiability concerns, with the proof being similar to that in \citep{shen2022weakly}. 
Second, in real-world experiments, if we take $\mathbf{u}$ as concept labels, it will function as regularization terms, thereby constraining the latent variable information, i.e., as $\boldsymbol{y}$.
% In this case, the overall supervision signal used by the model consists only of the labels for the generating factors. 
As highlighted in section \ref{sec:Causal Representation Learning}, existing methods primarily focus on supervised models, which are readily implementable in practice. This highlights the practicality and real-world applicability of our approach.
\section{Experiments}\label{6experiments}
In this section, we empirically evaluate DCVAE, demonstrating that the learned representation is causally disentangled. \textcolor{black}{This capability will enable the model to perform across various tasks effectively.}
\subsection{Datasets}
\begin{figure}[tbp]
\begin{center}
\subfigure[Pendulum]{
\begin{minipage}[t]{0.26\linewidth}
\centering
\includegraphics[width=\textwidth]{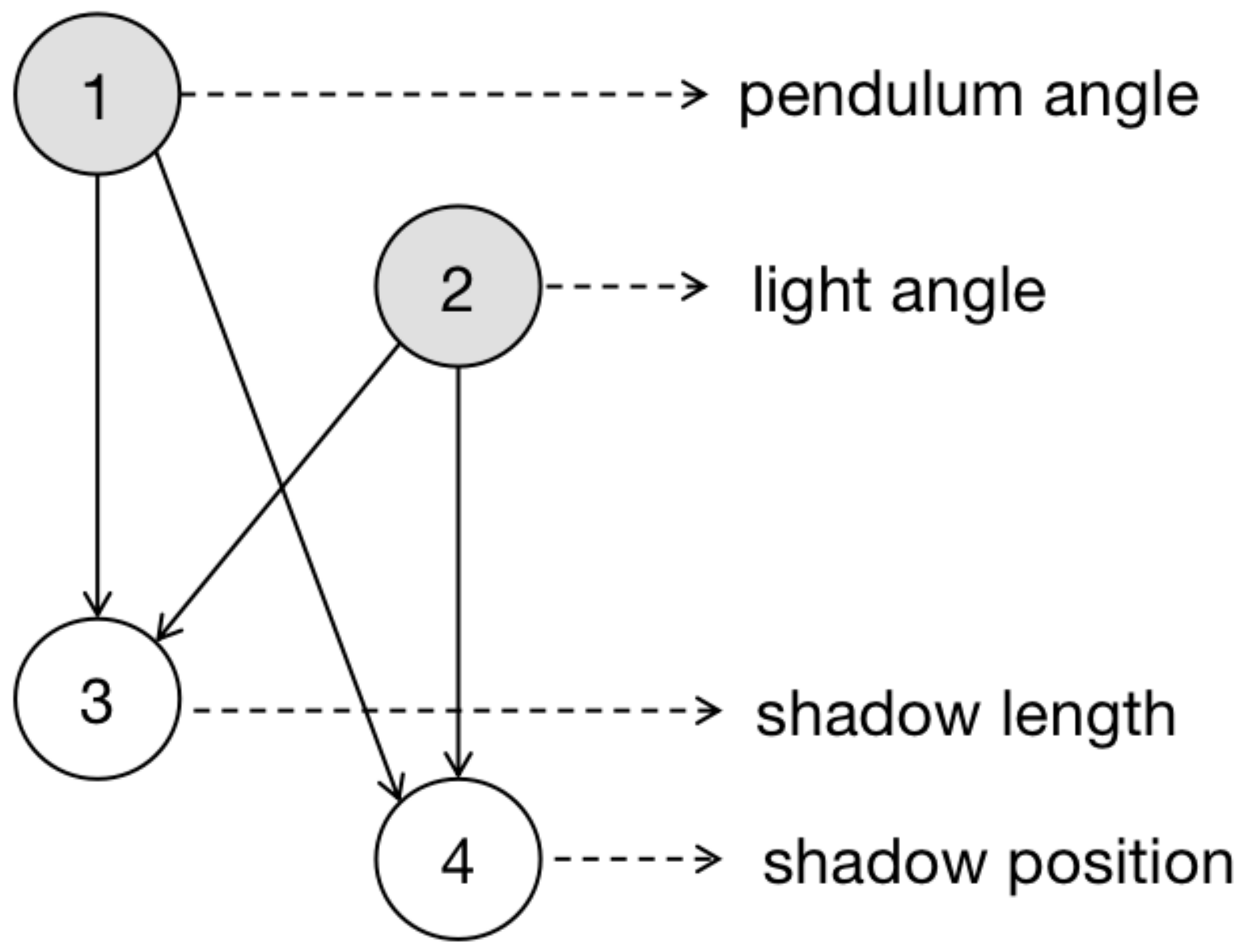}
% \caption{fig1}
\label{fig:pendulumgraph}
\end{minipage}%
}%
\subfigure[CelebA(Attractive)]{
\begin{minipage}[t]{0.3\linewidth}
\centering
\includegraphics[width=\textwidth]{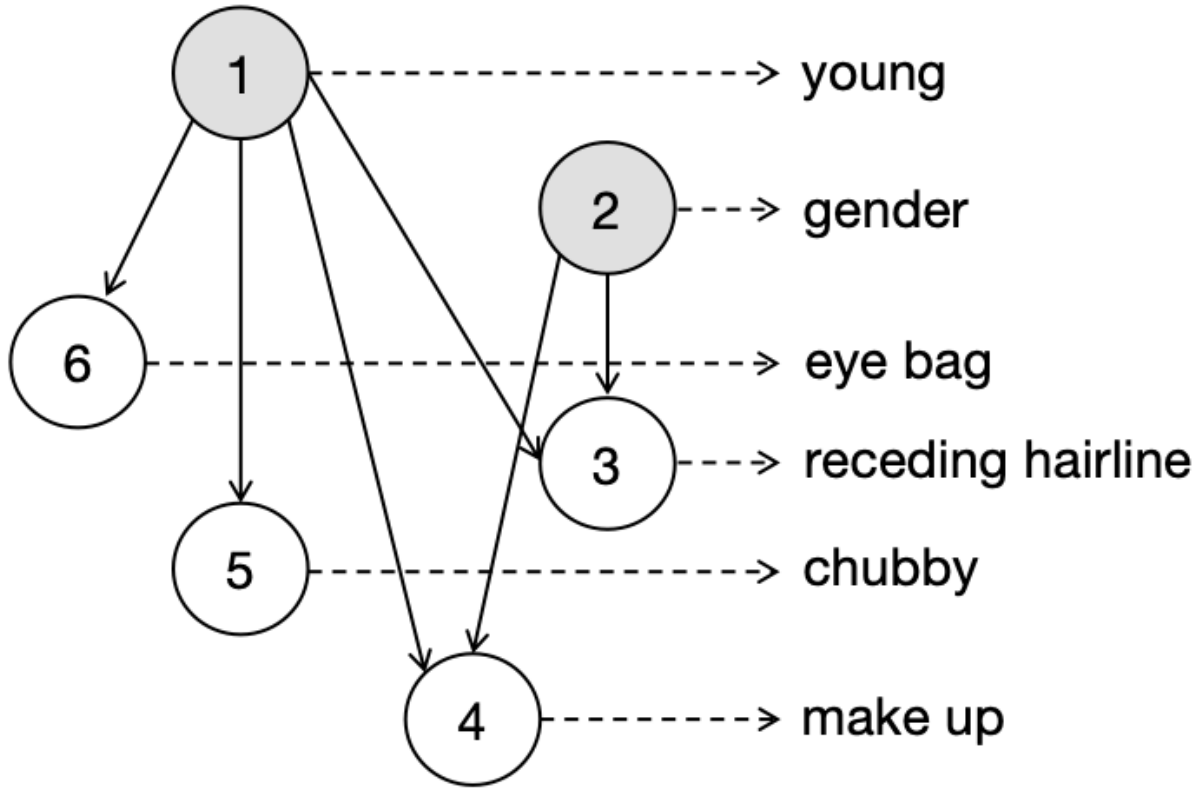}
%\caption{fig2}
\label{fig:attractivegraph}
\end{minipage}%
}%
\subfigure[CelebA(Smile)]{
\begin{minipage}[t]{0.28\linewidth}
\centering
\includegraphics[width=\textwidth]{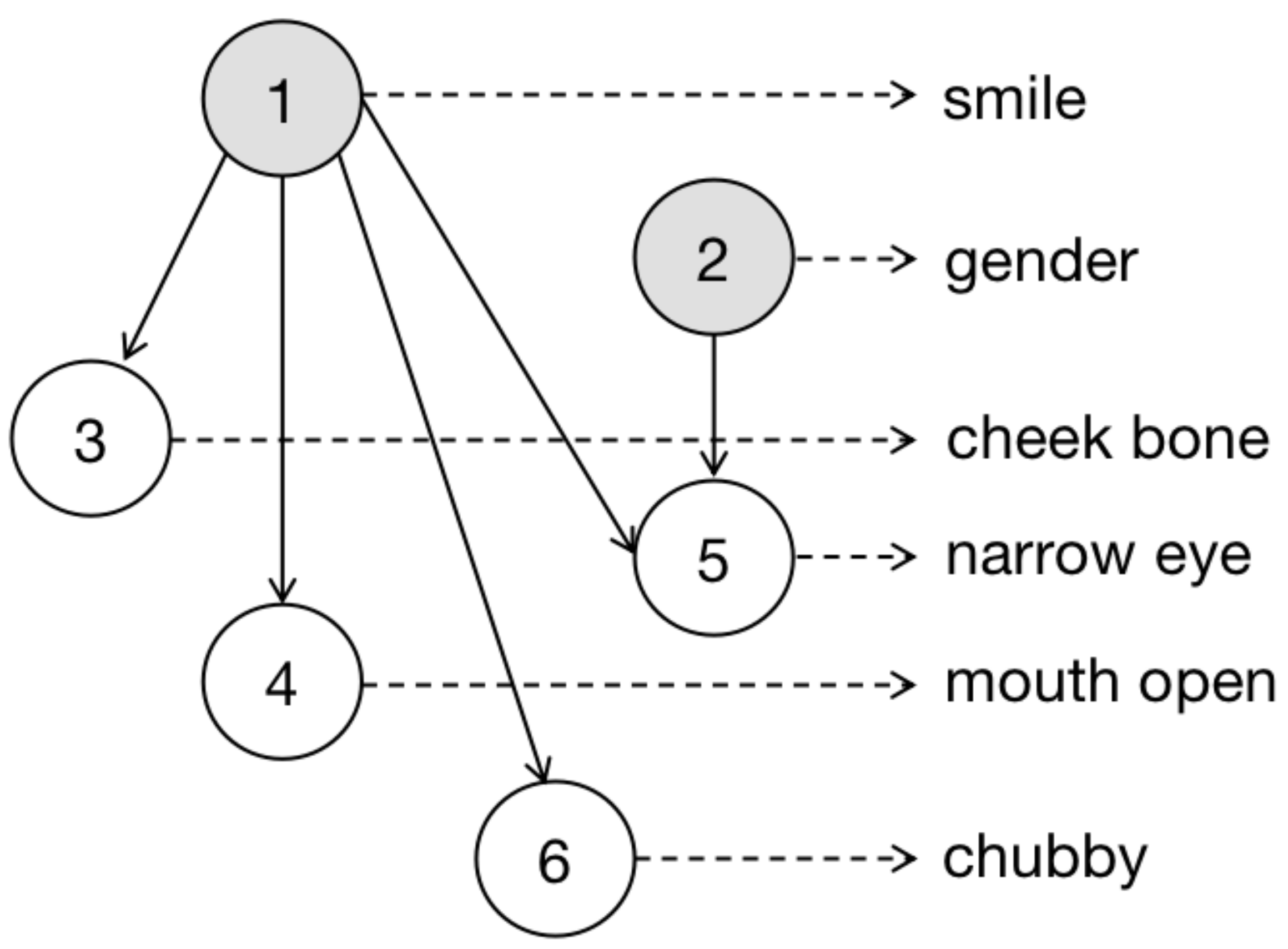}
%\caption{fig2}
\label{fig:smilegraph}
\end{minipage}
}%
\end{center}
 \caption{Causal graphs of Pendulum and CelebA. The gray circles represent the causal variables in the graphs. In Figures (a), (b), and (c), we label the underlying factors we are interested in each dataset.
}
\end{figure}

We utilize the same datasets from \citet{shen2022weakly}\textcolor{black}{,} where the underlying generative factors are causally related. The synthetic dataset is Pendulum, with four continuous factors whose causal graph of the factors is shown in Figure \ref{fig:pendulumgraph}. We generate the pendulum dataset using the synthetic simulators mentioned in \citet{yang2021causalvae}. The training and testing sets consist of 5847 and 1461 samples, respectively. The real human face dataset is CelebA \citep{liu2015deep}, with 40 discrete labels. We consider two sets of causally related factors named CelebA(Attractive) and CelebA(Smile) with causal graphs also depicted in Figure \ref{fig:attractivegraph} and \ref{fig:smilegraph}. \textcolor{black}{We set the values of features to $\left[-1, 0\right]$}. The training and testing sets consist of 162770 and 19962 samples, respectively. In the datasets, $\mathbf{x}$ represents the sample, $\boldsymbol{y}$ denotes the true label information of the generating factors, and $\mathbf{u}$ is the additional observed variable, which we have designated as label information for the experiments.
% , as detailed in section \ref{sec:Disentanglement Identifiability}.
All the images are resized to 64×64×3 resolution.
\subsection{Experimental Settings and Baselines}
We present the main settings used in our experiments. Our experiments on Pendulum utilize one NVIDIA GeForce RTX 2080ti GPU, while experiments on CelebA use one NVIDIA GeForce RTX
3080 GPU. To train DEAR, we use two NVIDIA GeForce RTX 2080ti GPUs.
\subsubsection{DCVAE} In DCVAE, regarding the setting of conditional prior of DCVAE, since it is generally difficult to directly fit the exponential family of distributions, we use a special form of exponential distribution, namely the Gaussian distribution in our experiments. For simplicity, we adopt a factorial distribution, as described in \citet{khemakhem2020variational} and \citet{yang2021causalvae}. However, unlike them, we set the $mean$ and $variance$ as learnable parameters for training, which enhances the flexibility of the prior distribution.
% further guarantees that $\widetilde{\mathbf{z}}_i$ is a causal disentangled representation, for $1\leq i \leq m$. This means that the latent variables can capture the true underlying factors successfully. 
In the implementation of the supervised loss for Pendulum, the factors' labels are resized to $\left[-1, 1\right]$ because they are continuous,  and Mean Squared Error (MSE) is employed as the loss function $\mathcal{L}_{sup}$. For CelebA, where the factors' labels are binary, we map $-$1 to 0 and utilize cross-entropy loss. 
\textcolor{black}{Since we incorporate additional information $\mathbf{u}$ relevant to the true concept labels as the supervision signal in our experiment, the encoder in DCVAE only takes $\mathbf{x}$ as input and does not use $\mathbf{u}$ to avoid information leakage.} The comprehensive details of the network architectures and hyperparameters are given in Appendix \ref{Appendix:Experimental Details}.
\subsubsection{Baseline models} We compare our method with several  state-of-the-art VAE-based models for disentanglement \citep{locatello2019challenging}, including $\beta$-VAE \citep{higgins2017beta}, $\beta$-TCVAE \citep{chen2018isolating}, DEAR \citep{shen2022weakly} and VAE \citep{kingma2013auto}.
We also investigate prior information with limited understanding of causal graph structures. Specifically, DCVAE-SP represents a DCVAE trained using the given super-graph of the true graph (i.e., Figure \ref{fig:super_graph_all_text}), rather than the full graph.
% to highlight the advantages of disentanglement techniques. 
In this context, $\beta$-VAE and $\beta$-TCVAE stand out as the most representative VAE-based disentanglement models, imposing independent constraints on the posterior and utilizing total correlation to encourage independence among aggregated posteriors, respectively. Currently, the most advanced causal disentanglement model based on VAE is DEAR. Therefore, overall, the selected comparison models are all representative. 
% Furthermore, except for DEAR, we utilize the same encoder and decoder network structures since we consider the design of DEAR's encoder and decoder to be part of its model innovation. 
We use the same conditional prior and loss term with labeled data for each of these methods as in DCVAE, except that DEAR's prior is SCM prior. 
% Furthermore, except for DEAR, we utilize the same encoder and decoder network structures since we consider the design of DEAR's encoder and decoder to be part of its model innovation. 
Furthermore, apart from models that specifically choose the architecture of  encoder and decoder, we employ identical encoder and decoder structures for the baselines.
The implementations of $\beta$-TCVAE and DEAR are each associated with publicly available source codes, which can be found at \url{https://github.com/AntixK/PyTorch-VAE} and \url{http://jmlr.org/papers/v23/21-0080.html}. 
\begin{figure}[b]
\begin{center}
\subfigure[Traverse of DCVAE]{
\begin{minipage}[b]{0.5035\linewidth}
\centering
\includegraphics[width=\textwidth]{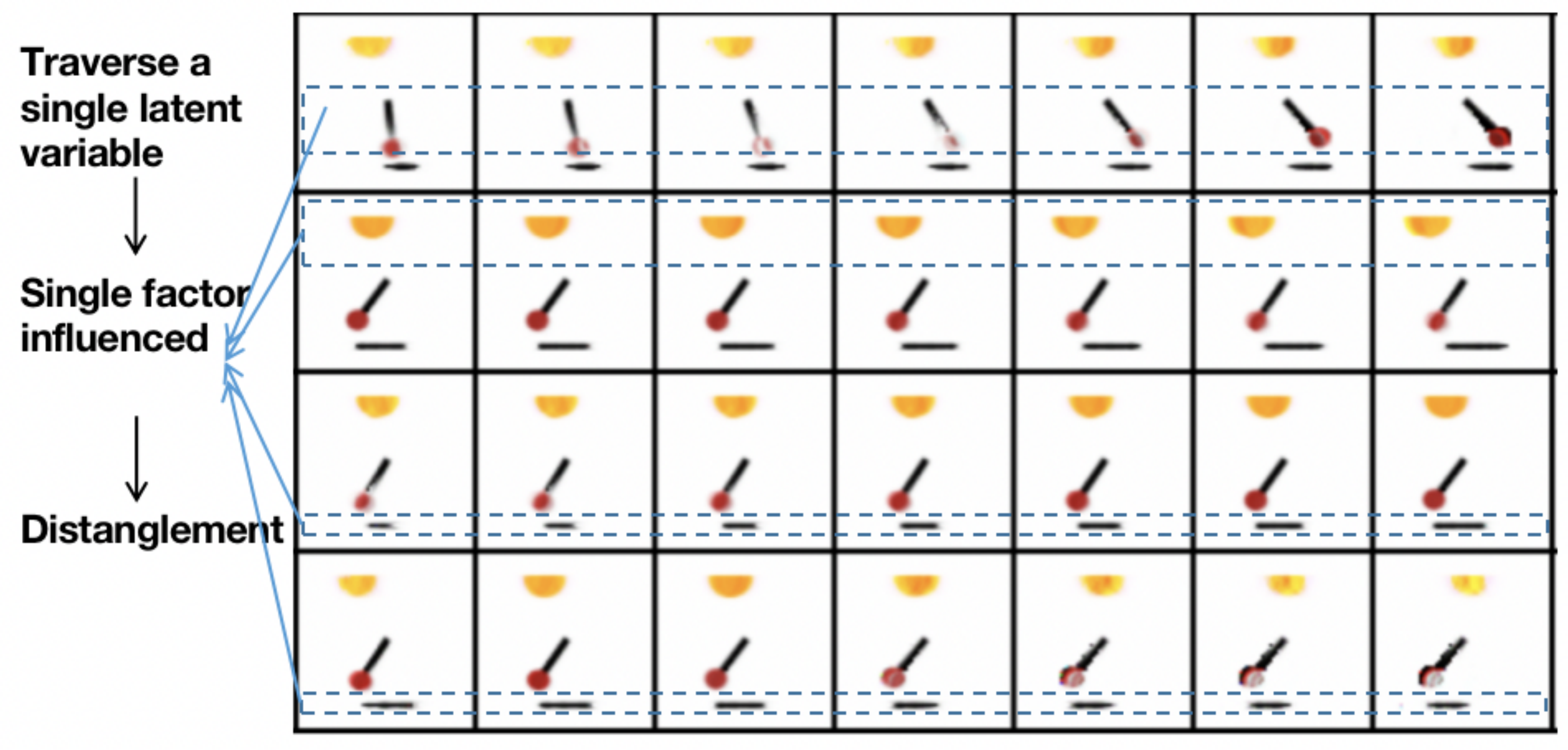}
% \caption{fig1}
\label{fig:cfvaependulumtraverse}
\end{minipage}%
}%
\subfigure[Traverse of DEAR]{
\begin{minipage}[b]{0.4978\linewidth}
\centering
\includegraphics[width=\textwidth]{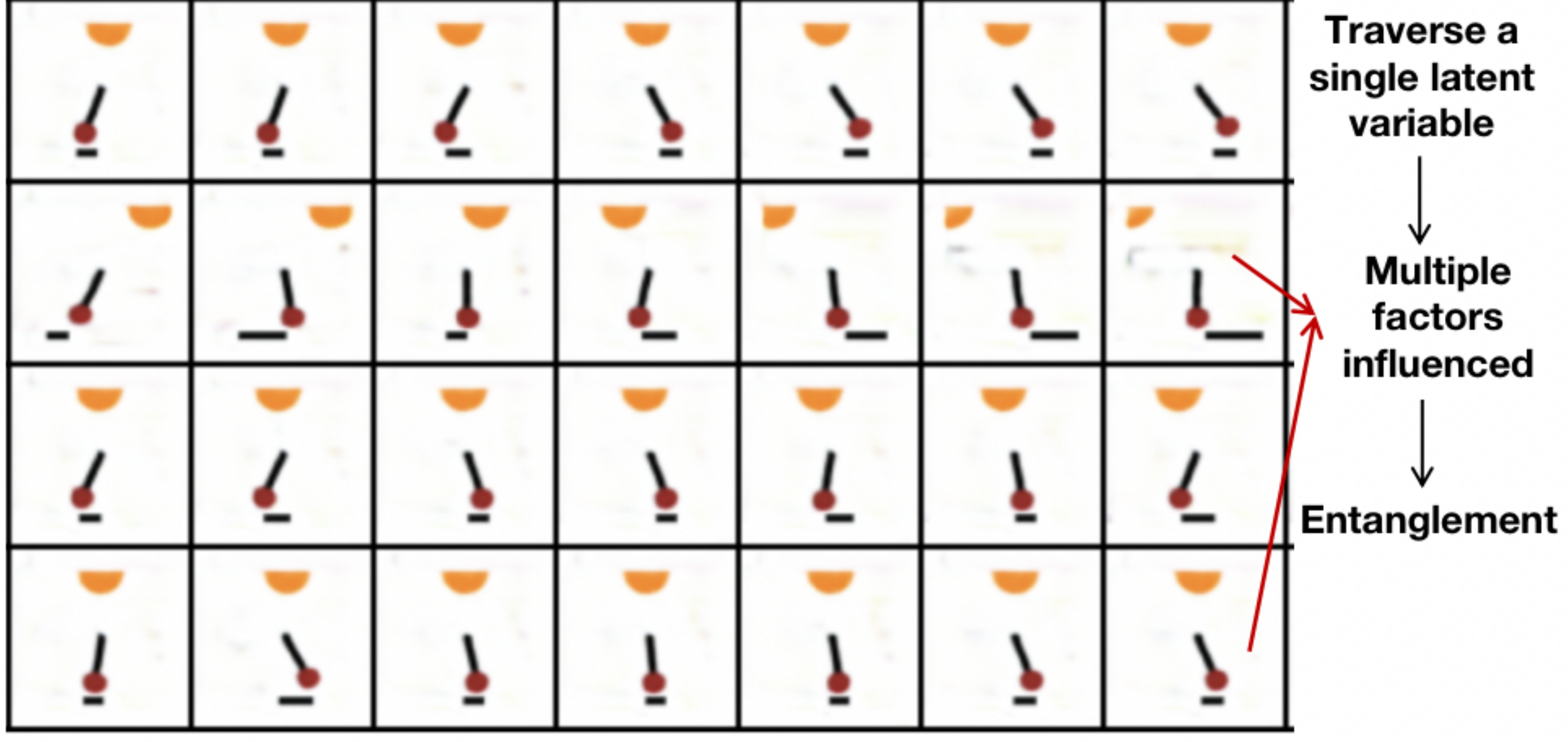}
%\caption{fig2}
\label{fig:dearpendulumtraverse}
\end{minipage}%
}%
\end{center}
\caption{Results of traverse experiments on Pendulum. Each row corresponds to a variable that we traverse on, specifically, pendulum angle, light angle, shadow length, and shadow position.}
\label{fig:pendulumtraverse}
\end{figure}
\begin{figure}[b]
\begin{center}
\subfigure[Traverse of DCVAE]{
\begin{minipage}[b]{0.52\linewidth}
\centering
\includegraphics[width=\textwidth]{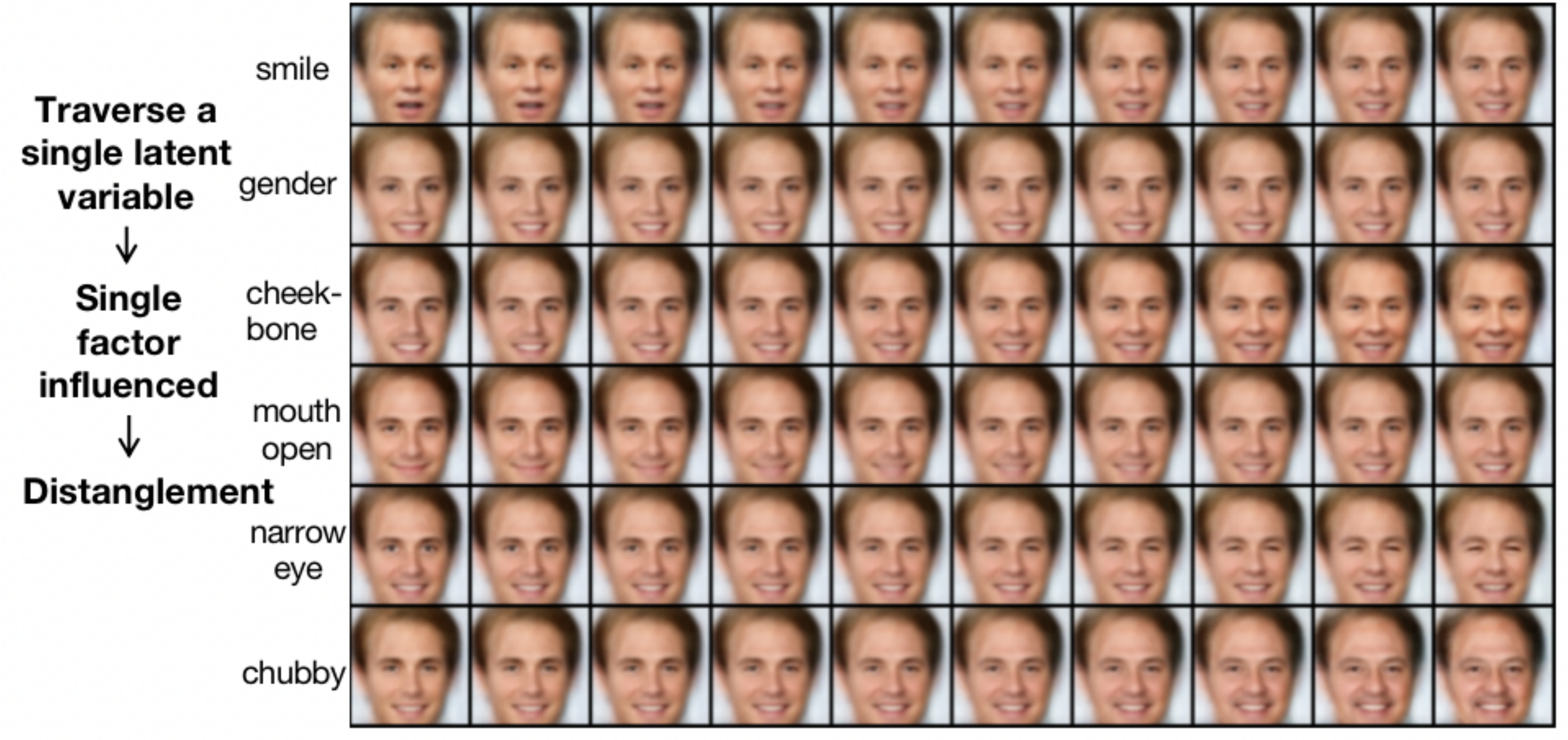}
%\caption{fig2}
\label{fig:cfvaesmiletraverse}
\end{minipage}
}%
\subfigure[Traverse of DEAR]{
\begin{minipage}[b]{0.49\linewidth}
\centering
\includegraphics[width=\textwidth]{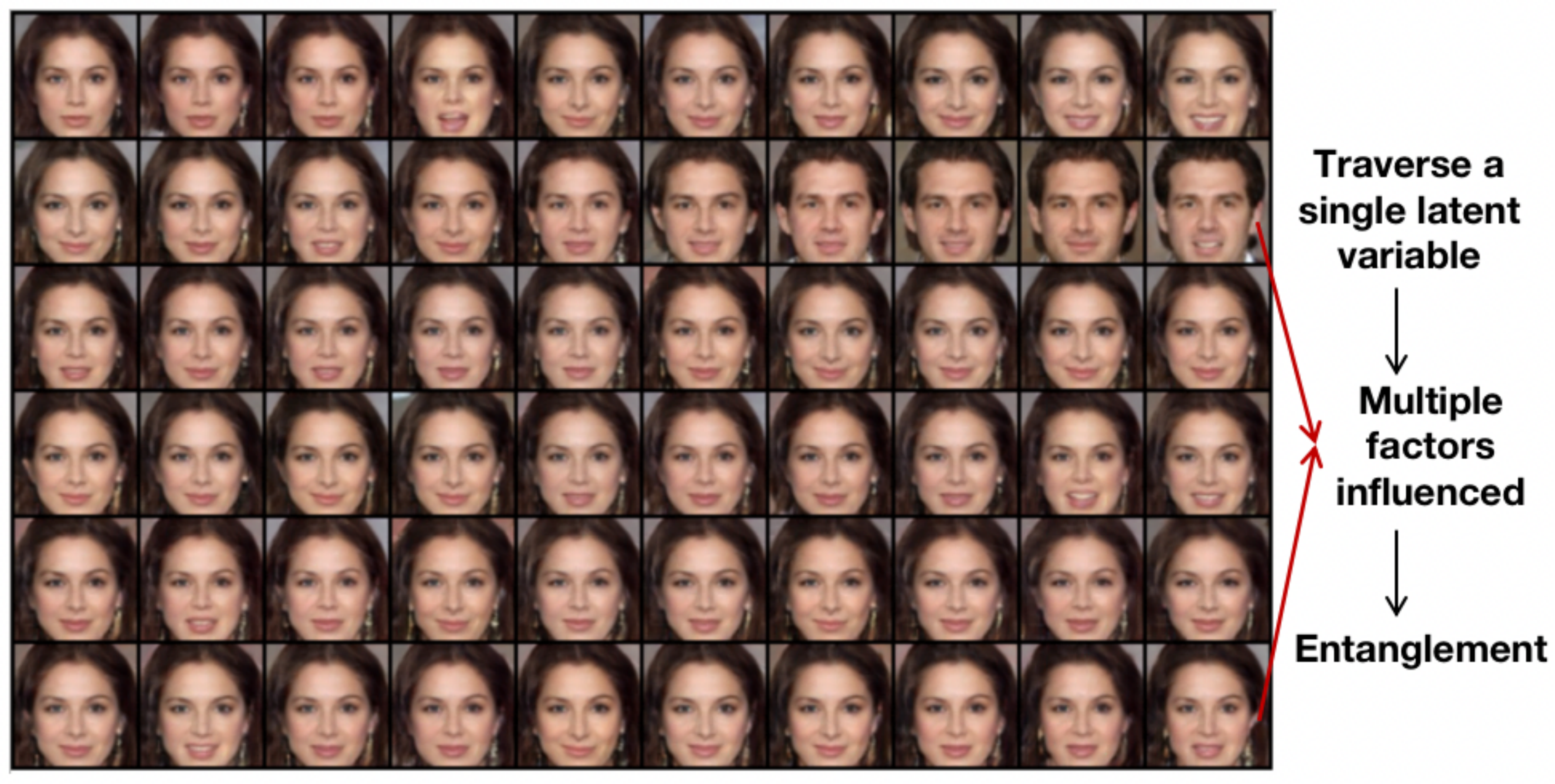}
%\caption{fig2}
\label{fig:dearsmiletraverse}
\end{minipage}
}%
\end{center}
\caption{Results of traverse experiments on CelebA(Smile). Each row corresponds to a variable that we traverse on, specifically, smile, gender, cheek bone, mouth open, narrow eye and chubby.
}
\label{fig:smiletraverse}
\end{figure}
\begin{figure}[t]
\begin{center}
\subfigure[Intervene factors of Pendulum]{
\begin{minipage}[t]{0.505\linewidth}
\centering
\includegraphics[width=\textwidth]{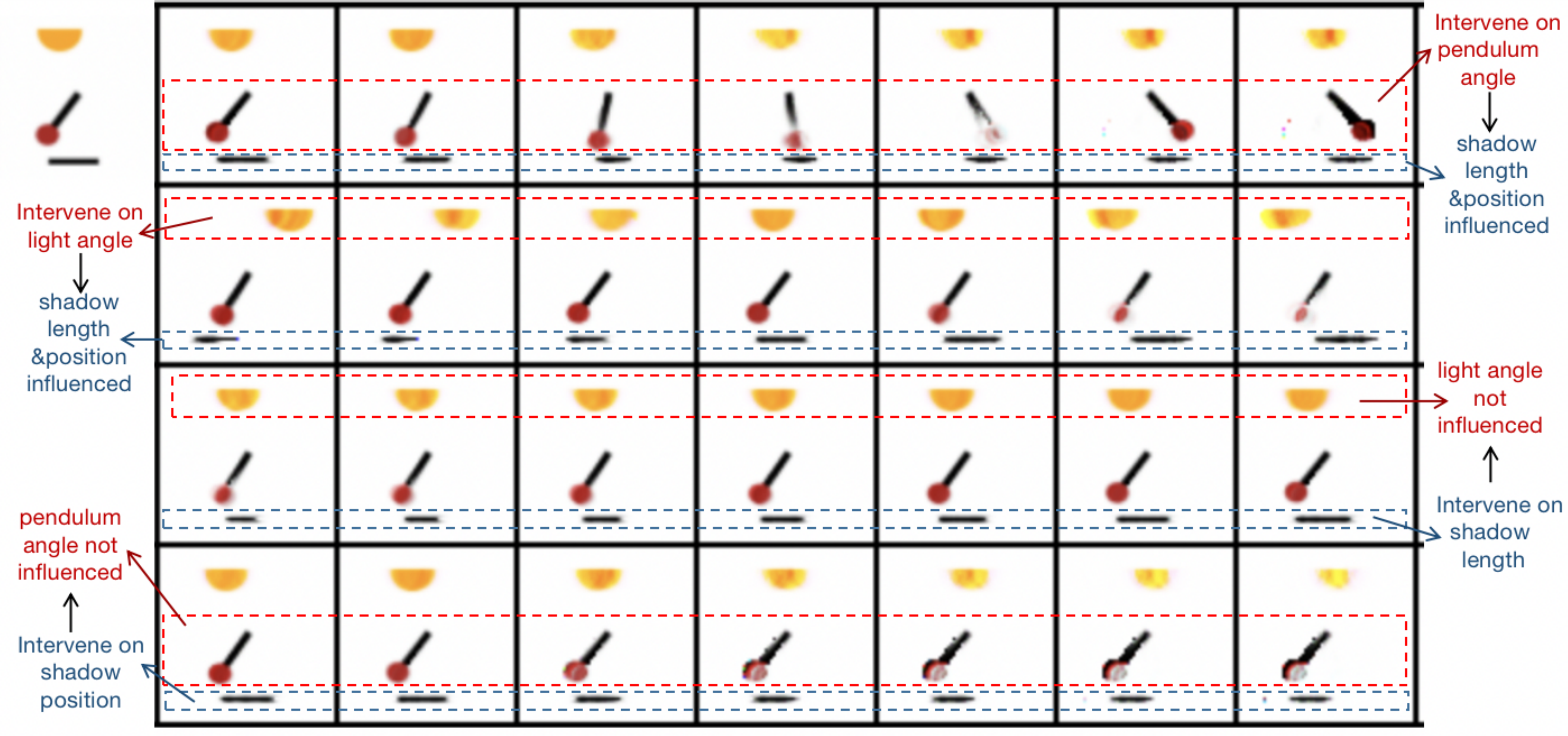}
% \caption{fig1}
\label{fig:cfvaependulumintervene}
\end{minipage}%
}%
\subfigure[Intervene factors of CelebA(Smile)]{
\begin{minipage}[t]{0.495\linewidth}
\centering
\includegraphics[width=\textwidth]{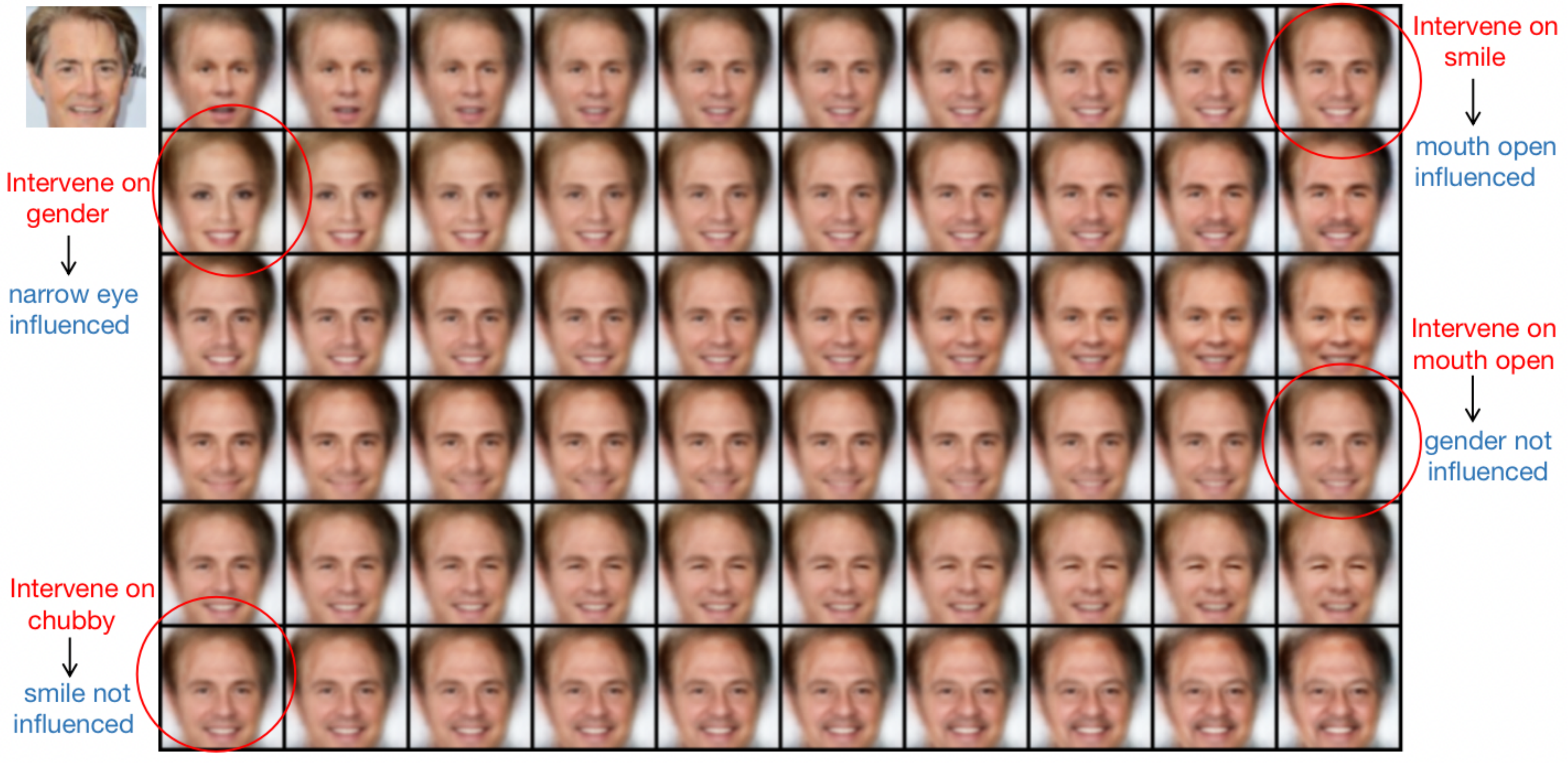}
%\caption{fig2}
\label{fig:CFVAEsmileintervene}
\end{minipage}%
}%
\end{center}
\caption{Results of intervention on only one variable for both Pendulum and CelebA(Smile). The image in the upper left corner of (a) and (b) are the test data we consider respectively.}
\label{fig:intervene}
\end{figure}
It is worth noting that we did not compare our results with the model in \citet{yang2021causalvae}. Firstly, due to the presence of a Mask layer in its decoder, it is not possible to observe changes in individual factors when traversing each dimension of the learned representation. Secondly, the latent variable dimension in the model corresponds to the number of interested latent factors, which, when applied to real-world datasets, may not enable the latent layer to capture all the generative factors of the images, thereby failing to ensure a one-to-one correspondence between latent units and generative factors, and consequently, not achieving causal disentanglement. Additionally, regarding the pendulum dataset, to ensure model performance, the authors of \citet{yang2021causalvae} assigned a multidimensional vector with a dimension of 4 to each latent factor which could not be aligned with our model. Consequently, due to the issue with the dimension of latent variables in their model, we did not perform a comparison. In contrast, DEAR, as a VAE-based causal disentanglement model, is more reasonable.
To ensure a fair comparison with equal amounts of supervised information, for each of these methods we use the same conditional prior and loss term as in DCVAE.
\subsection{Experimental Results}
Now, we proceed to evaluate our method through both qualitative and quantitative experiments and provide an analysis of the corresponding experimental results.
\subsubsection{Causal Disentangled Representations}\label{sec:Causal Disentangled Representations}
To qualitatively verify that DCVAE indeed learns causal disentangled representations, we conduct intervention experiments. 

Intervention experiments involve performing the "do-operation" in causal inference \citep{pearl2009causality}.
% which allows us to visually observe the causal disentanglement of representations. 
Taking a single-step causal flow as an example, we demonstrate step by step how our model performs "do-operation". First, given a trained model, we input the sample $\mathbf{x}$ into the inference model, obtaining an output $\widetilde{\mathbf{z}}$. Assuming we wish to perform the "do-operation" on $\widetilde{\mathbf{z}}_i$, i.e., $do(\widetilde{\mathbf{z}}_i=c)$, we follow the approach in \citet{khemakhem2021causal} by treating Eq. (\ref{flows formulation}) as SEMs. Specifically, we set the input and output of $\widetilde{\mathbf{z}}_i$ to the control value $c$, while other values are computed iteratively from input to output. Finally, the resulting $\widetilde{\mathbf{z}}$ is decoded to generate the desired image, which corresponds to generating images from the interventional distribution of $\widetilde{\mathbf{z}}$.

% To verify the effectiveness of our model in achieving causal disentangled representations, 
We perform intervention experiments by applying the "do-operation" to $m-1$ variables in the first $m$ dimensions of the latent variables, resulting in the change of only one variable. This operation, which has been referred to as "traverse", aims to test the disentanglement of our model \citep{shen2022weakly}. Figures \ref{fig:pendulumtraverse} and \ref{fig:smiletraverse} show the experimental results of the DCVAE and DEAR on Pendulum and CelebA(Smile). We observe that when traversing a latent variable dimension, DCVAE has almost only one factor changing, while DEAR has multiple factors changing. This is clearly shown by comparing "traverse" results of the third row for shadow length in Figures \ref{fig:cfvaependulumtraverse} and \ref{fig:dearpendulumtraverse}, as well as the second row for gender in Figures \ref{fig:cfvaesmiletraverse} and \ref{fig:dearsmiletraverse}. Therefore, our model performs better in achieving causal disentanglement.
%包含MIC/TIC，traverse，intervention，反事实图片。对比实验在附录

To demonstrate our model's capability to perform interventions hence generating new images beyond the dataset, we further conduct "do-operations" on individual latent variables. Figure \ref{fig:intervene} illustrates these operations, with each row representing an intervention on a single dimension. In Figure \ref{fig:cfvaependulumintervene}, we observe that intervening on the pendulum angle and light angle produces changes in shadow length in accordance with physical principles. However, intervening on shadow length has minimal impact on these two factors. Similarly, as illustrated in Figure \ref{fig:CFVAEsmileintervene}, intervening on gender influences narrow eye appearance, but the reverse is not true. This demonstrates that intervening on causal factors affects the resulting effects, but not the other way around. Hence, our latent variables have effectively learned factor representations, attributed to the design of the causal flows, which incorporate $A$. Additional traversal and intervention results are presented in Appendix \ref{Appendix:Additional Results}.

\textcolor{black}{The clarity of the images generated by our model, as shown in Figure \ref{fig:cfvaependulumtraverse}, is somewhat lower compared to those produced by \citet{an2023causally}. However, the images generated from the CelebA(Smile) dataset, as shown in Figure \ref{fig:cfvaesmiletraverse}, are clearer than those in Figure \ref{fig:cfvaependulumtraverse}. One possible solution is to increase the sampling size of the datasets used in future practical applications as much as possible when using the model. On the other hand, the Causally Disentangled Generation (CDG) model by \citet{an2023causally} significantly enhances the VAE decoder, resulting in clearer counterfactual images by accurately reflecting the data generation process. In contrast, our model focuses on optimizing the encoder to learn disentangled representations and relies on the decoder to generate high-quality images. 
% While \citet{an2023causally} concentrates on improving decoder accuracy, their approach has significant limitations, including the need for complete causal information for supervision and its unsuitability for large-scale datasets like CelebA. 
Therefore, to address the issue of image clarity, we will explore methods to improve our model's performance, including integrating techniques from \citet{an2023causally} or other methods to enhance the decoder’s performance. This would help ensure higher image quality in our model's output. In summary, addressing this challenge will be a key focus of our future work.}
\begin{table}[b]
\begin{center}
\caption{Test accuracy and sample efficiency of different models on Pendulum and CelebA datasets. Mean$\pm$standard deviations are
included in the Table. }
\scalebox{0.9}{
\begin{tabular}{ccccccc}
\toprule
% \multirow{2}{1.2cm}{\textbf{Model}} 
& \multicolumn{3}{c}{\textbf{Pendulum}} & \multicolumn{3}{c}{\textbf{CelebA}} \\
\cmidrule{2-7}  % 这部分是画一条横线在2-6 排之间
\textbf{Model}  &   100(\%) & All(\%) & \footnotesize{Sample Eff} & 100(\%) & 10000(\%) & \footnotesize{Sample Eff} \\
\midrule
  DCVAE & $\boldsymbol{99.00_{\pm 0}}$ & $99.43_{\pm0.34}$  & $\boldsymbol{99.57_{\pm 0.34}}$ &$\boldsymbol{81.00_{\pm 1.73}}$ & $\boldsymbol{81.54_{\pm 1.76}}$ & $\boldsymbol{99.34_{\pm 0.27}}$\\
  DCVAE-SP & $98.67_{\pm 0.58}$ & $\boldsymbol{99.79_{\pm0.14}}$  & $98.87_{\pm 0.70}$ &$79.67_{\pm 1.15}$ & $81.32_{\pm 0.47}$ & $97.98_{\pm 1.69}$\\
  DEAR & $88.00_{\pm 0}$&	$88.55_{\pm 0.04}$ & $98.63_{\pm 1.33}$&$61.00_{\pm 3.60}$  & $68.50_{\pm 0}$&$89.05_{\pm 5.26}$\\
  $\beta$-VAE & $98.67_{\pm 1.15}$	&$99.59_{\pm 0.07}$  & $98.94_{\pm 0.92}$&$62.33_{\pm 5.69}$  & $68.49_{\pm 0.02}$	&$91.01_{\pm 8.28}$\\
  $\beta$-TCVAE & $97.67_{\pm 1.15}$& $99.38_{\pm 0.48}$  & $98.27_{\pm 0.79}$ & $75.33_{\pm 3.21}$  &$78.72_{\pm 4.93}$	& $95.83_{\pm 4.10}$\\
  VAE & $98.33_{\pm 0.58}$	& $99.48_{\pm 0.39}$  & $98.72_{\pm 0.37}$& $60.33_{\pm 2.89}$ & $68.50_{\pm 0}$& $88.08_{\pm 4.21}$\\
\bottomrule
\end{tabular}}
\label{tab:sample_efficiency}
\end{center}
\end{table}
\begin{table}[t]
\begin{center}
    \caption{Distributional robustness of different models.}
    \scalebox{0.9}{
    \begin{tabular}{ccc}
    \toprule
    \textbf{Model}  &  \textbf{TestAvg(\%)} & \textbf{TestWorst(\%)} \\
    \midrule
      DCVAE & $\boldsymbol{97.83_{\pm1.18}}$  & $\boldsymbol{94.70_{\pm3.41}}$\\
      DCVAE-SP & $97.24_{\pm0.49}$  & $92.43_{\pm0.54}$\\
      DEAR & $80.40_{\pm 0.47}$ & $64.50_{\pm2.67}$\\
      $\beta$-VAE & $96.48_{\pm2.06}$  & $90.05_{\pm5.44}$\\
      $\beta$-TCVAE & $96.57_{\pm1.33}$  & $90.27_{\pm3.74}$\\
      VAE & $95.14_{\pm3.46}$  & $88.81_{\pm5.44}$\\
    \bottomrule
    \end{tabular}}
    \label{tab:robust}
\end{center}
\end{table}
\subsubsection{Downstream tasks}To quantitatively illustrate \textcolor{black}{the} benefits of causal disentangled representations, we consider its impact on downstream tasks in terms of sample efficiency and distributional robustness. 
%A common downstream task is classification, thus 
We introduce two downstream prediction tasks to compare our model with baseline models. First, for Pendulum, we normalize factors to $\left[-1, 1\right]$ during preprocessing. Then, we manually create a classification task: if $pendulum\,angle> 0$ and $light\,angle>0$, the target label $y=1$; otherwise, $y=0$. 
%This represents a positive class if both the pendulum and light are on the right side and a negative class otherwise. 
For the CelebA(Attractive), we adopt the same classification task as  presented in \citep{shen2022weakly}.
We employ multilayer perceptron (MLP) to train classification models, where both the training and testing sets consist of the latent representations $\widetilde{\mathbf{z}}$
% obtained from the encoding of the input samples $\mathbf{x}$, 
and their corresponding labels $\mathbf{y}$. 

\textbf{Sample Efficiency:} The experimental results are presented in Table \ref{tab:sample_efficiency}. We adopt the statistical efficiency score defined in \citet{locatello2019challenging} %and \citet{shen2022weakly} 
as a measure of sample efficiency, which is defined as the classification accuracy of 100 test samples divided by the number of all (Pendulum)/10,000 test samples (CelebA). Table \ref{tab:sample_efficiency} shows that DCVAE achieves the best sample efficiency on both datasets. \textcolor{black}{DCVAE-SP exhibits slightly inferior performance compared to DCVAE because the causal information input into DCVAE-SP is weaker than that in DCVAE, which contributes to its reduced performance,} but it remains competitive when compared to other baseline methods. 
Hence, while the model's optimal performance is achieved by integrating information from the full graph, causal flows can still contribute with slightly diminished effectiveness in the case of the super-graph. This emphasizes the significance of incorporating causal flow in our model.
% However, on the complex CelebA dataset, DCVAE significantly outperforms $\beta$-VAE. 
% Among the baseline models, none stand out and all perform worse than DCVAE. 
% It is worth noting that the encoder structures of baseline models are identical, except for DEAR, which uses ResNet as the encoder but also exhibits strong learning capacity.
We attribute the superiority of DCVAE to our modeling approach, which leverages the capabilities of causal flows and incorporates conditional prior. This greatly enhances the encoder's ability to learn semantically meaningful and expressive representations.
% where we report both the test accuracy and sample efficiency.

\textbf{Distributional robustness:} To assess distributional robustness, we modify the controllable synthetic Pendulum dataset during training to inject spurious correlations between the target label and some spurious attributes. We choose $background\_color \\\in \{blue(+), white(-)\}$ as a spurious feature. Specifically, in 80\% of the examples, the target label and the spurious attribute are both positive or negative, while in 20\% of the examples, they are opposite. For instance, in the manipulated training set, 80\% \textcolor{black}{of the} positive examples in Pendulum are masked with a blue background. \textcolor{black}{However}, in the test set, we do not inject this correlation, resulting in a distribution shift. The results are summarized in Table \ref{tab:robust}.

The results include average and worst-case test accuracy, evaluating overall classification performance and distributional robustness. Worst-case accuracy identifies the group with the lowest accuracy among four groups categorized based on target and spurious binary labels. It often involves opposing spurious correlations compared to training data. The classifiers trained using DCVAE representations demonstrate significant superiority over all baseline models in both evaluation metrics. Notably, DCVAE experiences a smaller decrease in worst-case accuracy compared to average accuracy, indicating robustness to distributional shifts.

\textcolor{black}{In the aforementioned experiments, we analyzed the reasons why our model outperforms DEAR, which are primarily twofold. First, DEAR uses an SCM prior to guide latent representation learning indirectly through loss minimization, whereas our model integrates causal flow directly into the VAE, using flow model characteristics to enhance latent representations. Second, DEAR employs a fixed transformation function in its SCM prior, which may not align with the true causal function and affects performance. In contrast, our model utilizes causal flows, leveraging the benefits of autoregressive flows to adaptively learn superior transformation functions and improve VAE’s generative capability.}
\begin{figure}[b]
\begin{center}
\subfigure[Original $A$]{
\begin{minipage}[t]{0.2\linewidth}
\centering
\includegraphics[width=\textwidth]{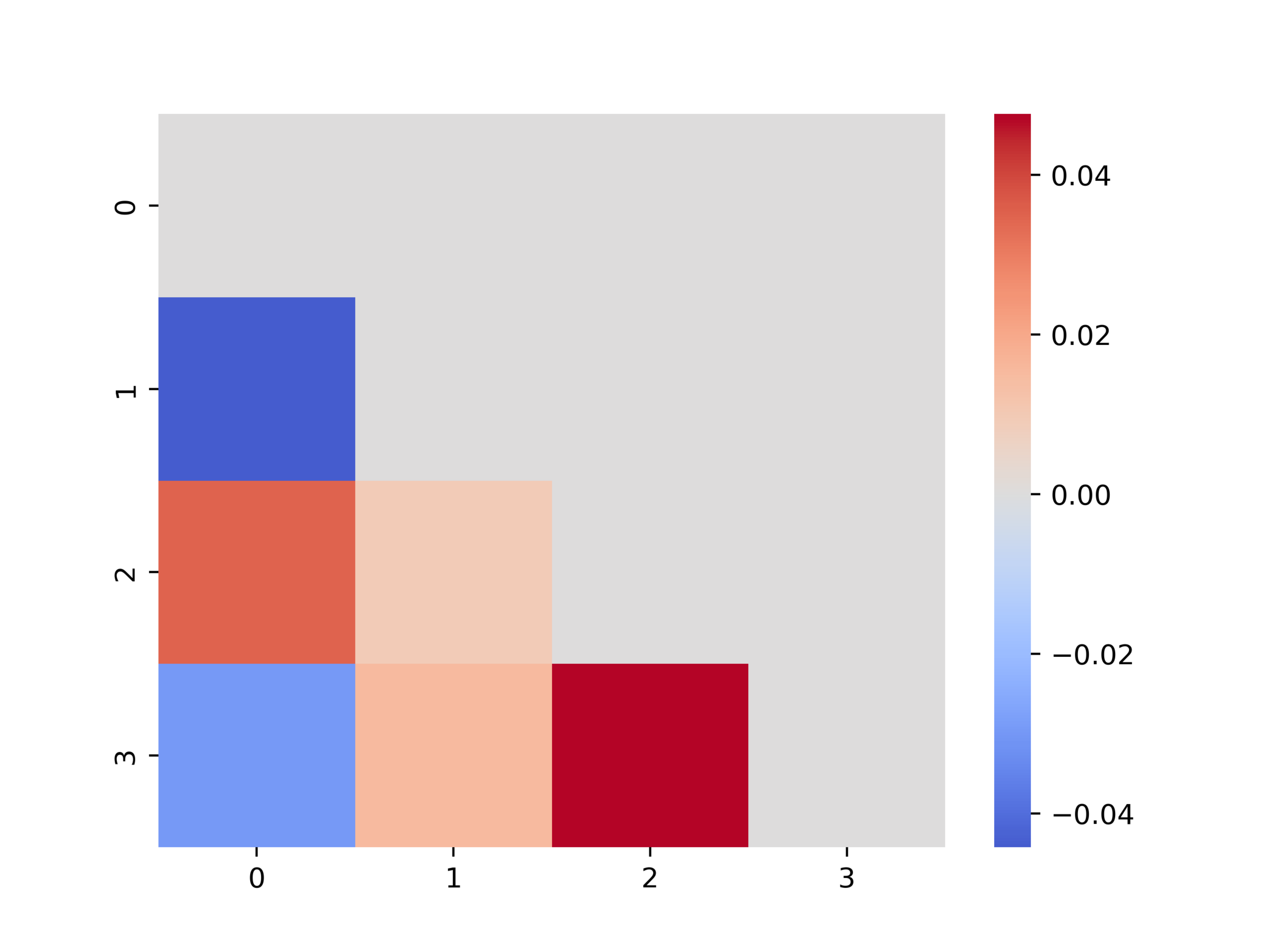}
% \caption{fig1}
\label{fig:pendulumprioraoa0}
\end{minipage}%
}%
\subfigure[After 20 epochs]{
\begin{minipage}[t]{0.2\linewidth}
\centering
\includegraphics[width=\textwidth]{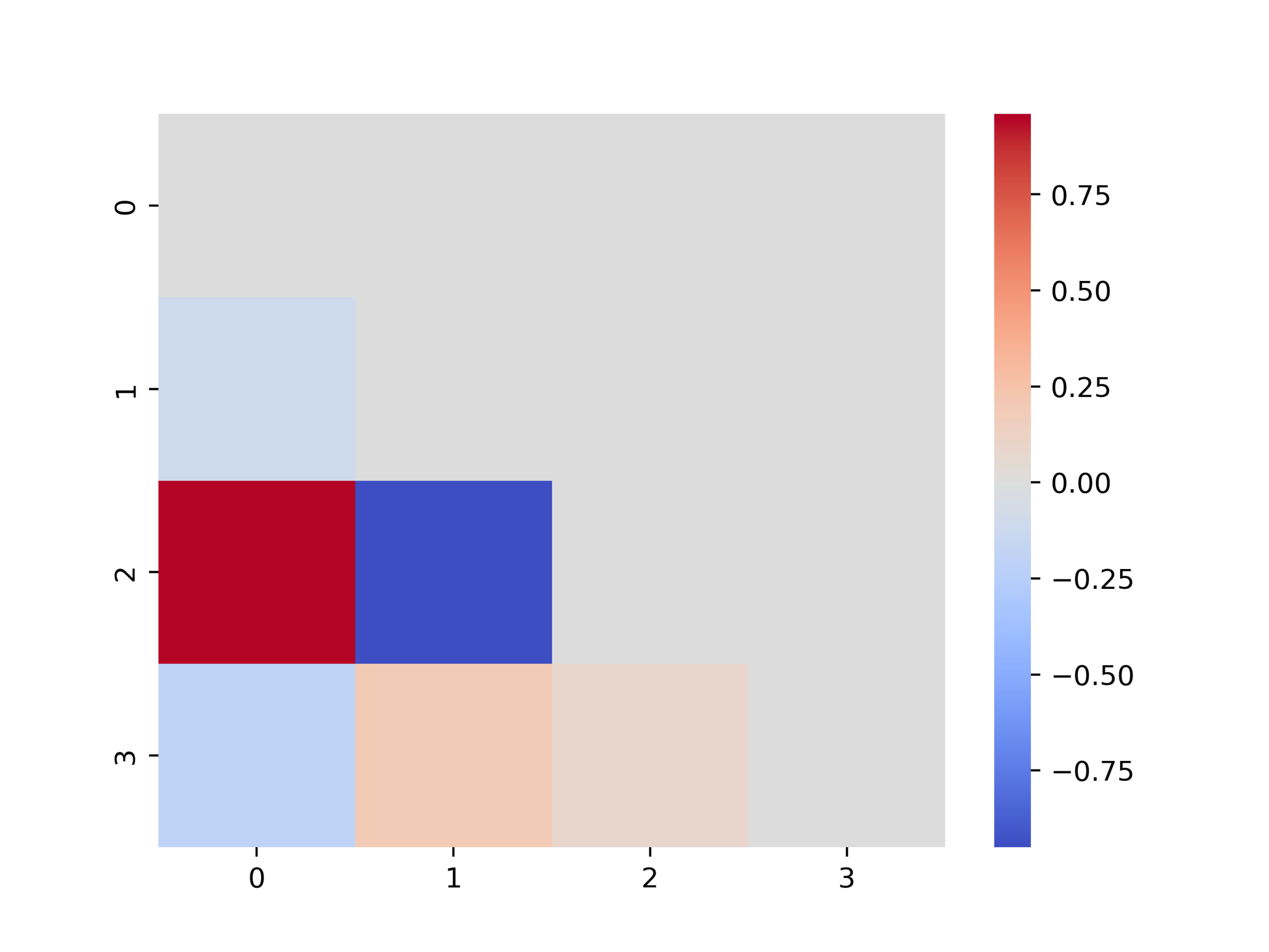}
%\caption{fig2}
\label{fig:pendulumprioraoa20}
\end{minipage}%
}%
\subfigure[After 50 epochs]{
\begin{minipage}[t]{0.2\linewidth}
\centering
\includegraphics[width=\textwidth]{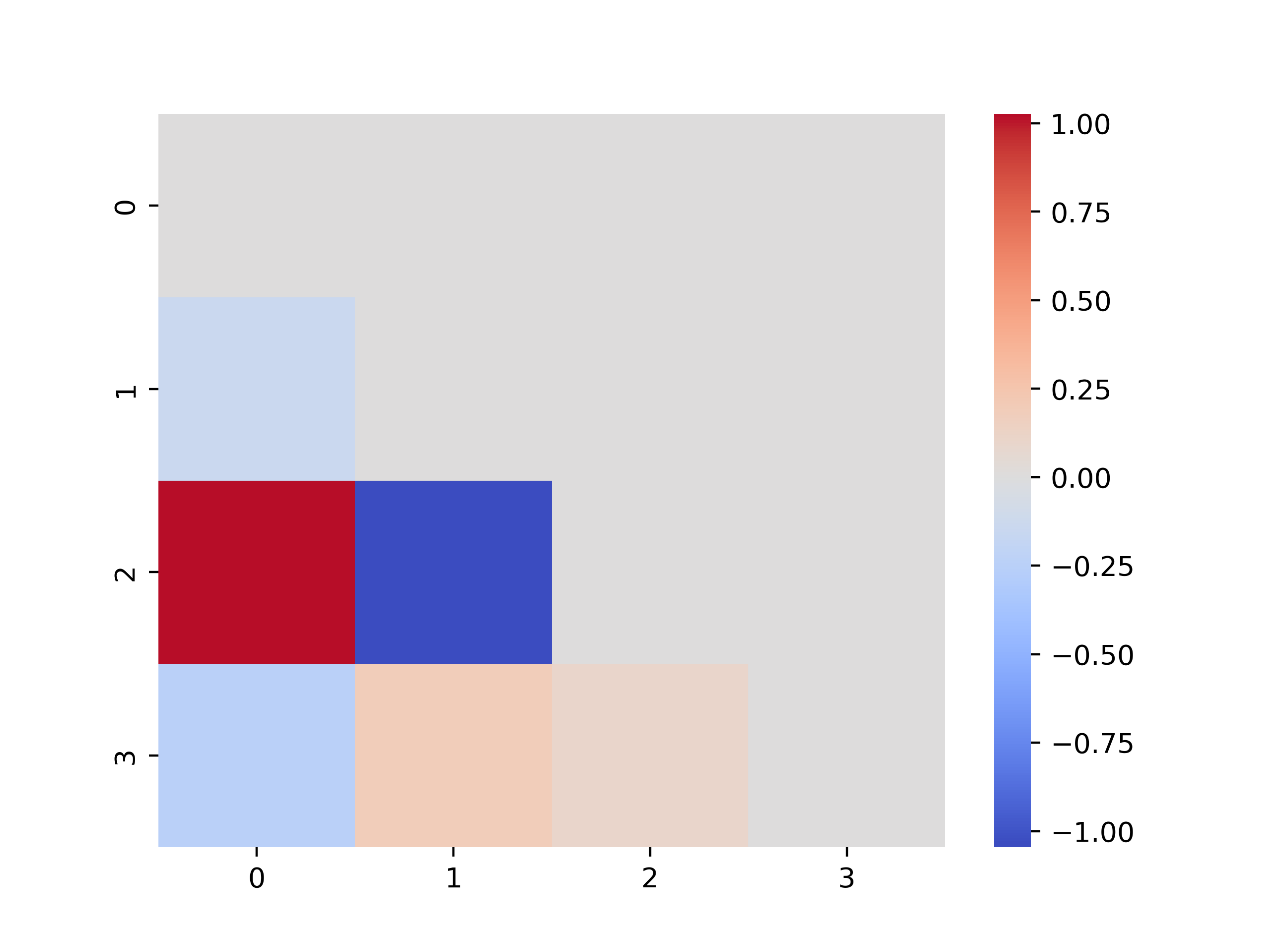}
%\caption{fig2}
\label{fig:pendulumprioraoa50}
\end{minipage}%
}%
\subfigure[After 80 epochs ]{
\begin{minipage}[t]{0.2\linewidth}
\centering
\includegraphics[width=\textwidth]{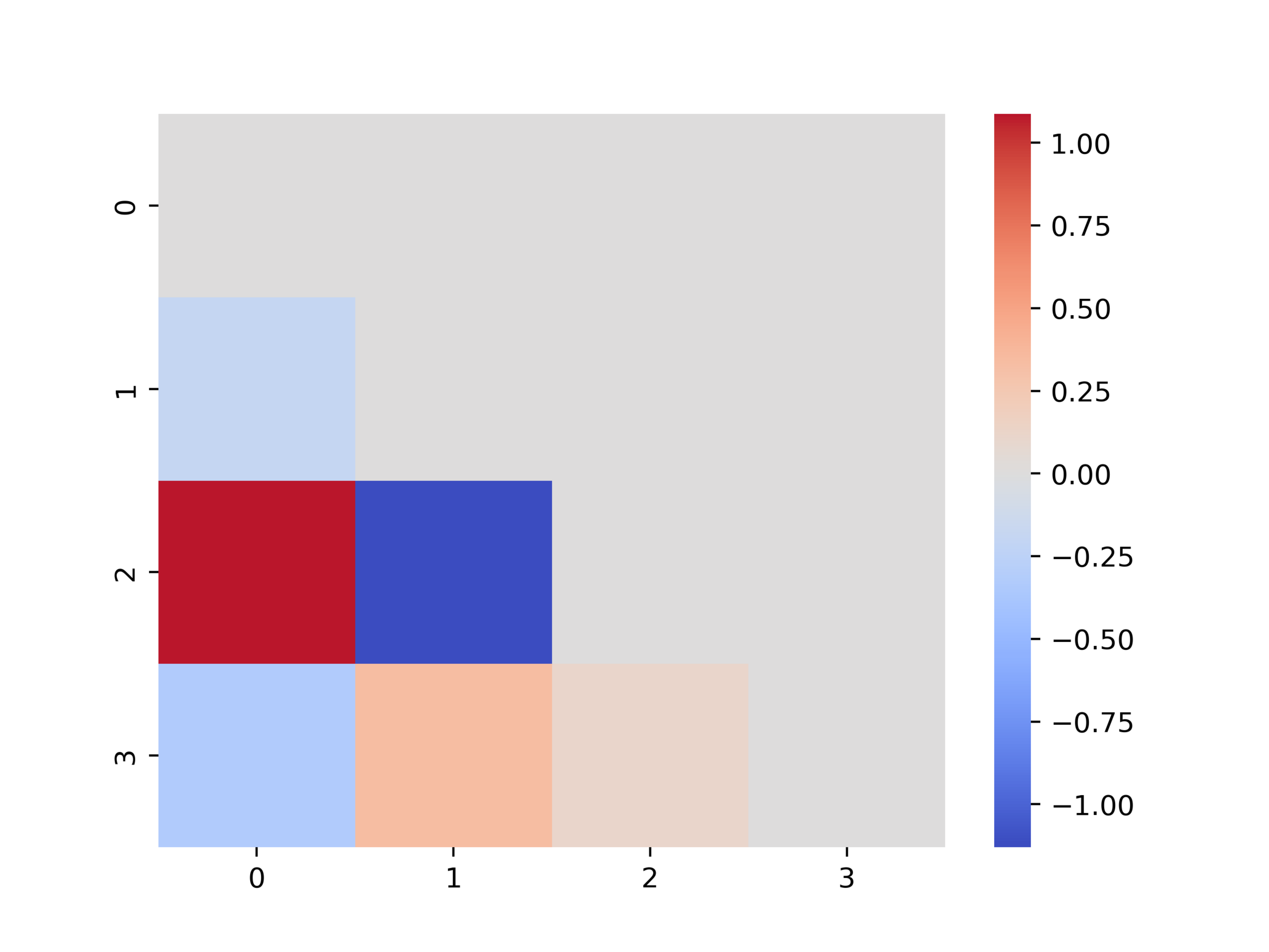}
%\caption{fig2}
\label{fig:pendulumprioraoa80}
\end{minipage}%
}%
\subfigure[Final $A$]{
\begin{minipage}[t]{0.2\linewidth}
\centering
\includegraphics[width=\textwidth]{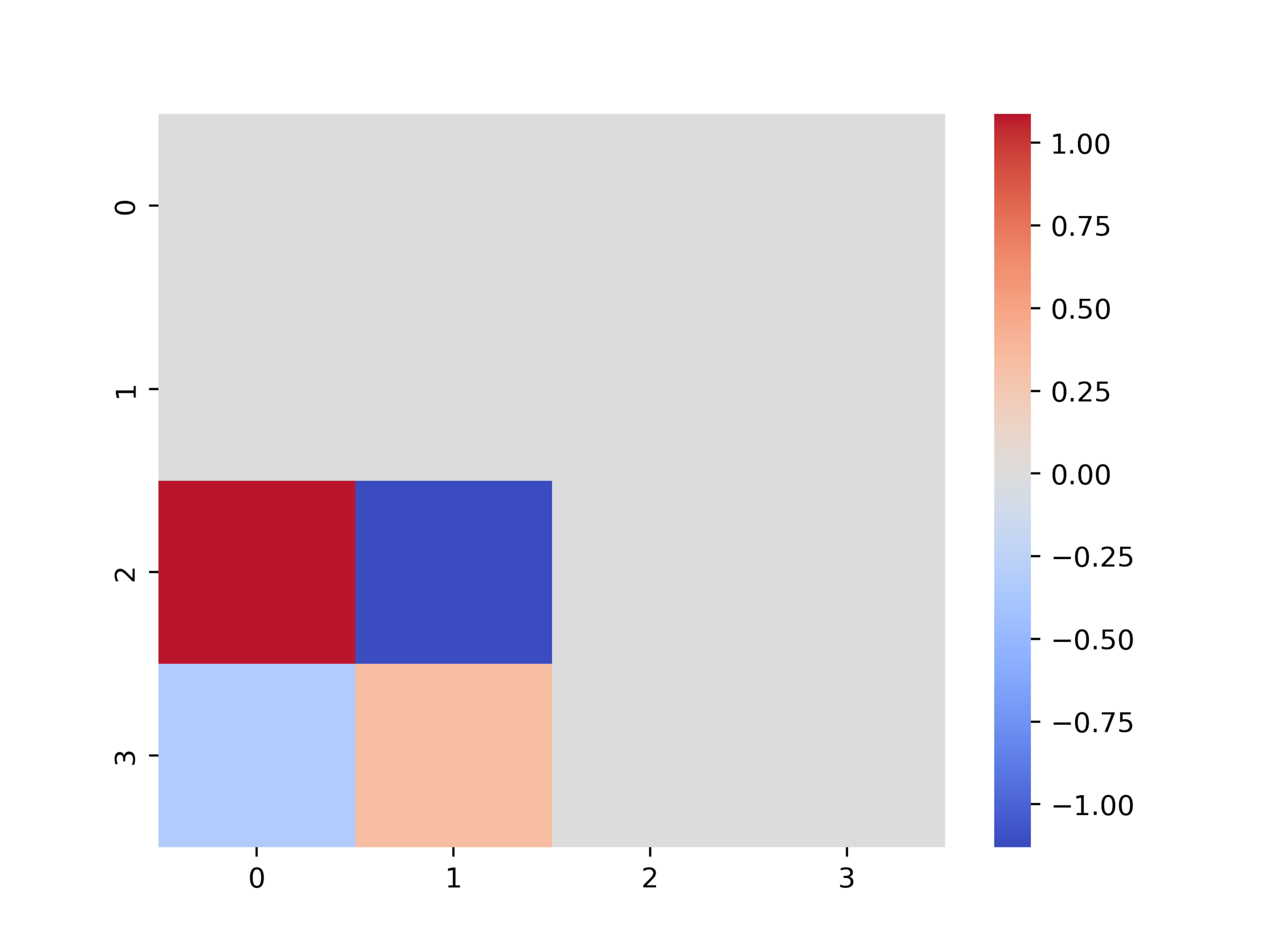}
%\caption{fig2}
\label{fig:pendulumprioraoa80filter}
\end{minipage}%
}%
\end{center}
 \caption{The learned weighted adjacency matrix $A$ given the causal ordering on Pendulum. (a)-(d) illustrate the changes in $A$ as the training progresses. (e) represents $A$ after edge pruning.
}
\label{fig:A}
\end{figure}
\begin{figure}[!t]
\centering
\subfigure[Pendulum]{
\begin{minipage}[t]{0.5\linewidth}
\centering
\includegraphics[width=0.5\textwidth]{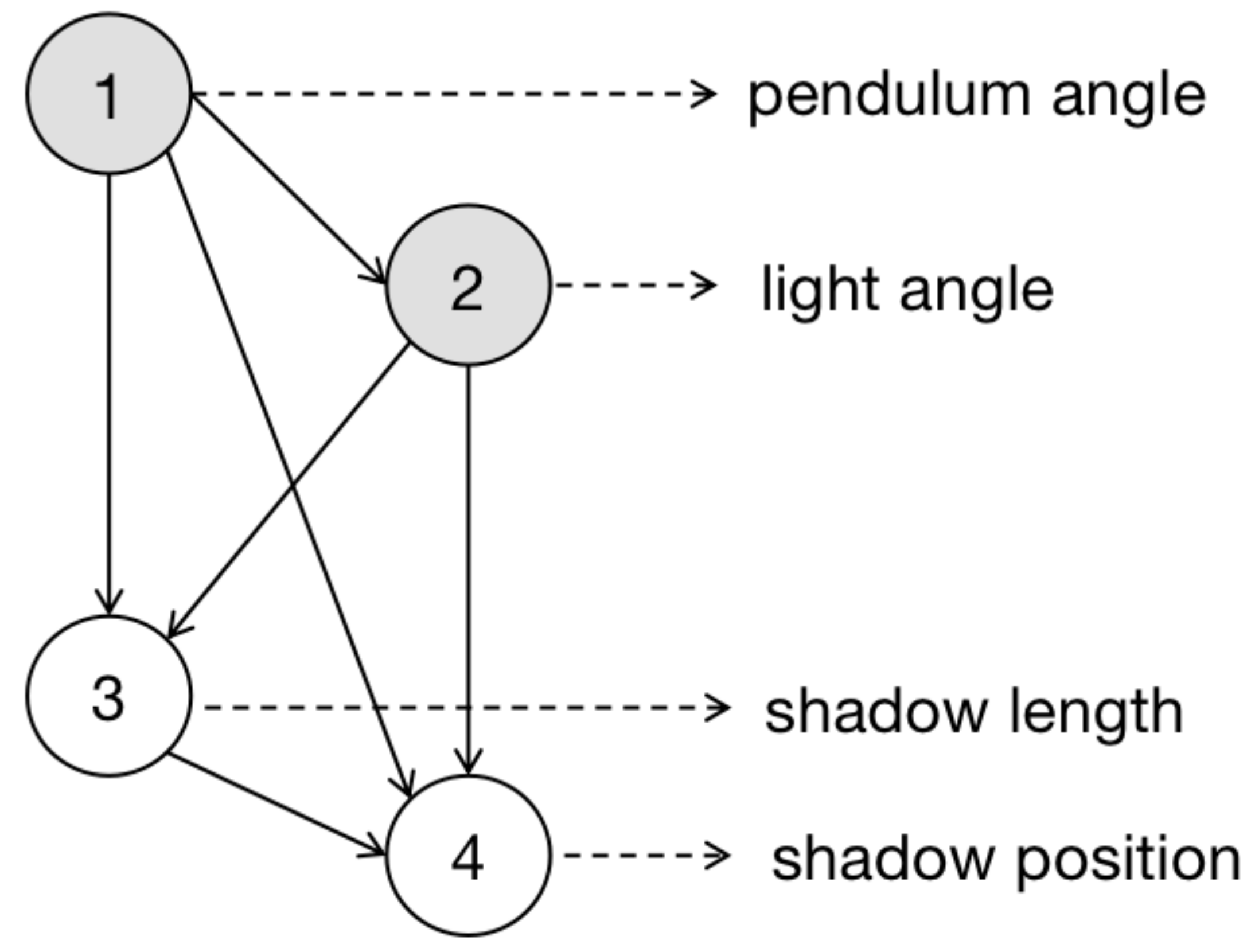}
% \caption{fig1}
\label{fig:supergraphpendulum}
\end{minipage}%
}%
\subfigure[CelebA(Attractive)]{
\begin{minipage}[t]{0.5\linewidth}
\centering
\includegraphics[width=0.6\textwidth]{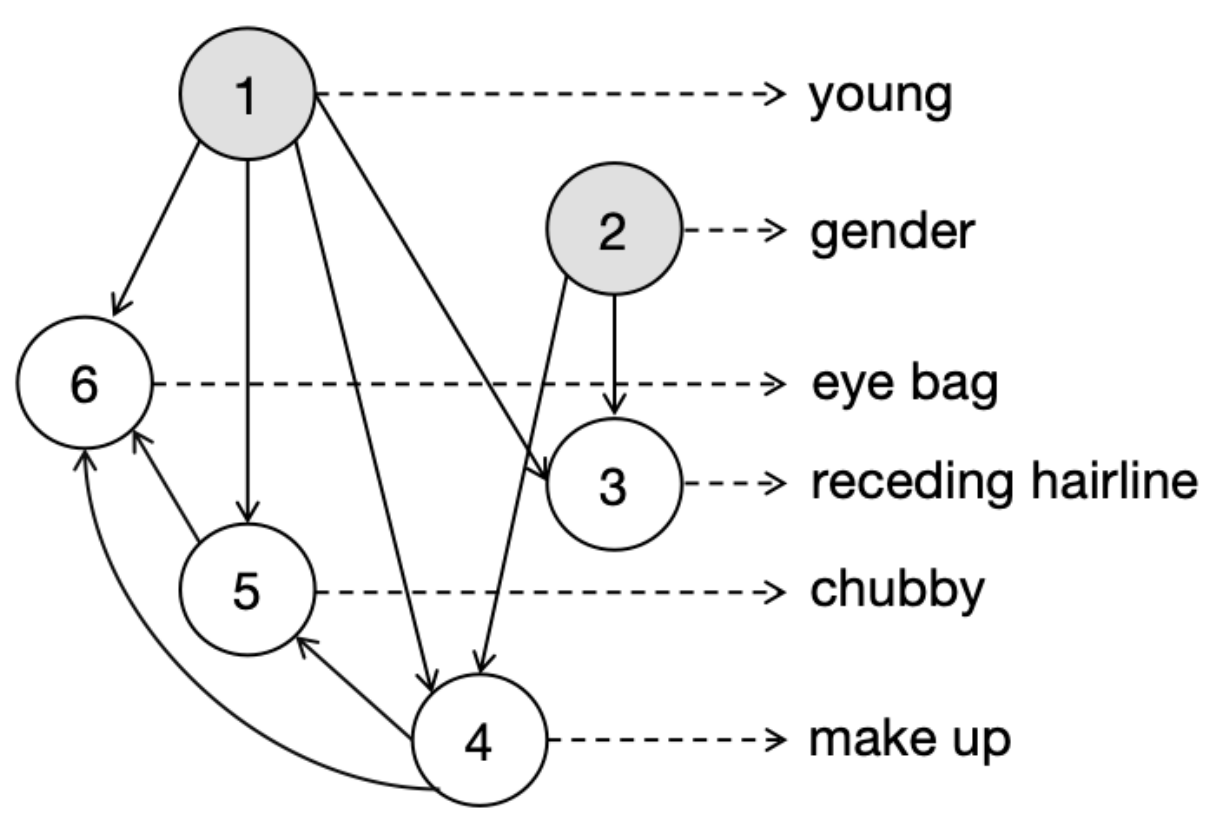}
%\caption{fig2}
\label{fig:super_graph_attractive_text}
\end{minipage}%
}%
\centering
\caption{Super-graph of Pendulum and CelebA(Attractive).}
\label{fig:super_graph_all_text}
\end{figure}
\subsubsection{Exploring the potential for learning the  structure \textit{A}}
Apart from the aforementioned applications, DCVAE has the potential to learn true causal relationships between factors, despite the fact that it is not the primary focus of our work.
% even without using a SCM \citep{yang2021causalvae,pearl2000models}. 
As shown in Figure \ref{fig:pendulumprioraoa0}-\ref{fig:pendulumprioraoa80}, for Pendulum, when our model $A$ adopts the full graph shown in Figure \ref{fig:supergraphpendulum}, though the corresponding $A$ is initialized randomly around 0, it gradually approaches the true causal structure during the training process. 
If we apply a threshold and prune edges in the causal graph,
%graph with values below this threshold
we obtain Figure \ref{fig:pendulumprioraoa80filter}, which corresponds to the true causal structure depicted in Figure \ref{fig:pendulumgraph}. 
At this point, only four positions in matrix A are relatively large non-zero values, indicating accurate causal relationships. For the experiments conducted on CelebA dataset, please refer to Appendix \ref{Appendix:Additional Results}.
This preliminary result suggests that our model has the potential to learn causal structures even without relying on a Structural Causal Model (SCM) \citep{yang2021causalvae,pearl2000models}, inspiring us to explore this topic further in future research.

\section{Conclusion and Points of Future Research}\label{7conclusion}
In this paper, we focus on addressing the significant challenge of learning causal disentangled representations using VAE when the underlying generative factors are causally related. We propose the Disentangled Causal Variational Auto-Encoder (DCVAE), which leverages causal flows to integrate the causal structure information of generative factors into the model. Empirical results, including quantitative and qualitative experiments on synthetic and real-world datasets, validate our method's success in learning causal disentangled representations and generating counterfactual outputs. 

Future research directions present several opportunities. First, there is potential to extend this approach to model causal relationships among a broader set of generative factors, which is particularly relevant in the context of big data. Currently, our work offers a framework-based approach in this regard.
Second, future studies could explore alternative forms of supervised information for disentanglement learning, as well as investigate weakly supervised or unsupervised methods to achieve causal disentangled representations within the flow model framework.
Finally, a deeper exploration of the inherent structure of data could facilitate the learning of low-dimensional causal disentangled representations. Additionally, applying these representations to areas such as video generation presents a promising research direction.
\section*{Acknowledgments}
This work was funded by the National Nature Science Foundation of China under Grant No. 12320101001 and 12071428.
% \clearpage
\bibliographystyle{ACM-Reference-Format}
\bibliography{sample-base}
\appendix
\setcounter{equation}{0}
\gdef\theequation{A.\arabic{equation}}

\section{Experimental Details}\label{Appendix:Experimental Details}
\setcounter{table}{0}
\gdef\thetable{A.\arabic{table}}
\begin{table}[b]
\renewcommand{\arraystretch}{0.8}
\begin{center}
\caption{Architecture for the Encoder and Decoder in DCVAE ($d=4$ for Pendulum and $d=100$ for CelebA).}

\begin{tabular}{cc}
\toprule
\textbf{Encoder}  &   \textbf{Decoder} \\
\midrule
- & Input ${\widetilde{\mathbf{z}}}\in \mathbb{R}^d$\\
  3$\times$3 conv, MaxPool, 8 SELU, stride 1 & FC, 256 SELU\\
  3$\times$3 conv, MaxPool, 16 SELU, stride 1 & FC, 16$\times$12$\times$12 SELU\\
  3$\times$3 conv, MaxPool, 32 SELU, stride 1 & 2$\times$2 conv, 32 SELU, stride 1\\
  3$\times$3 conv, MaxPool, 64 SELU, stride 1 & 2$\times$2 conv, 64 SELU, stride 1\\
  3$\times$3 conv, MaxPool, 8 SELU, stride 1 & 2$\times$2 conv, 128 SELU, stride 1\\
  FC 256$\times$2 & FC, 3$\times$64$\times$64 Tanh\\
\bottomrule
\end{tabular}
\label{tab:Architecture for the Encoder and Decoder in DCVAE}
\end{center}
\end{table}
\begin{table}[b]
\renewcommand{\arraystretch}{0.8}
\begin{center}
\caption{Hyperparameters of DCVAE.}
\begin{tabular}{lcc}
\toprule
\textbf{Parameters}&\textbf{Values (Pendulum)}  &   \textbf{Values (CelebA)} \\
\midrule
Batch size & 128 & 128\\
Epoch & 801 & 101\\
Latent dimension & 4 & 100\\
$\sigma$ & 0.1667 & 0.1667\\
 $\beta_{sup}$ & 8 & 5\\
 $\beta_{1}$ & 0.2 & 0.2\\
 $\beta_{2}$ & 0.999 & 0.999\\
 $\epsilon$ & 1e$-$8 & 1e$-$8\\
 Learning rate of Encoder & 5e$-$5 & 3e$-$4\\
 Learning rate of Causal Flow & 5e$-$5 & 3e$-$4\\
 Learning rate of $A$ & 1e$-$3 & 1e$-$3\\
 Learning rate of Conditional prior & 5e$-$5 & 3e$-$4\\
 Learning rate of Decoder & 5e$-$5 & 3e$-$4\\
\bottomrule
\end{tabular}
\label{tab:Hyperparameters of DCVAE}
\end{center}
\end{table}
We present the network architecture and hyperparameters used in our experiments. 
The network structures of encoder and decoder are presented in Table \ref{tab:Architecture for the Encoder and Decoder in DCVAE}. The encoder's output, i.e., $mean$ and ${\rm log}\,variance$, share parameters except for the final layer. A single-layer causal flow implemented by an MLP implementation is used for fitting the posterior distribution, and a causal weight matrix $A$ is added in a manner inspired by \citet{wehenkel2021graphical}. \textcolor{black}{Since the original encoder output $\mathbf{z}$ has already learned the information from $\mathbf{x}$, we only input the $\mathbf{z}$ into the flow.}

% We set the additional information $\mathbf{u}$ to be the same as $\boldsymbol{y}$, so that we only need to use one type of supervised information.
The decoder's output is resized to generate pixel values for a three-channel color image. 
% The $\beta_{sup}$ must be chosen carefully. If it is too small, the supervision loss will have little effect, while if it is too large, it will affect the capacity of the encoder and the flow model. 
% $\beta_{sup}$ is roughly tuned during our hyperparameter selection process. 
We train the model using the Adam optimizer and all the training parameters are shown in Table \ref{tab:Hyperparameters of DCVAE}. Although we cannot guarantee finding the optimal solution in the experiments, the empirical results still demonstrate the excellent performance of our model.
\section{Additional Results}\label{Appendix:Additional Results}
\setcounter{figure}{0}
\gdef\thefigure{B.\arabic{figure}}
\subsection{Samples from interventional distributions} 
\begin{figure}[t]
\begin{center}
%\vspace{-0.35cm} %设置与上面正文的距离
%\setlength{\abovecaptionskip}{0pt}
% \subfigtopskip=5pt %设置子图与上面正文或别的内容的距离
% \subfigbottomskip=0pt %设置第二行子图与第一行子图的距离，即下面的头与上面的脚的距离
% %\subfigcapskip=-5pt %设置子图与子标题之间的距离
	\subfigure[Female gradually with receding hairline
]{
		\begin{minipage}{0.45\linewidth} %[b]%{0.2\textwidth} 
  \centering   
                        \includegraphics[width=\textwidth]{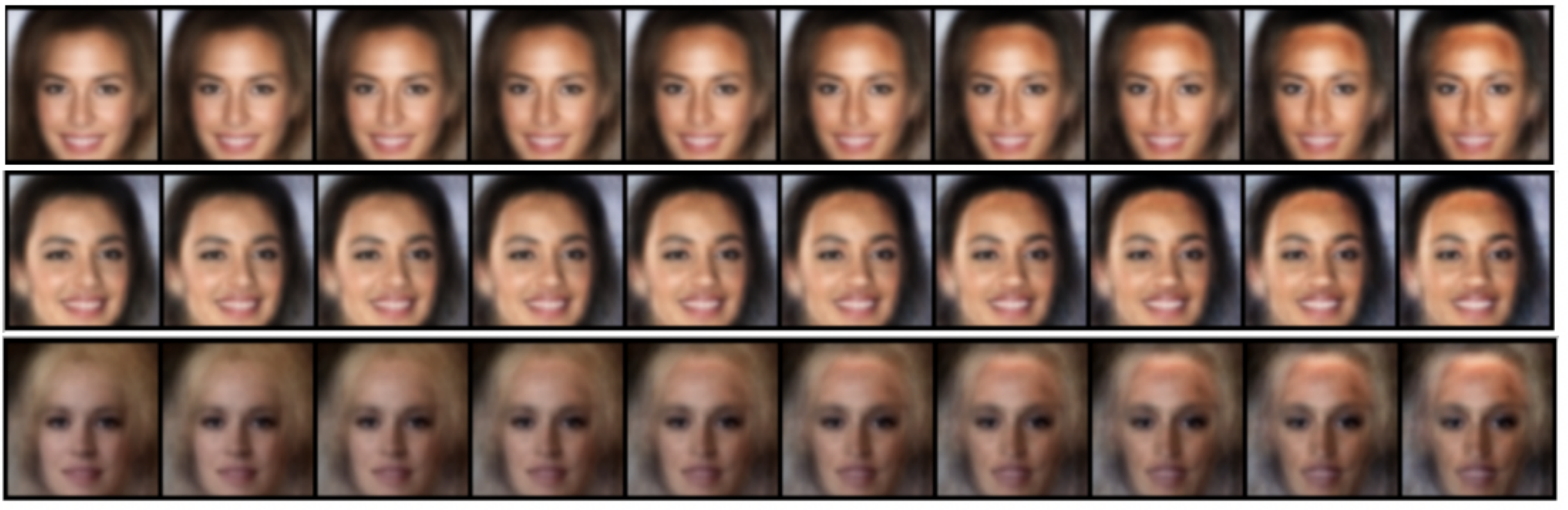} \\
                        \label{fig:intervene1}
		\end{minipage}
	}
	\subfigure[Male gradually with make up]{
		\begin{minipage}{0.45\linewidth}%[b]%{0.2\textwidth}
  \centering   
			\includegraphics[width=\textwidth]{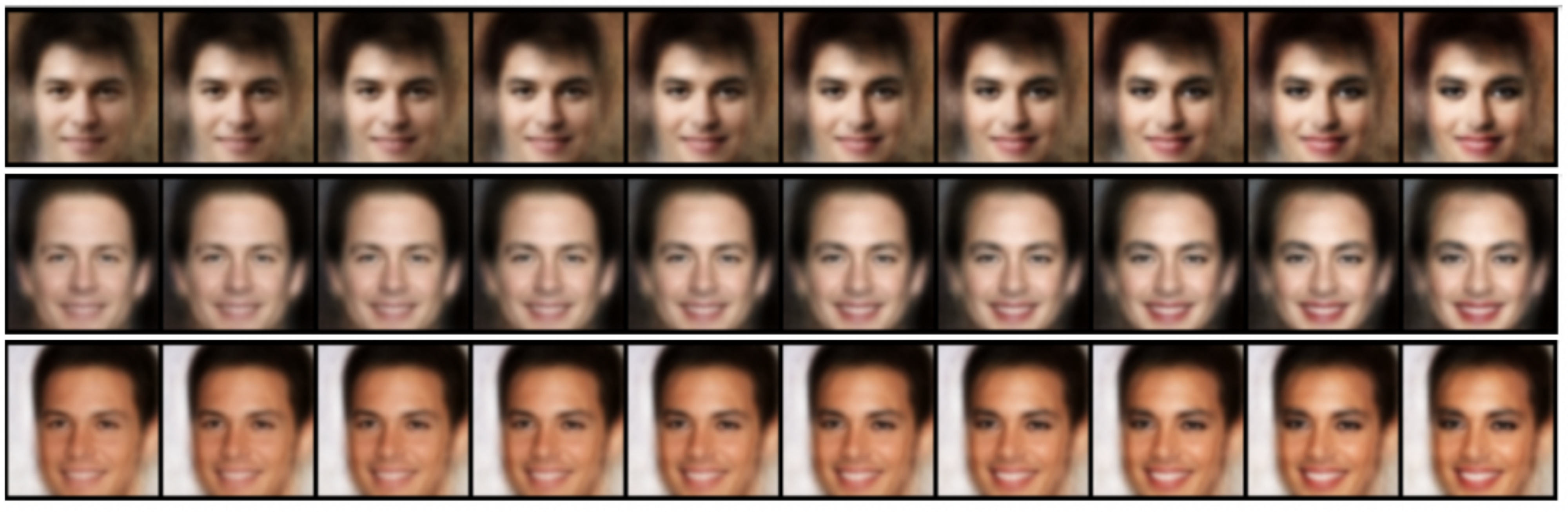} \\
			\label{fig:intervene2}
		\end{minipage}
	}
 \caption{Sample from interventional distributions. } 
 \label{fig:counteractual}
\end{center}
\end{figure}

InSection \ref{sec:Causal Disentangled Representations}, we describe the capability of our model to perform interventions by generating new images that do not exist in the dataset. Specifically, our model utilizes causal flows to sample from the interventional distributions, even though the model is trained on observational data. The steps for intervening on one factor are explained in section \ref{sec:Causal Disentangled Representations}, and the same applies to intervening on multiple factors. As depicted in Figure \ref{fig:intervene1}, we intervene on the values of two factors by fixing gender as female and gradually adjusting the value of receding hairline. This produces a series of images showing women with a gradually receding hairline. Furthermore, as shown in Figure \ref{fig:intervene2}, we intervene on gender and makeup, generating a series of images of men with gradually applied makeup. These images are not commonly found or may not even exist in the training data, highlighting the ability of our model to sample from interventional distributions.
% \begin{figure}[b]
% \centering
% \subfigure[Original $A$]{
% \begin{minipage}[t]{0.2\linewidth}
% \centering
% \includegraphics[width=\textwidth]{pendulum-prior-ao-a-0.pdf}
% % \caption{fig1}
% \label{fig:pendulumprioraoa0-appendix}
% \end{minipage}%
% }%
% \subfigure[After 20 epochs]{
% \begin{minipage}[t]{0.2\linewidth}
% \centering
% \includegraphics[width=\textwidth]{pendulum-prior-ao-a-20.pdf}
% %\caption{fig2}
% \label{fig:pendulumprioraoa20-appendix}
% \end{minipage}%
% }%
% \subfigure[After 50 epochs]{
% \begin{minipage}[t]{0.2\linewidth}
% \centering
% \includegraphics[width=\textwidth]{pendulum-prior-ao-a-50.pdf}
% %\caption{fig2}
% \label{fig:pendulumprioraoa50-appendix}
% \end{minipage}%
% }%
% \subfigure[After 80 epochs ]{
% \begin{minipage}[t]{0.2\linewidth}
% \centering
% \includegraphics[width=\textwidth]{pendulum-prior-ao-a-80.pdf}
% %\caption{fig2}
% \label{fig:pendulum-prior-ao-a-80-appendix}
% \end{minipage}%
% }%
% \subfigure[Final $A$]{
% \begin{minipage}[t]{0.2\linewidth}
% \centering
% \includegraphics[width=\textwidth]{pendulum-prior-ao-a-80-filter.pdf}
% %\caption{fig2}
% \label{fig:pendulumprioraoa80filter-appendix}
% \end{minipage}%
% }%
% \centering
%  \caption{The learned weighted adjacency matrix $A$ given the causal ordering on Pendulum. (a)-(d) illustrate the changes in $A$ as the training progresses. (e) represents $A$ after edge pruning.
% }
% \label{fig:A_pendulum_appendix}
% \end{figure}
\begin{figure}[b]
\centering
\subfigure[Original $A$]{
\begin{minipage}[t]{0.2\linewidth}
\centering
\includegraphics[width=\textwidth]{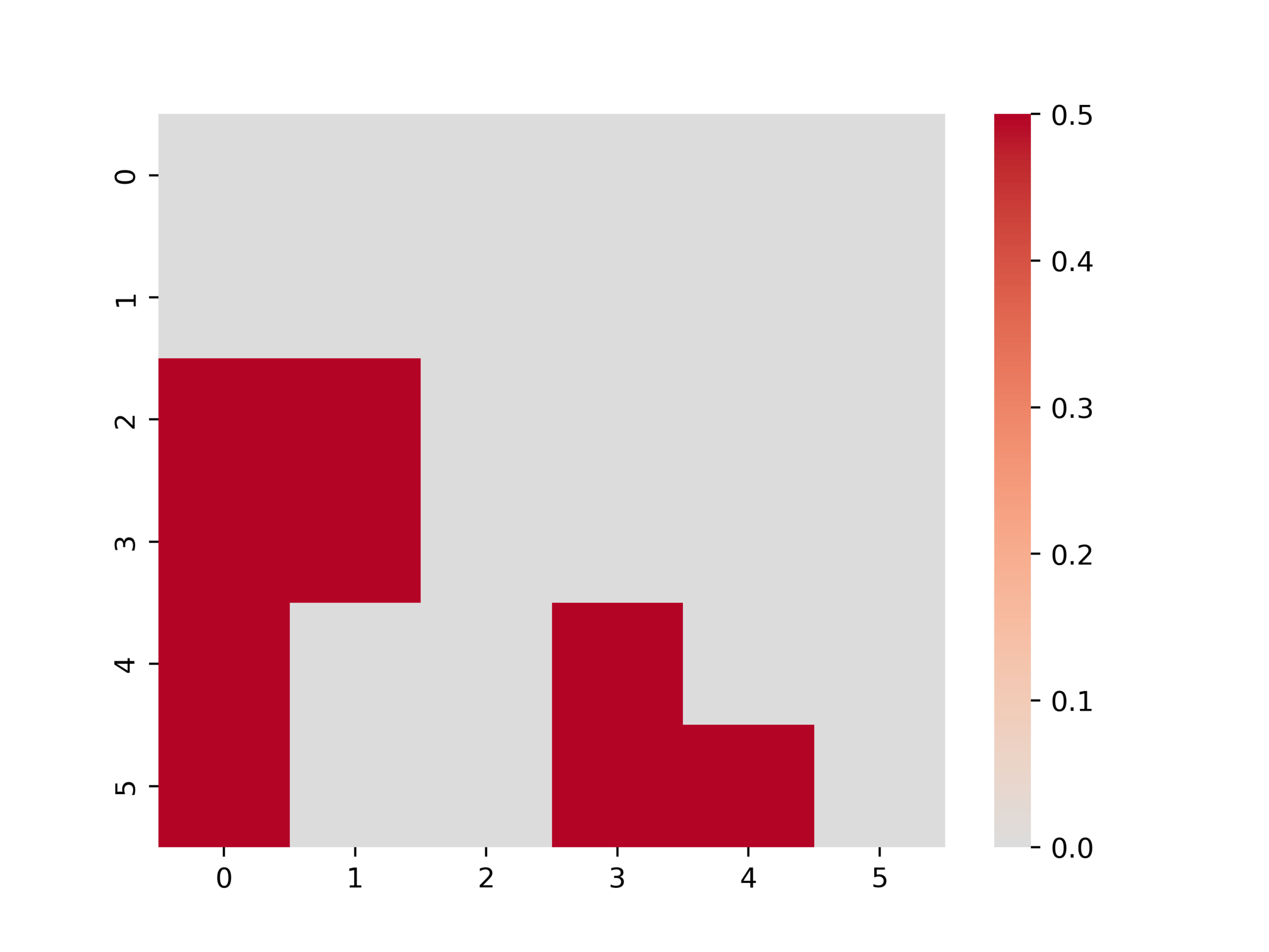}
% \caption{fig1}
\label{fig:attractive-prior-ao-a-0}
\end{minipage}%
}%
\subfigure[After 5 epochs]{
\begin{minipage}[t]{0.2\linewidth}
\centering
\includegraphics[width=\textwidth]{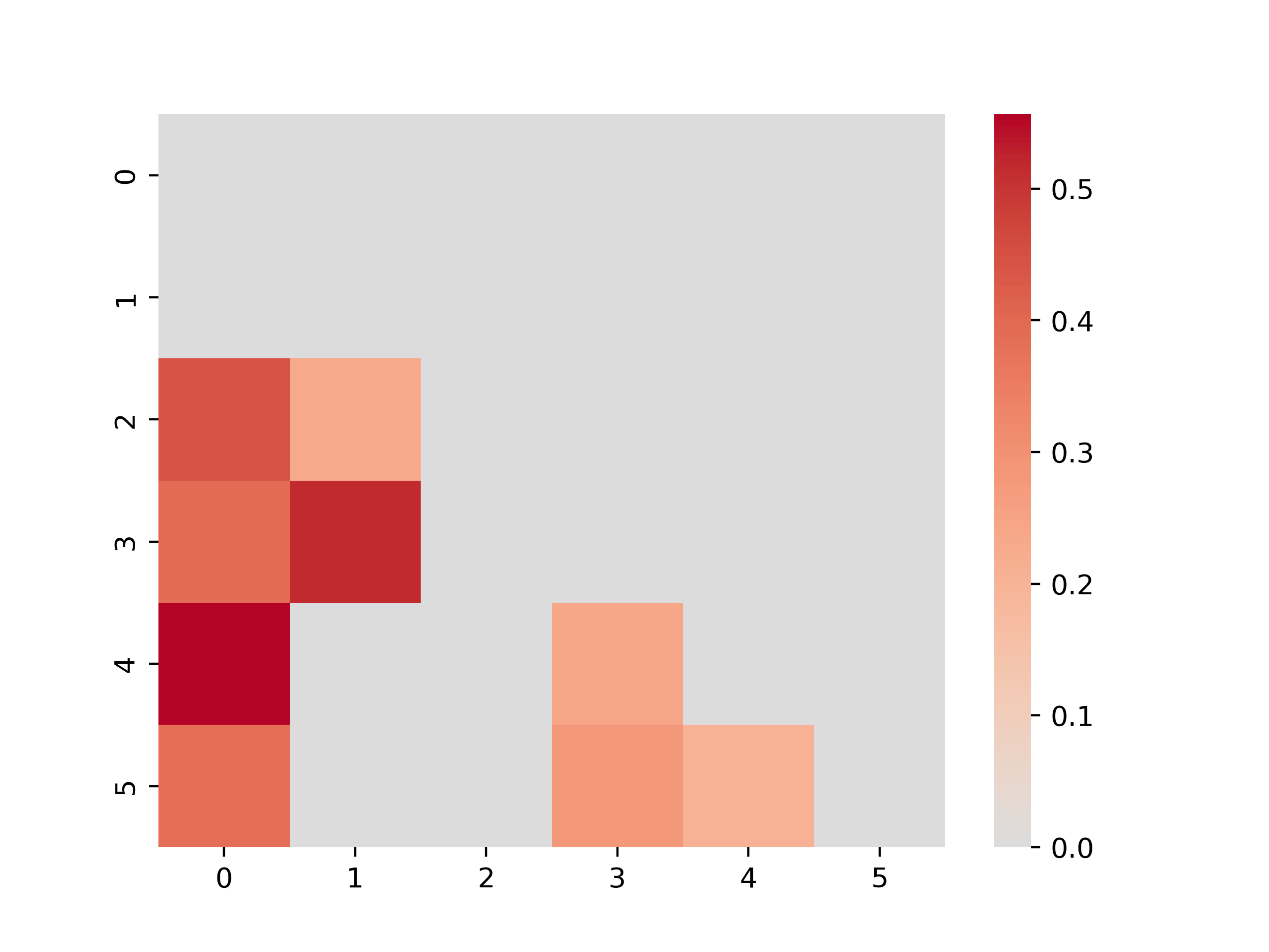}
%\caption{fig2}
\label{fig:attractive-prior-ao-a-5}
\end{minipage}%
}%
\subfigure[After 20 epochs]{
\begin{minipage}[t]{0.2\linewidth}
\centering
\includegraphics[width=\textwidth]{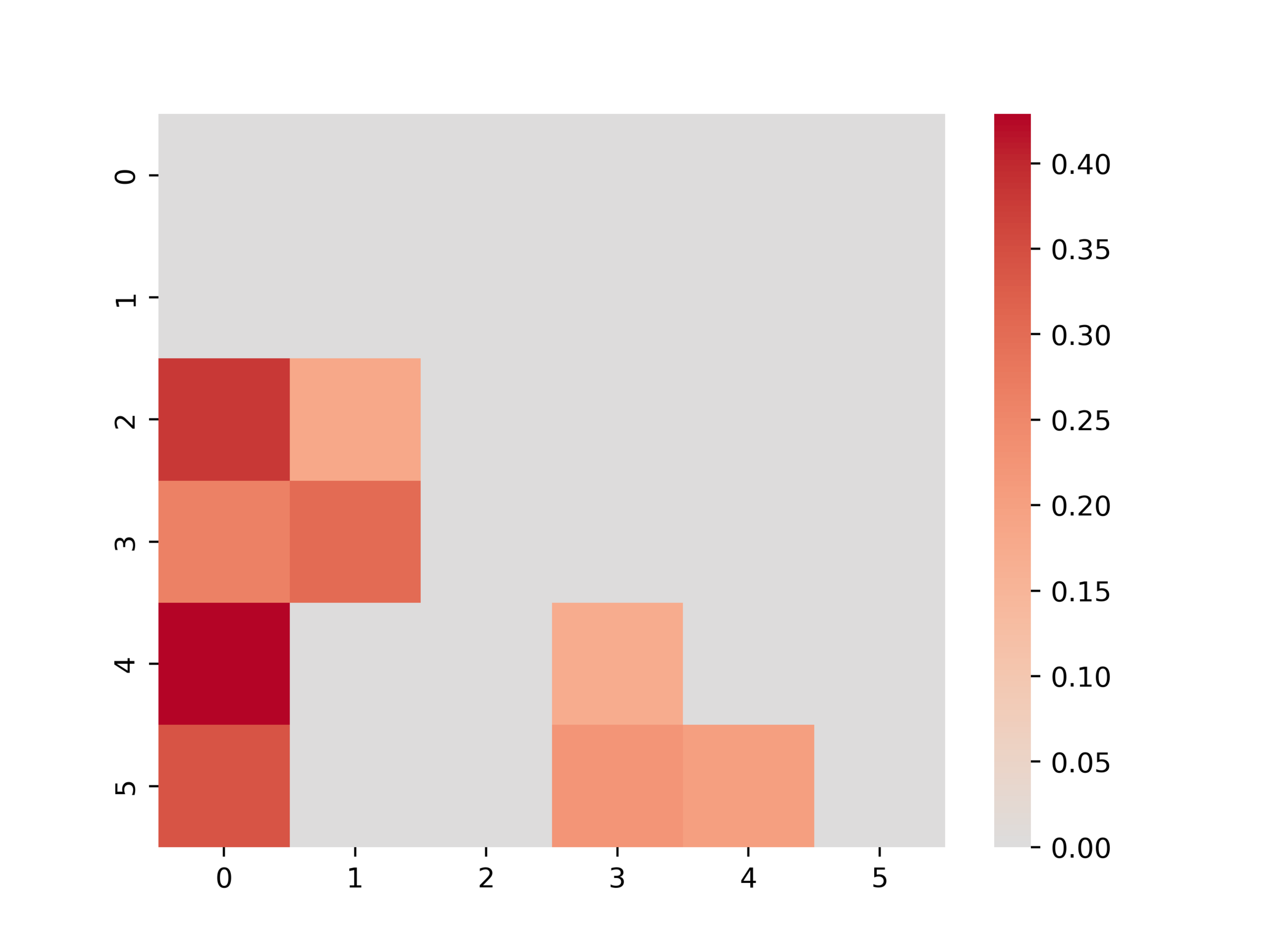}
%\caption{fig2}
\label{fig:attractive-prior-ao-a-20}
\end{minipage}%
}%
\subfigure[After 120 epochs]{
\begin{minipage}[t]{0.2\linewidth}
\centering
\includegraphics[width=\textwidth]{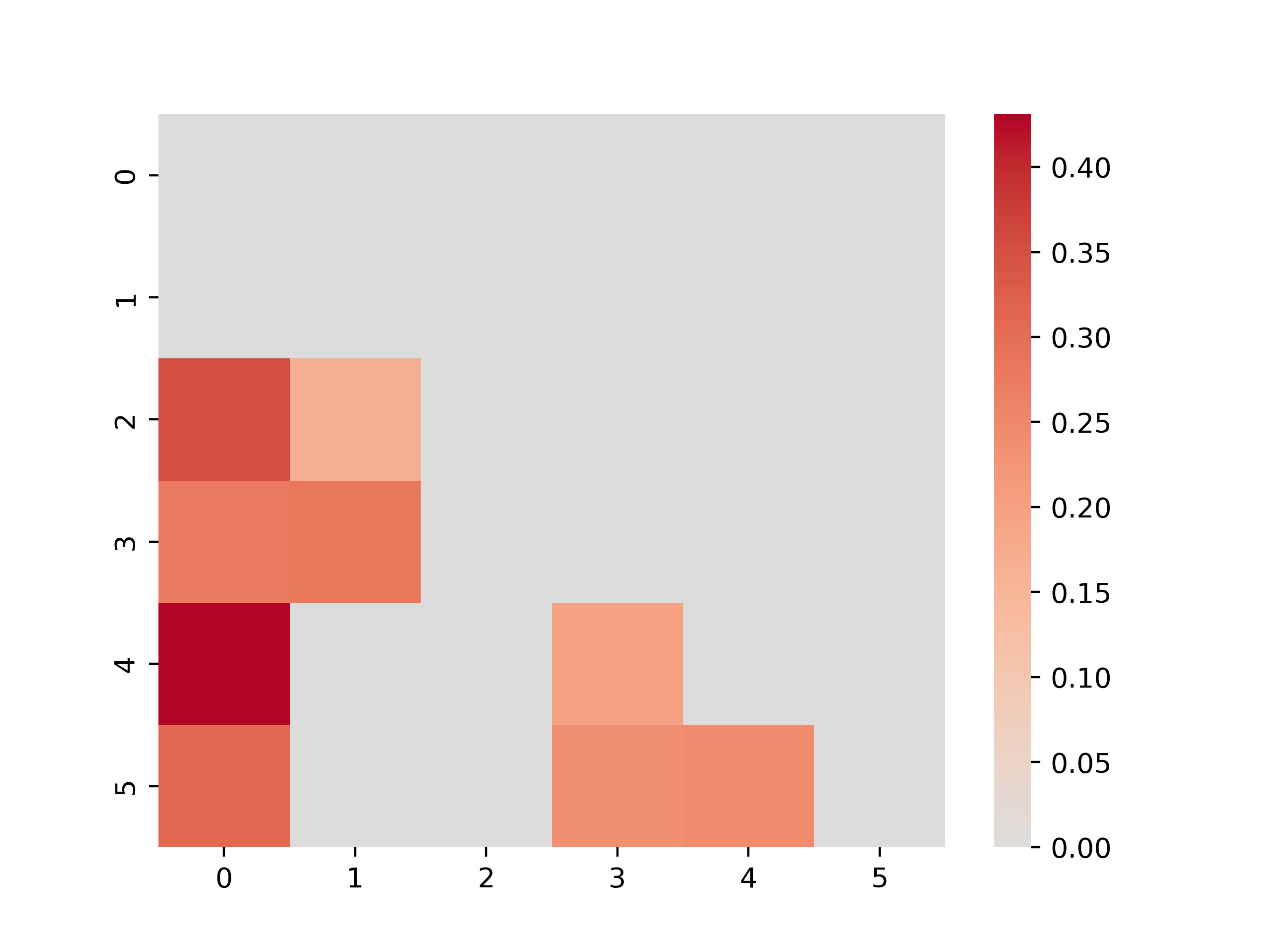}
%\caption{fig2}
\label{fig:attractive-prior-ao-a-120}
\end{minipage}%
}%
\subfigure[Final $A$]{
\begin{minipage}[t]{0.2\linewidth}
\centering
\includegraphics[width=\textwidth]{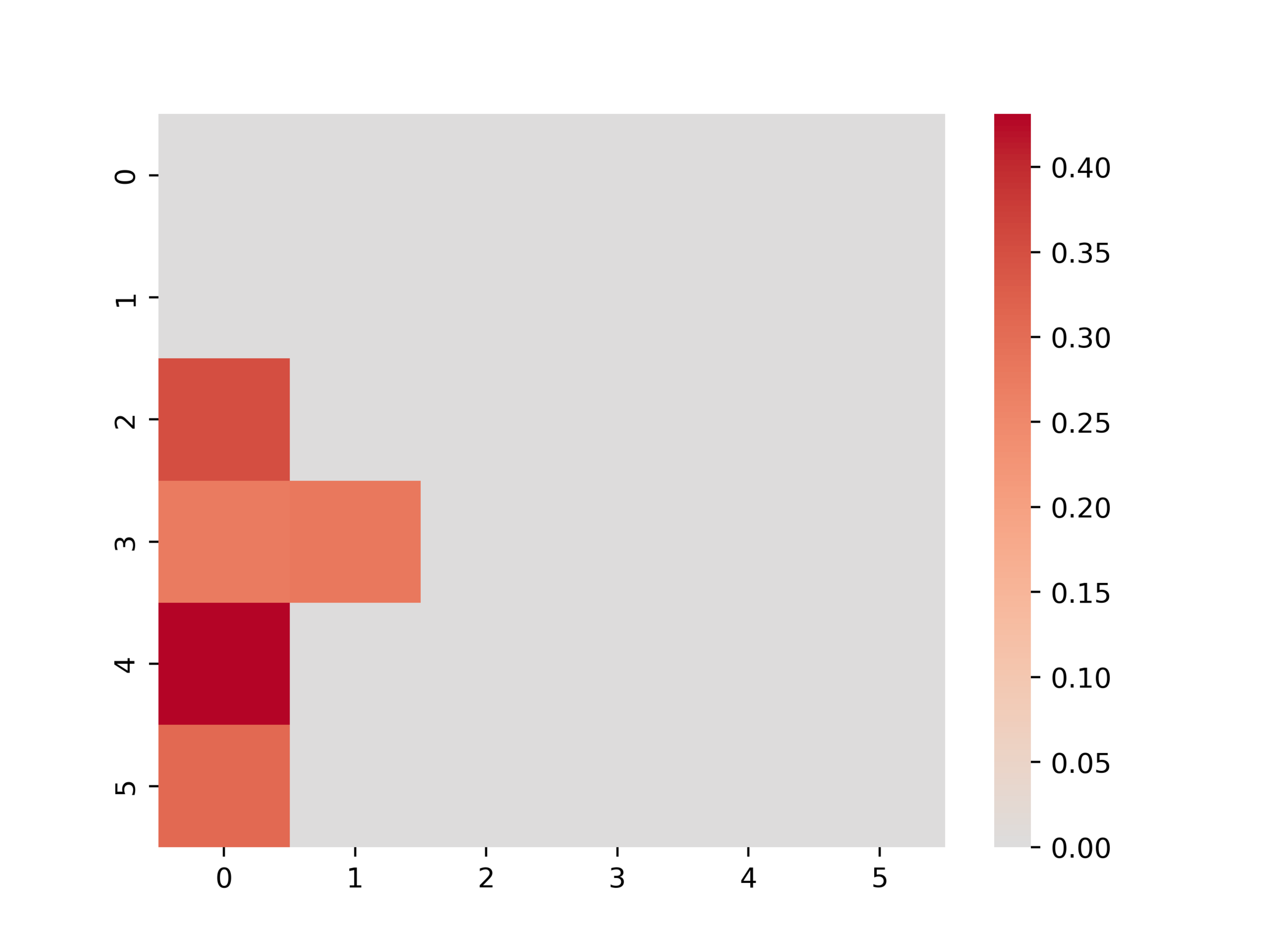}
%\caption{fig2}
\label{fig:attractive-prior-ao-a-120-filter}
\end{minipage}%
}%
\centering
 \caption{The learned weighted adjacency matrix $A$ given a super-graph on CelebA(Attractive). (a)-(d) illustrate the changes in $A$ as the training progresses. (e) represents $A$ after edge pruning.
}
\label{fig:A_attractive_appendix}
\end{figure}
\subsection{Learning of causal structure} DCVAE has the potential to learn causal structure between underlying factors, even though our work does not specifically focus on this aspect, and it does not use the Structural Causal Model (SCM) framework. Here we present the learning process of the adjacency weight matrix $A$, whose super-graph is shown in Figure \ref{fig:super_graph_all_text}. Figure \ref{fig:A_attractive_appendix} shows the learning process of CelebA (Attractive). If we set a threshold of 0.25, i.e., considering edges in the causal graph smaller than the threshold as non-existing, we can obtain Figure \ref{fig:attractive-prior-ao-a-120-filter}. %We find that the final result differs from the true causal structure by one edge, but this also indicates that causal discovery can be further improved through future design.
Our observation indicates that DCVAE effectively eliminates a significant portion of redundant variables and ultimately achieves a nearly accurate graph. This suggests the potential for further improvements in causal discovery through future research work.
\subsection{Examining Causal Disentangled Representations}
\begin{figure}[t]
\centering
\subfigure[Traverse of DCVAE on CelebA(Attractive)]{
\begin{minipage}[t]{0.5\linewidth}
\centering
\includegraphics[width=\textwidth]{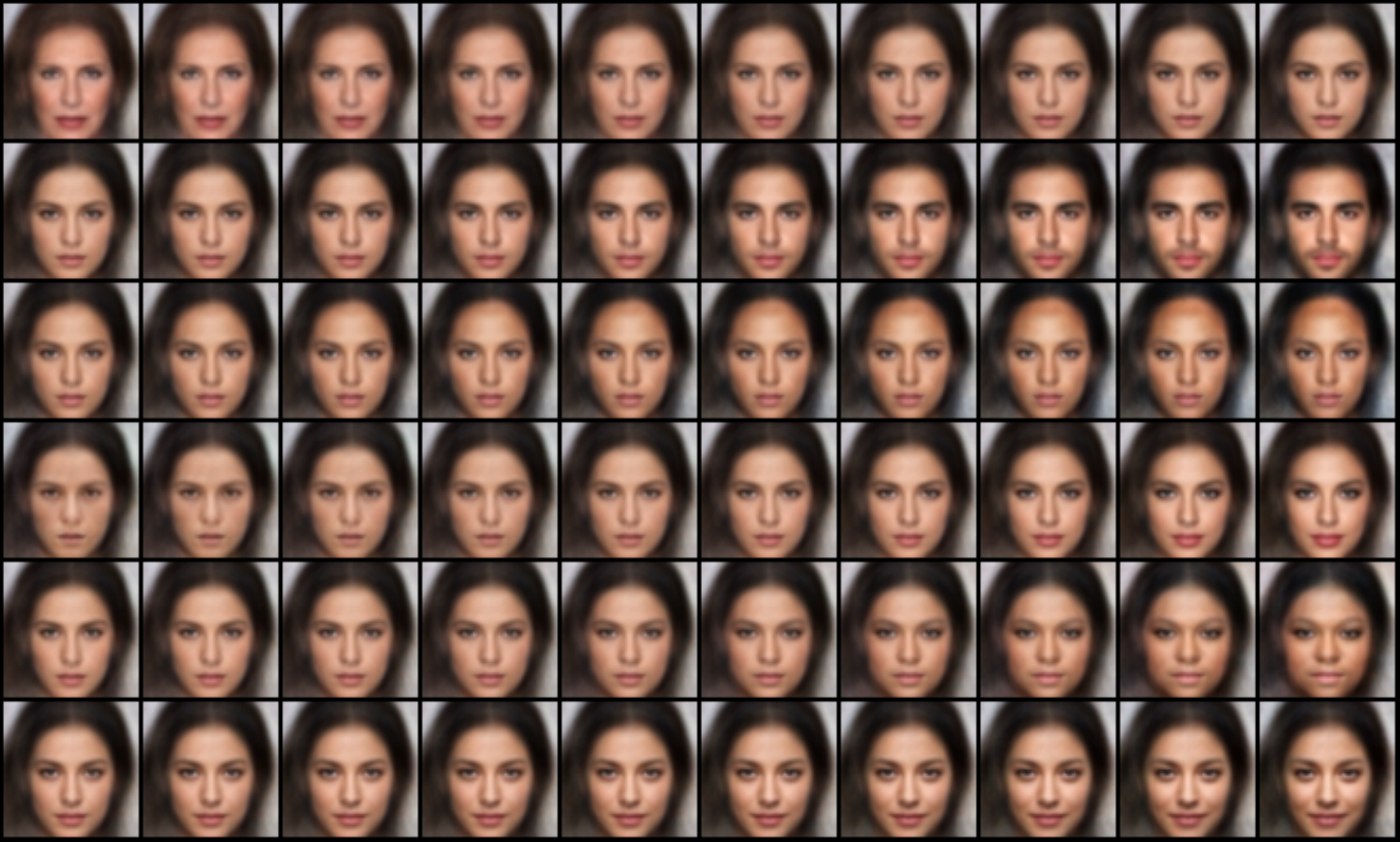}
% \caption{fig1}
\label{fig:Traverse of DCVAE on CelebA(Attractive)}
\end{minipage}%
}%
\subfigure[Intervention of DCVAE on CelebA(Attractive)]{
\begin{minipage}[t]{0.5\linewidth}
\centering
\includegraphics[width=\textwidth]{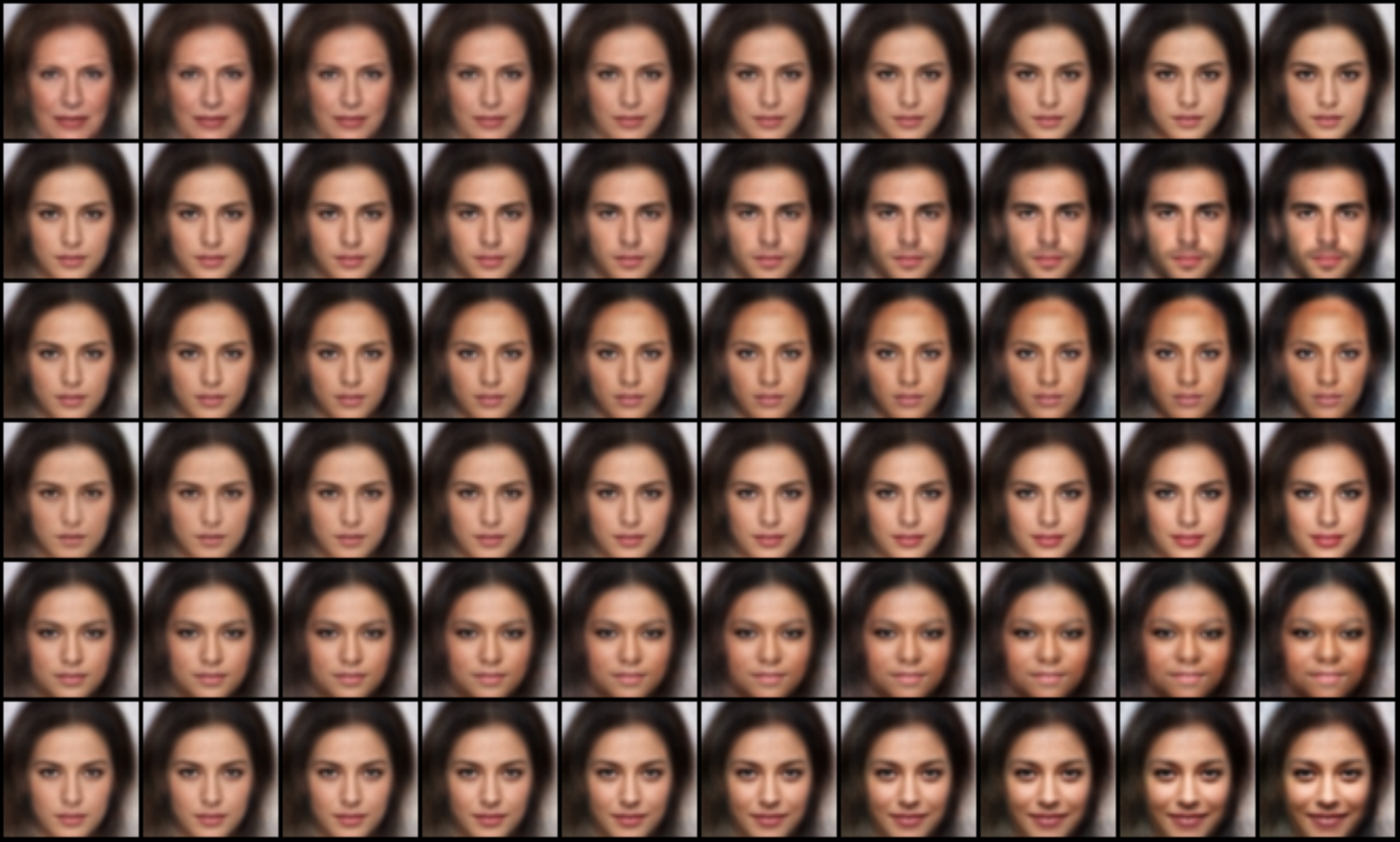}
%\caption{fig2}
\label{fig:Intervention of DCVAE on CelebA(Attractive)}
\end{minipage}%
}%
\centering
\caption{Results of the DCVAE model under two types of interventions on CelebA(Attractive). Each row corresponds to one factor, in the same order as in Figure \ref{fig:attractivegraph}. We observe that our model achieves disentanglement, and when intervening on the causal variables, it affects the effect variables, while the opposite is not true.
}
\label{fig:DCVAE on CelebA(Attractive)}
\end{figure}

\begin{figure}[t]
\centering
%\vspace{-0.35cm} %设置与上面正文的距离
	\subfigure[DEAR
]{
		\begin{minipage}{\textwidth} %[b]%{0.2\textwidth} 
                        \centering          
                        \includegraphics[width=0.7\textwidth]{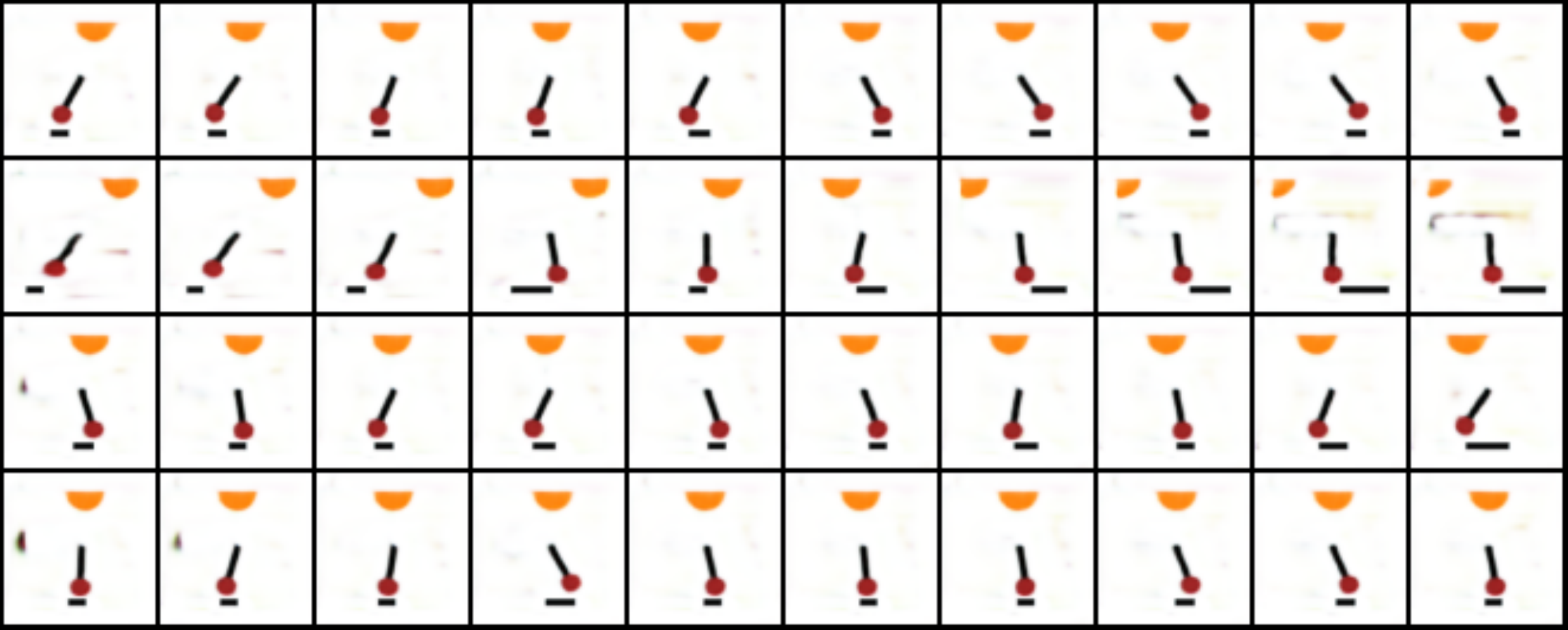} \\
                        \label{fig:dear-400-pendulum-traverse}
		\end{minipage}
	}
	\subfigure[$\beta$-VAE]{
		\begin{minipage}{\textwidth}%[b]%{0.2\textwidth}
  \centering   
			\includegraphics[width=0.7\textwidth]{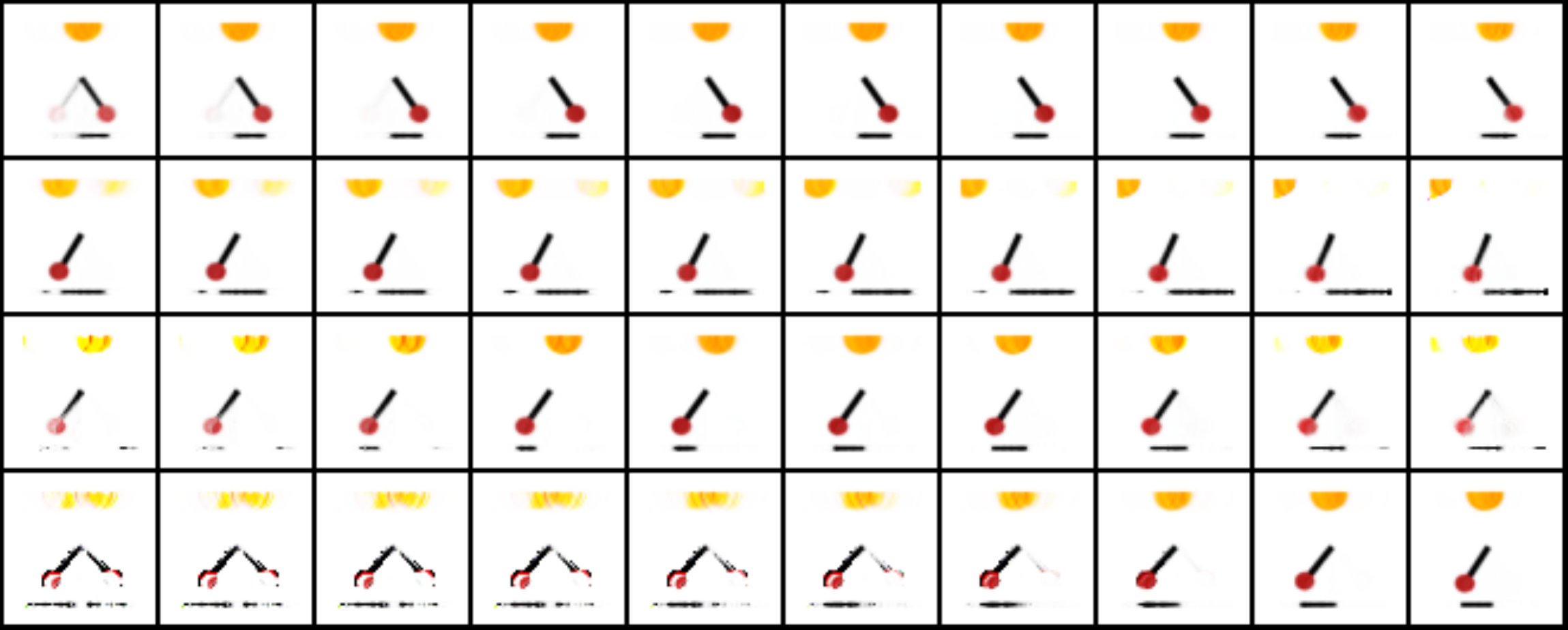} \\
			\label{fig:sbetavae-500-pendulum-traverse}
		\end{minipage}
	}
 \subfigure[$\beta$-TCVAE]{
		\begin{minipage}{\textwidth}%[b]%{0.2\textwidth}
  \centering   
			\includegraphics[width=0.7\textwidth]{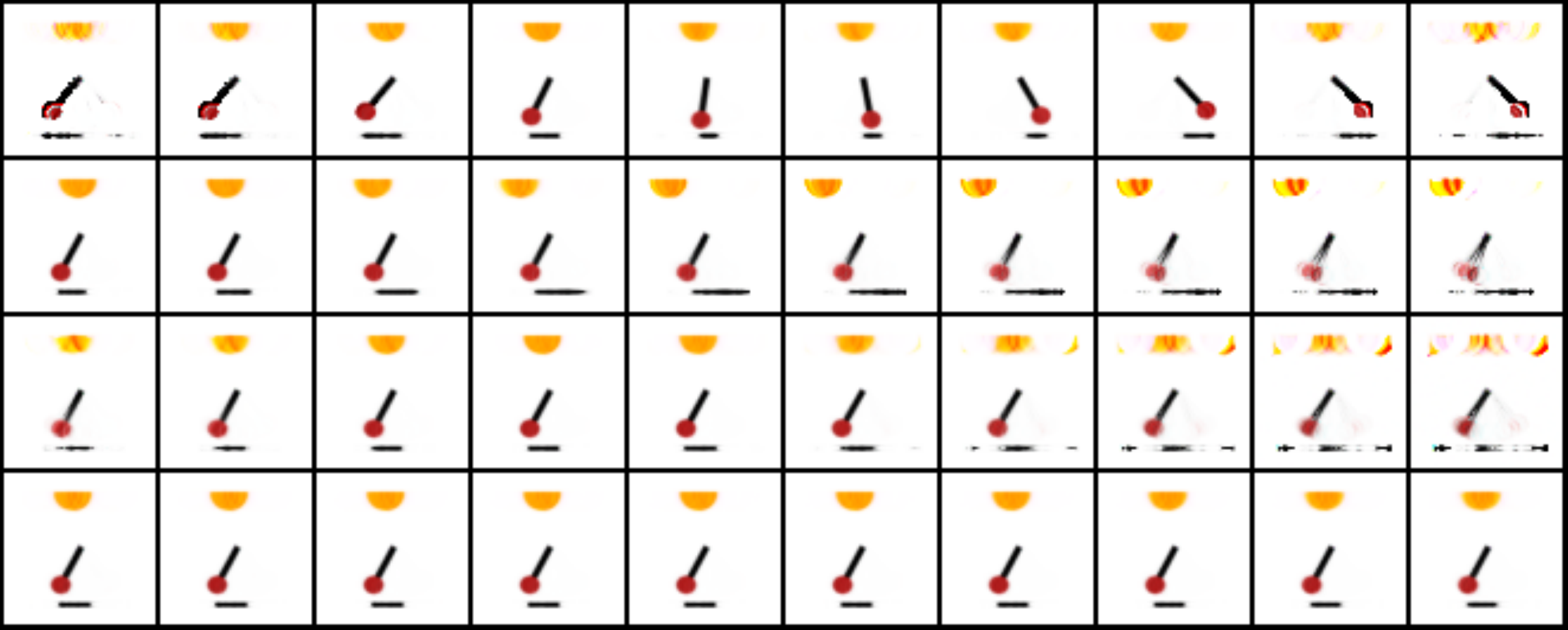} \\
			\label{fig:stcvae-500-pendulum-traverse}
		\end{minipage}
	}
 \subfigure[VAE]{
		\begin{minipage}{\textwidth}%[b]%{0.2\textwidth}
  \centering   
			\includegraphics[width=0.7\textwidth]{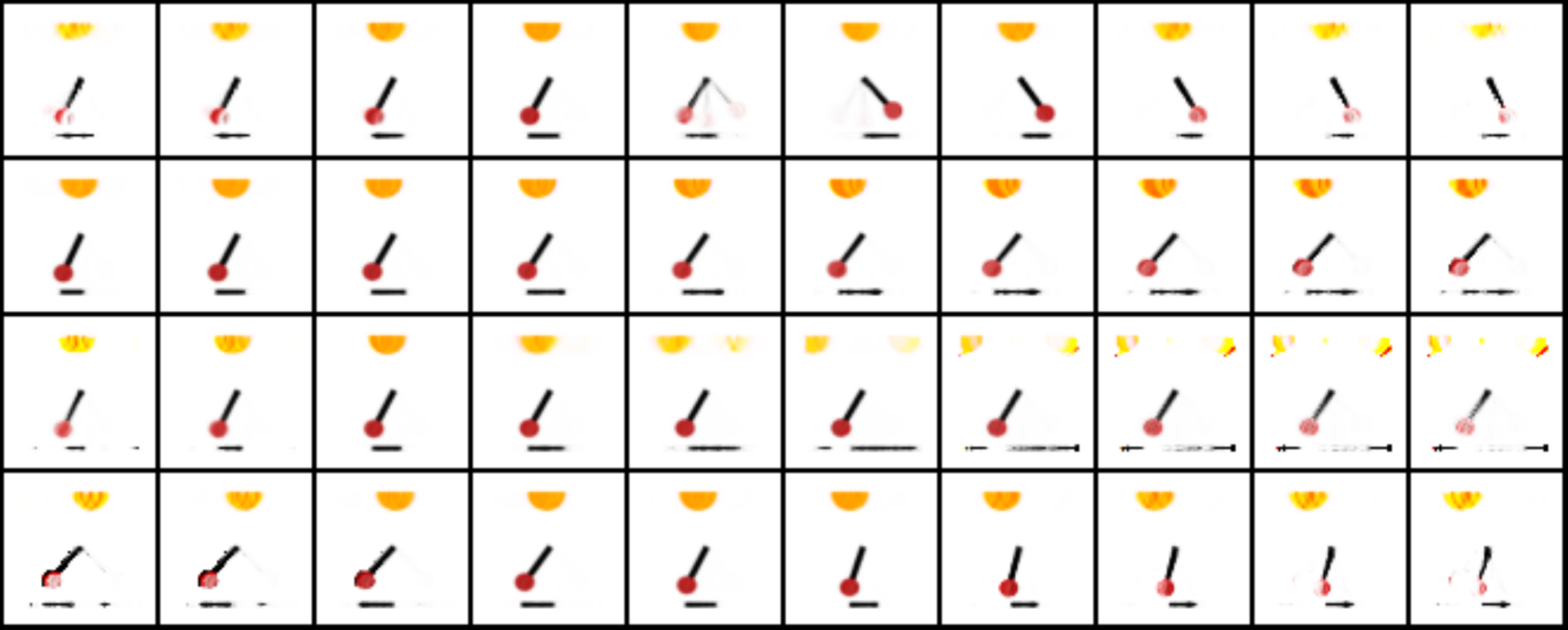} \\
			\label{fig:svae-500-pendulum-traverse}
		\end{minipage}
	}
 \centering
 \caption{Traverse results of four baseline models on Pendulum. We observe that changing one factor may result in changes in multiple factors, or no changes in any factor, such as the shadow length in the $\beta$-TCVAE. Therefore, their representations are all entangled on Pendulum.} 
 \label{fig:pendulum-traverse}
\end{figure}

\begin{figure}[t]
\centering
%\vspace{-0.35cm} %设置与上面正文的距离
	\subfigure[DEAR
]{
		\begin{minipage}{0.45\linewidth} %[b]%{0.2\textwidth} 
                        \centering          
                        \includegraphics[width=\textwidth]{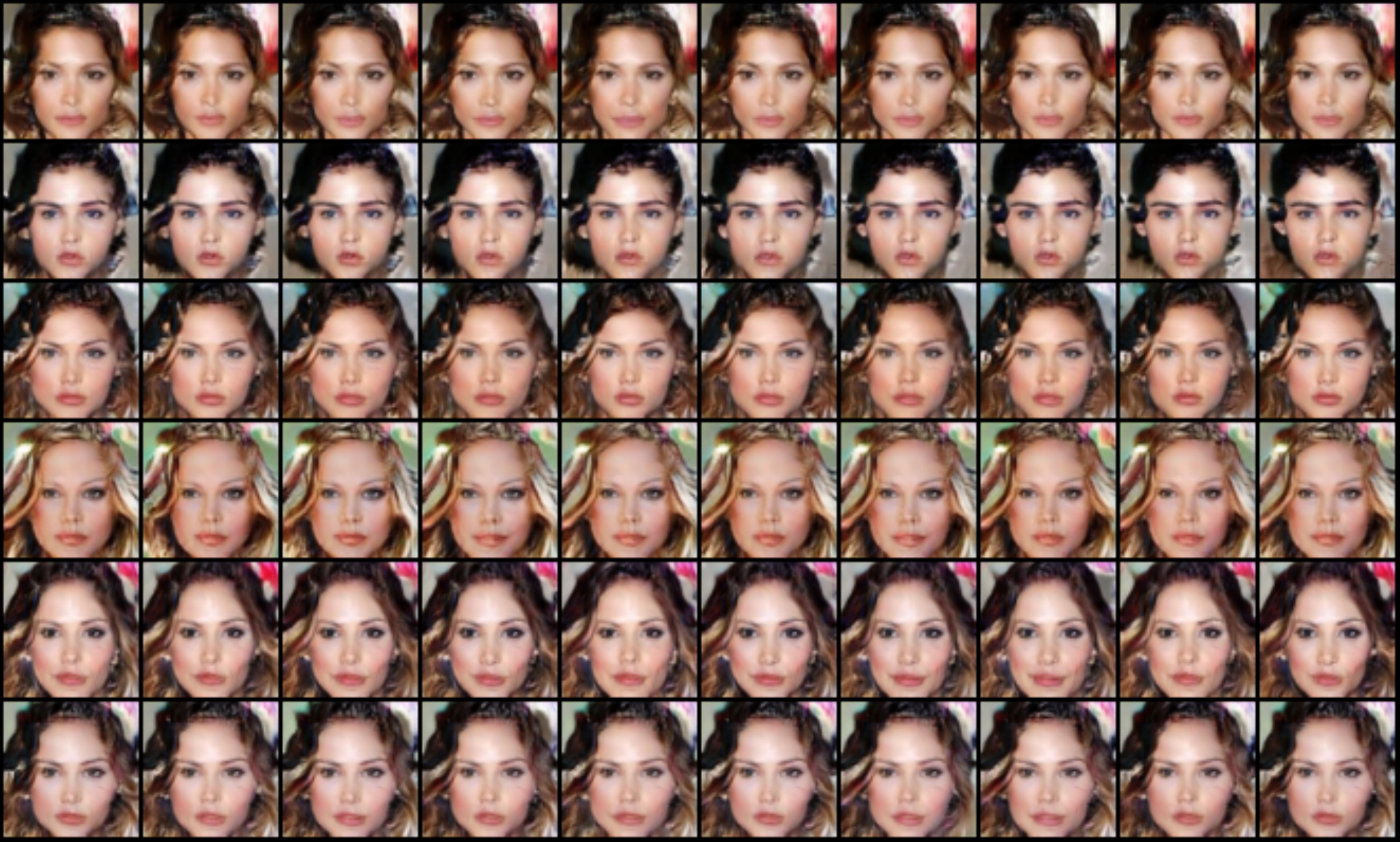} \\
                        \label{fig:traverse_attractive_dear}
		\end{minipage}
	}
	\subfigure[$\beta$-VAE]{
		\begin{minipage}{0.45\textwidth}%[b]%{0.2\textwidth}
  \centering   
			\includegraphics[width=\textwidth]{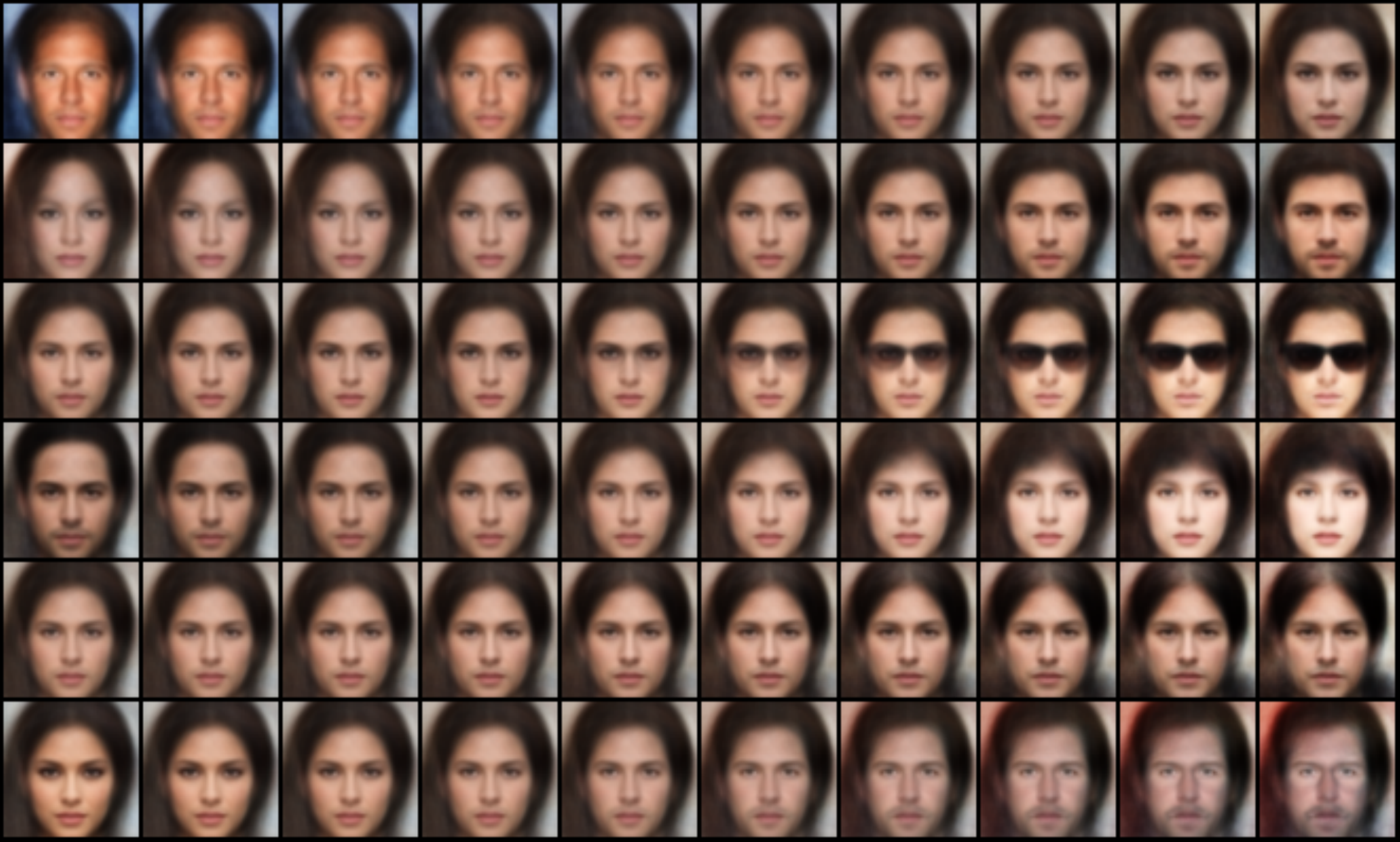} \\
			\label{fig:traverse_attractive_sbetavae}
		\end{minipage}
	}
 \subfigure[$\beta$-TCVAE]{
		\begin{minipage}{0.45\textwidth}%[b]%{0.2\textwidth}
  \centering   
			\includegraphics[width=\textwidth]{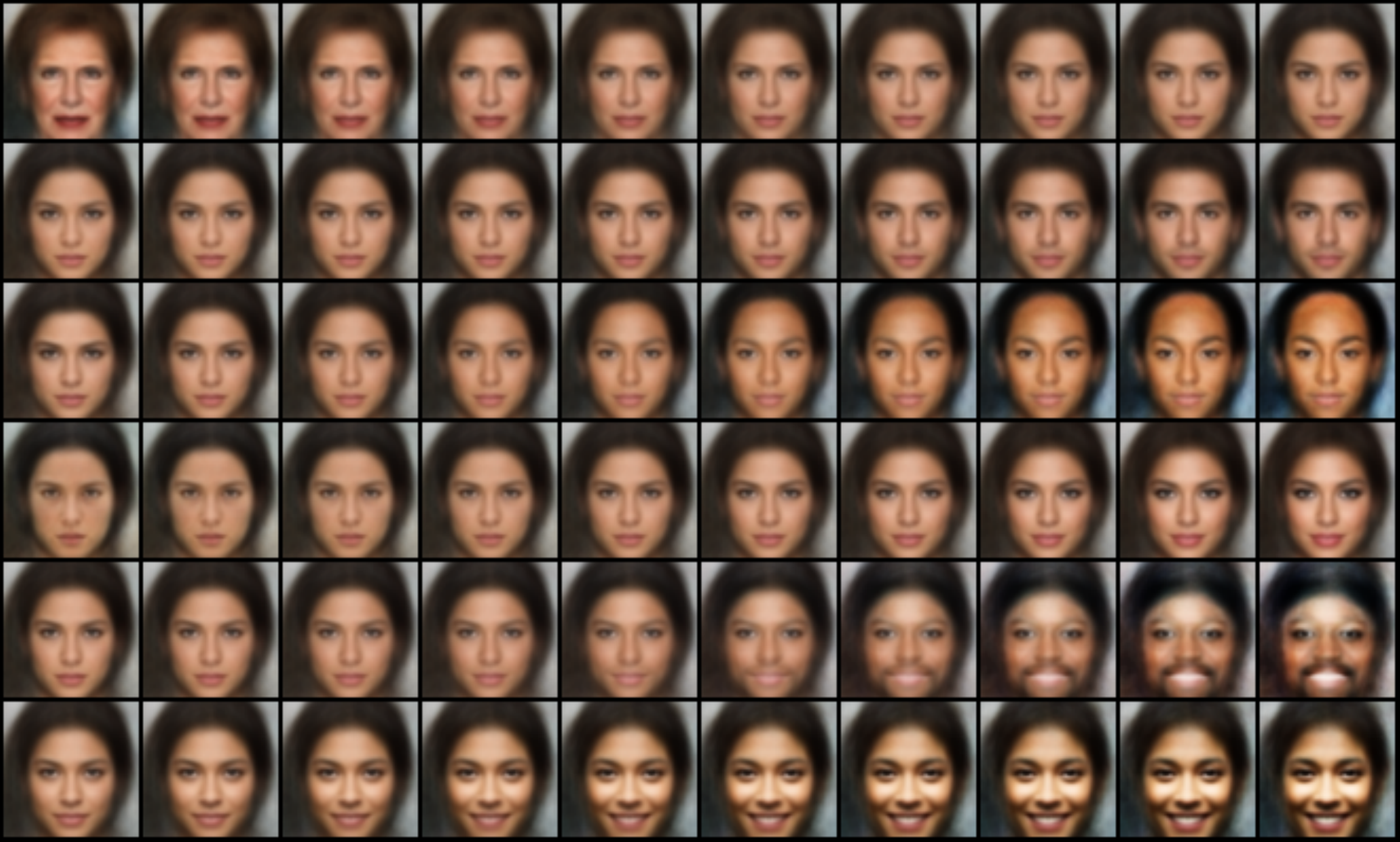} \\
			\label{traverse_attractive_stcvae}
		\end{minipage}
	}
 \subfigure[VAE]{
		\begin{minipage}{0.45\textwidth}%[b]%{0.2\textwidth}
  \centering   
			\includegraphics[width=\textwidth]{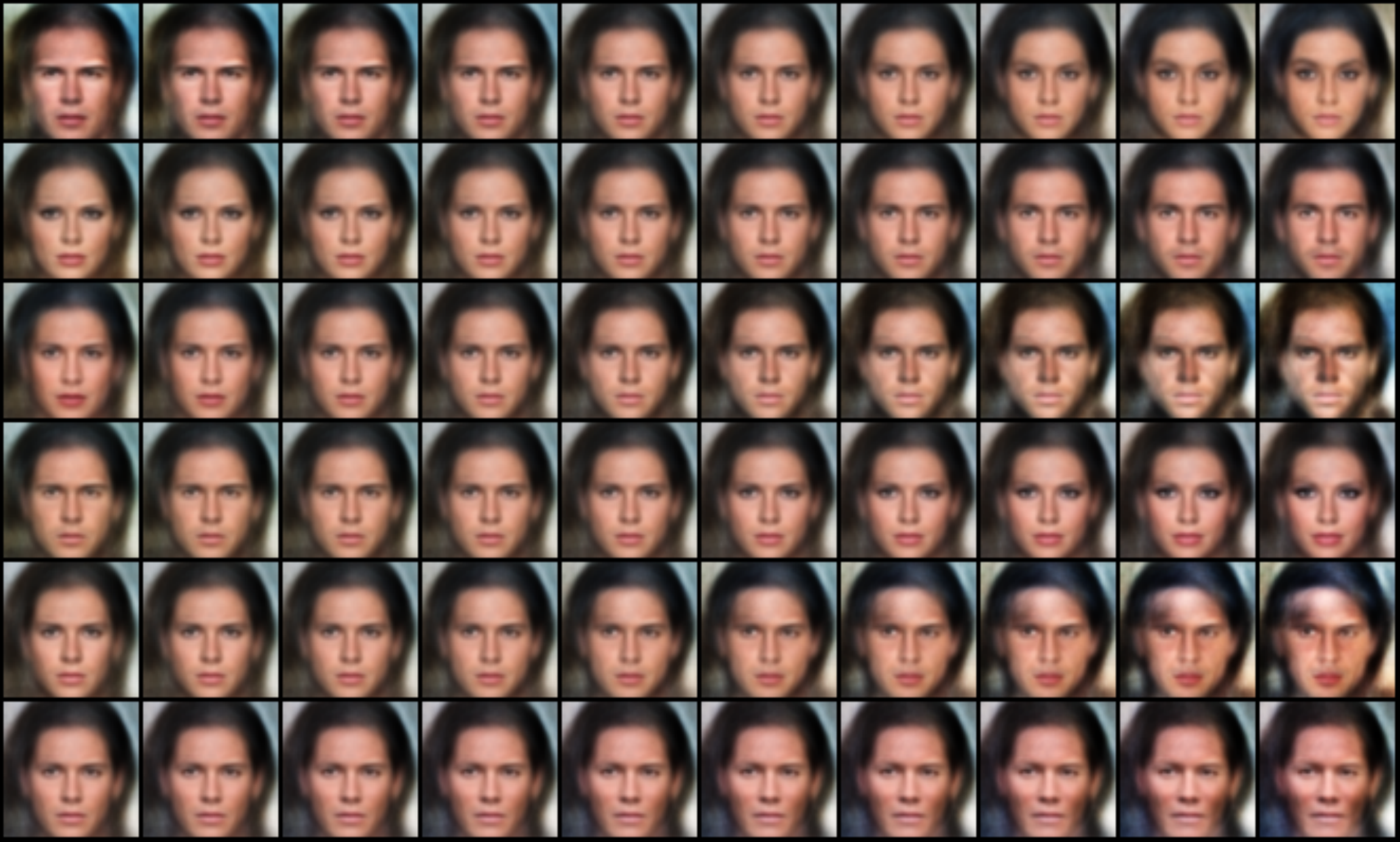} \\
			\label{fig:traverse_attractive_svae}
		\end{minipage}
	}
 \centering
 \caption{Traverse results of four baseline models on CelebA(Attractive).} 
 \label{fig:attractive-traverse-2}
\end{figure}

\begin{figure}[t]
\centering
%\vspace{-0.35cm} %设置与上面正文的距离
	\subfigure[DEAR
]{
		\begin{minipage}{0.45\linewidth} %[b]%{0.2\textwidth} 
                        \centering          
                        \includegraphics[width=\textwidth]{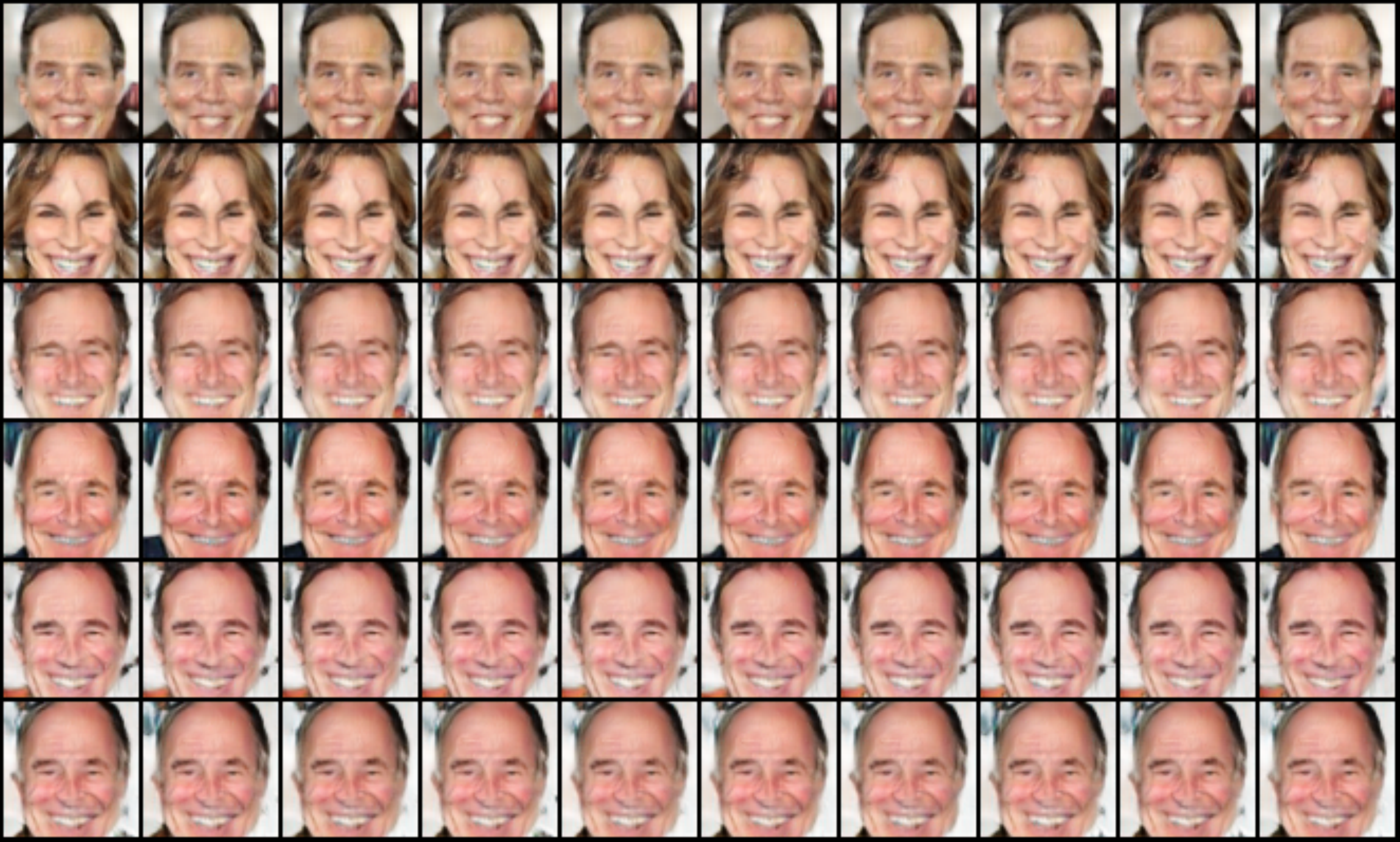} \\
                        \label{fig:traverse_smile_dear}
		\end{minipage}
	}
	\subfigure[$\beta$-VAE]{
		\begin{minipage}{0.45\linewidth}%[b]%{0.2\textwidth}
  \centering   
			\includegraphics[width=\textwidth]{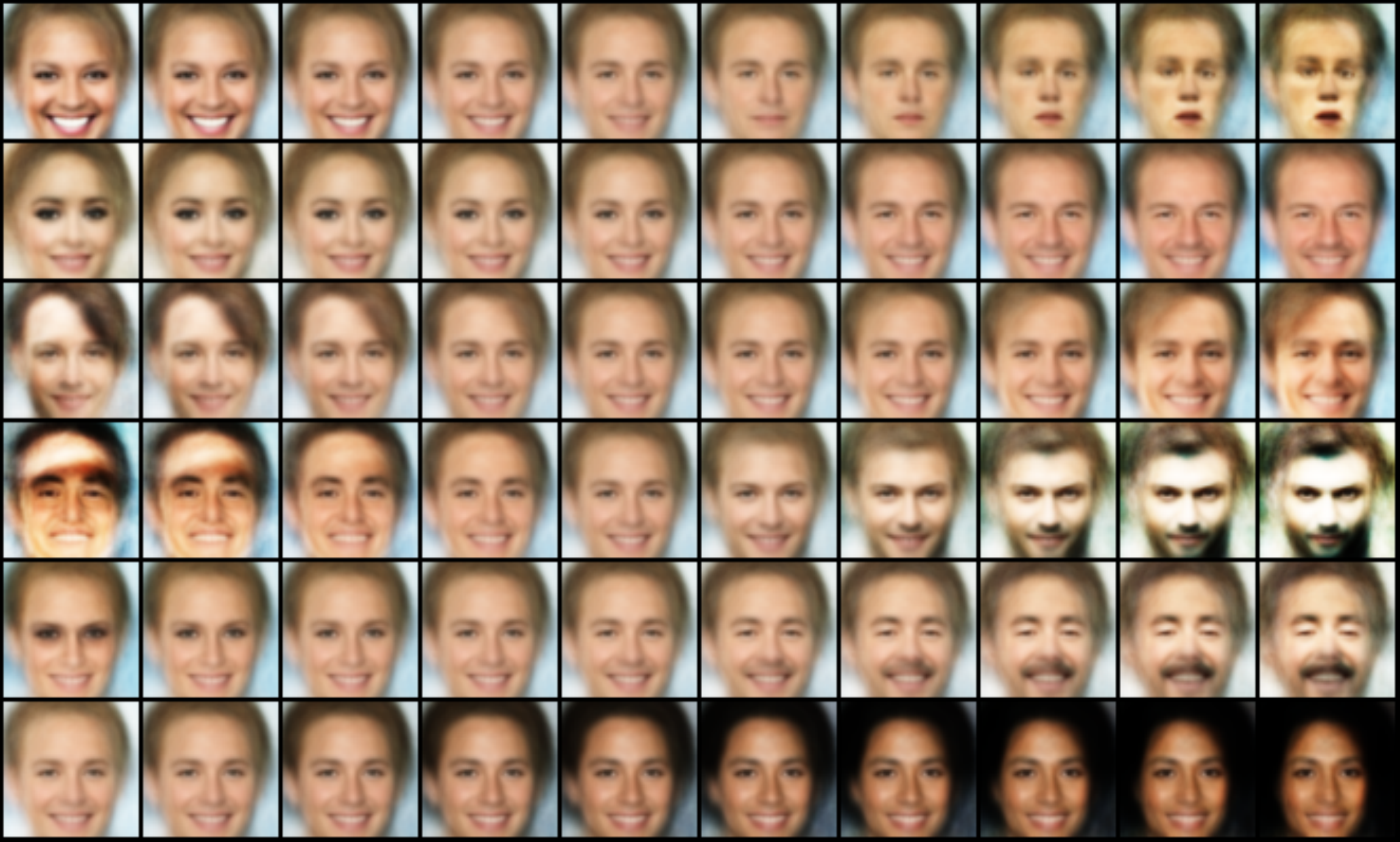} \\
			\label{fig:traverse_smile_sbvae}
		\end{minipage}
	}
 \subfigure[$\beta$-TCVAE]{
		\begin{minipage}{0.45\linewidth}%[b]%{0.2\textwidth}
  \centering   
			\includegraphics[width=\textwidth]{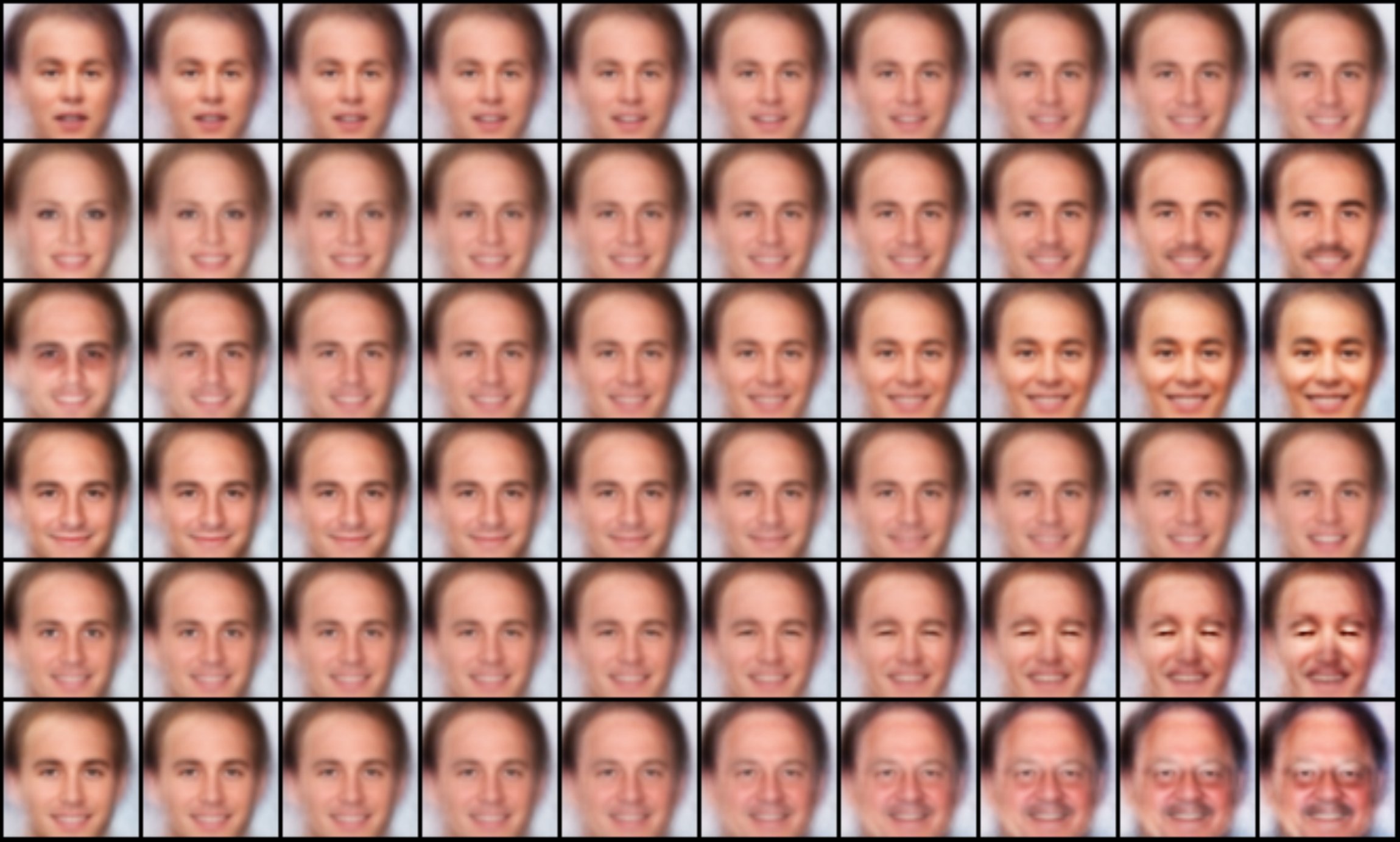} \\
			\label{traverse_smile_stcvae}
		\end{minipage}
	}
 \subfigure[VAE]{
		\begin{minipage}{0.45\linewidth}%[b]%{0.2\textwidth}
  \centering   
			\includegraphics[width=\textwidth]{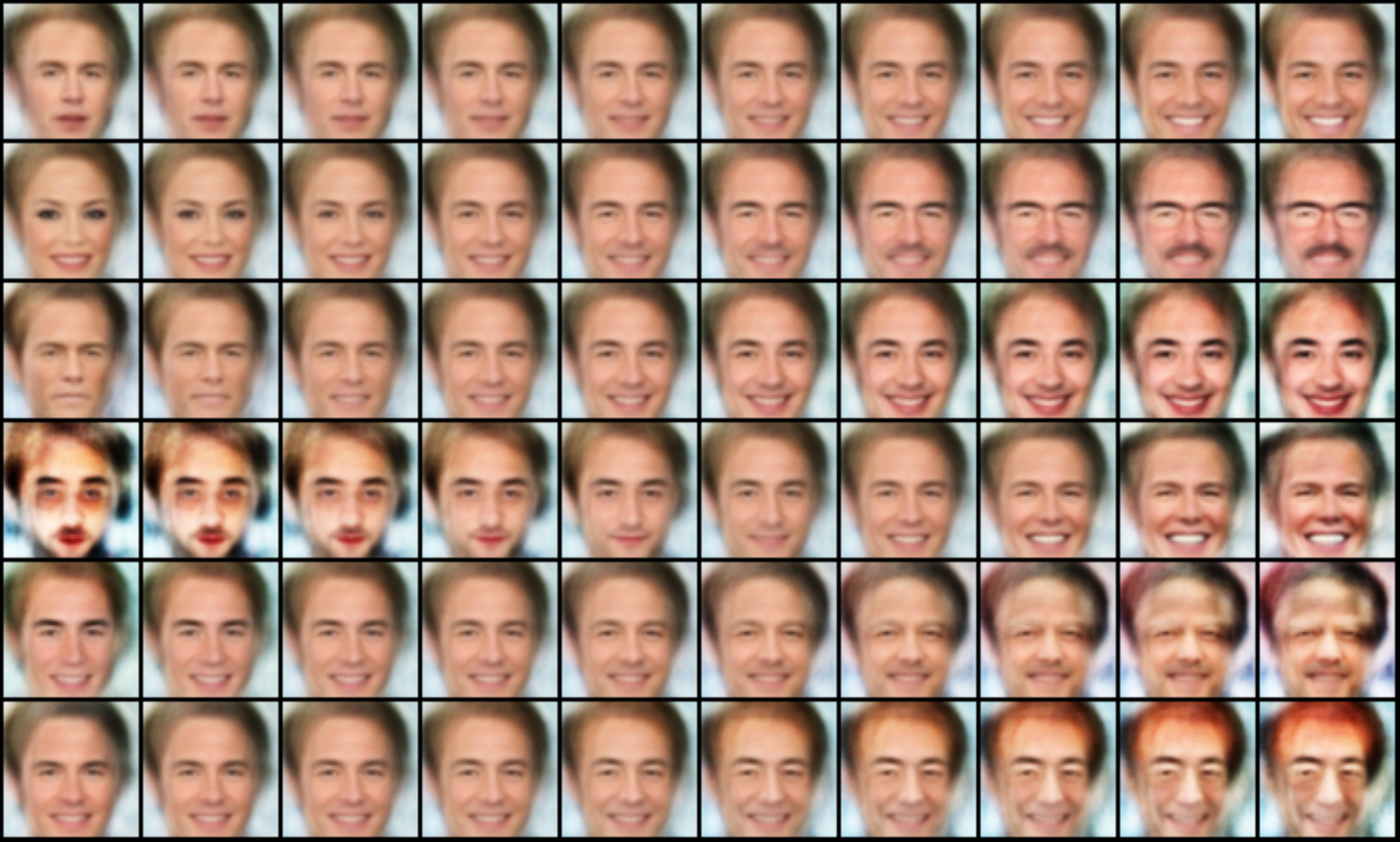} \\
			\label{fig:traverse_smile_svae}
		\end{minipage}
	}
 \centering
 \caption{Traverse results of four baseline models on CelebA(Smile).} 
 \label{fig:smile-traverse-2}
\end{figure}

To verify whether our model has learned causal disentangled representations, we consider two types of intervention operations: the "traverse" operation and the "intervention" operation introduced in section \ref{sec:Causal Disentangled Representations}. We present the experimental results of DCVAE on CelebA(Attractive) in Figure \ref{fig:DCVAE on CelebA(Attractive)} and the results of baseline models on three datasets in Figure \ref{fig:pendulum-traverse}, \ref{fig:attractive-traverse-2}, and \ref{fig:smile-traverse-2}.

\begin{figure}[htbp]
    \centering
    \rotatebox{90}{ % Rotate the entire block of images by 90 degrees
        \begin{minipage}{\textheight} % This makes sure the width/height fits the page after rotation
            \centering
            \includegraphics[height=0.08\textwidth]{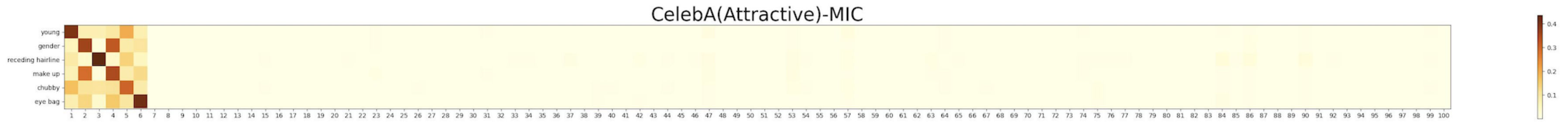} % First image
            \hfill
            \includegraphics[height=0.08\textwidth]{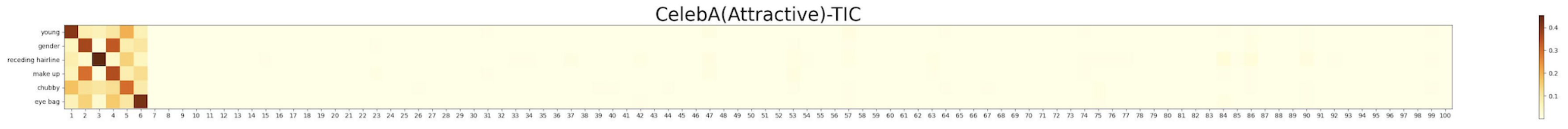} % Second image
            \hfill
            \includegraphics[height=0.079\textwidth]{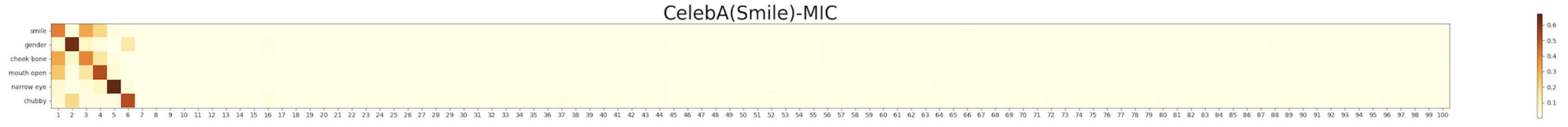} % Third image
            \hfill
            \includegraphics[height=0.079\textwidth]{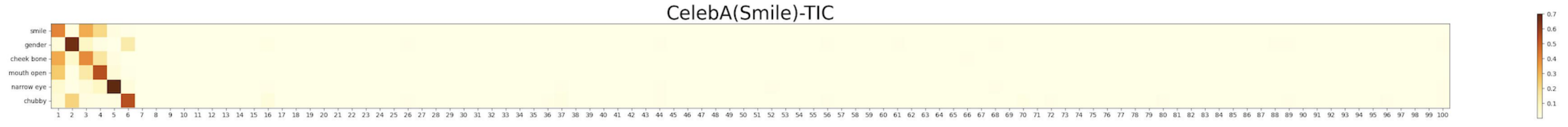} % Fourth image
            \caption{The mutual information (MIC/TIC) between the learned representation and the ground truth concept labels on the CelebA (Attractive) and CelebA (Smile) datasets.}
            \label{fig:additional information}
        \end{minipage}
    }
\end{figure}

\textcolor{black}{\subsection{Do Remaining Latent Variables Contain Useful Information about
the underlying factors we we focus on?}
We further conducted experiments to measure the influence of the remaining $m-d$ dimensions on the generated images in the CelebA dataset.
As shown in Figure \ref{fig:additional information}, we measured the maximal information coefficient (MIC) and the total information coefficient (TIC) between the learned representations and the ground truth concept labels in the CelebA (Attractive) and CelebA (Smile) datasets. In these experimental settings, with d=100 and m=6, the first six units in the latent variable layer correspond to the six factors. From the figures, it is clear that the remaining 94 latent variable units contain almost no information about the generative factors we focus on. Therefore, the nature of these remaining latent variables ensures that the generation of counterfactual images can guarantee causal disentangled representation. }

\section{Reflections on Disentanglement Metrics} \label{Appendix:Reflections on Disentanglement Metrics}
Numerous disentanglement studies propose their own metrics, such as the $\beta$-VAE \citep{higgins2017beta}, FactorVAE \citep{kim2018disentangling}, Mutual Information Gap (MIG) \citep{chen2018isolating}, and DCI \citep{eastwood2018framework}. A comprehensive review of these metrics can be found in \citet{locatello2019challenging}. However, these metrics are typically limited to scenarios where generative factors are independent and do not extend to cases with factor correlations. For instance, the MIG score evaluates the normalized gap in mutual information between the highest and second-highest coordinates in $\bar E({\mathbf x},{\mathbf u})$. In cases where $\boldsymbol{\xi}_1$ is correlated with $\boldsymbol{\xi}_2$, and a disentangled representation aligns each factor with one coordinate (i.e., $\bar E_1({\mathbf x},{\mathbf u}) = r_1(\xi_1)$ and $\bar E_2({\mathbf x},{\mathbf u}) = r_1(\xi_2)$), both coordinates will exhibit significant mutual information with their respective factors, leading to a minimal gap. Consequently, such a representation yields a low MIG score despite being properly disentangled.
Metrics for causal disentanglement remain scarce and have limitations. \citet{shen2022weakly} proposed a metric based on FactorVAE, which is the most suitable for our model compared to others like IRS \citep{suter2019robustly} and (UC and GC) \citep{reddy2022causally}, which assume conditional independence. However, \citet{kim2019relevance} demonstrated that this metric is not entirely reliable. In our experiments, FactorVAE scores \citep{shen2022weakly} for models such as DCVAE, VAE, $\beta$-TCVAE, $\beta$-VAE, and DEAR were all 0.50, with DEAR scoring 0.28. These results suggest that, at minimum, our model outperforms the latest models like DEAR in causal disentanglement, though the FactorVAE metric has clear limitations.
Therefore, to provide a more robust quantitative assessment, we conduct experiments on downstream tasks, where our model's superiority is evident both quantitatively and qualitatively.


\end{document}